\documentclass[10pt,twocolumn,letterpaper]{article}

 \usepackage{cvpr}              %
\usepackage[percent]{overpic}
\pdfoutput=1
\usepackage[dvipsnames]{xcolor}

\usepackage{mathtools}
\usepackage[inkscapelatex=false]{svg}

\definecolor{cvprblue}{rgb}{0.21,0.49,0.74}
\usepackage[pagebackref,breaklinks,colorlinks,citecolor=cvprblue]{hyperref}
\usepackage{multirow}
\usepackage{tikz}
\usepackage[normalem]{ulem}
\usepackage{setspace}
\def\paperID{7398} %
\def\confName{CVPR}
\def\confYear{2024}

\title{SOAC: \textbf{S}patio-Temporal \textbf{O}verlap-\textbf{A}ware Multi-Sensor Calibration \\using Neural Radiance Fields}

\author{
    Quentin Herau$^
    {1,2}$ \qquad
    Nathan Piasco$^{1}$ \qquad
    Moussab Bennehar$^{1}$ \qquad
    Luis Roldão$^{1}$ \\
    Dzmitry Tsishkou$^{1}$ \qquad
    Cyrille Migniot$^{3}$ \qquad
    Pascal Vasseur$^{4}$ \qquad
    Cédric Demonceaux$^{2}$ \\
    $^{1}$Noah's Ark, Huawei Paris Research Center \qquad
    $^{2}$ICB UMR CNRS 6303, Université de Bourgogne\\
    $^{3}$ImViA UR 7535, Université de Bourgogne \qquad
    $^{4}$MIS UR 4290, Université de Picardie Jules Verne\\
    {\tt\small \{Quentin.Herau, Nathan.Piasco, Moussab.Bennehar, Luis.Roldao, Dzmitry.Tsishkou\}@huawei.com}\\
    {\tt\small \{Quentin.Herau@etu., Cyrille.Migniot@, Cedric.Demonceaux@\}u-bourgogne.fr} \\
    {\tt\small Pascal.Vasseur@u-picardie.fr}
    }

\begin{document}
\maketitle
\begin{abstract}

In rapidly-evolving domains such as autonomous driving, the use of multiple sensors with different modalities is crucial to ensure high operational precision and stability. To correctly exploit the provided information by each sensor in a single common frame, it is essential for these sensors to be accurately calibrated. 
In this paper, we leverage the ability of Neural Radiance Fields (NeRF) to represent different sensors modalities in a common volumetric representation to achieve robust and accurate spatio-temporal sensor calibration.
By designing a partitioning approach based on the visible part of the scene for each sensor, we formulate the calibration problem using only the overlapping areas. This strategy results in a more robust and accurate calibration that is less prone to failure. We demonstrate that our approach works on outdoor urban scenes by validating it on multiple established driving datasets. Results show that our method is able to get better accuracy and robustness compared to existing methods.

\end{abstract}

\section{Introduction}

Multi-sensor calibration plays a key role in autonomous systems as it ensures accuracy, reliability, and robustness in safety-critical tasks such as localization~\cite{fayyad2020deep} and perception~\cite{marti2019review} in self-driving.
In typical multi-sensor setups, the sensors are attached to a common rigid body where the spatial relationship between them can be obtained through a rigid transformation matrix.
It is therefore important to identify the exact values of those matrices to correctly exploit and merge the data provided by the sensors.
The process of finding these spatial transformations is called extrinsic calibration, which is a topic that has been and is still being heavily studied thanks to the increasing popularity of multi-sensor algorithms. 
In addition to spatial calibration, without an external synchronization system, it is also necessary to perform temporal calibration. Using temporally miscalibrated sensors, performance on different tasks can be severely hindered. 
Although certain approaches in the literature address temporal misalignment~\cite{taylor2016motion,park2020spatiotemporal,herau2023moisst}, the prevailing assumption among these methods is the presence of properly synchronized sensors.
Due to the importance of sensor calibration, a multitude of calibration solutions exist in the literature, as highlighted in the review from Li et al.~\cite{li2023automatic} and summarized in Tab.~\ref{tab:methods_comparison}. 
They can be classified into two main categories: target-based and targetless methods.
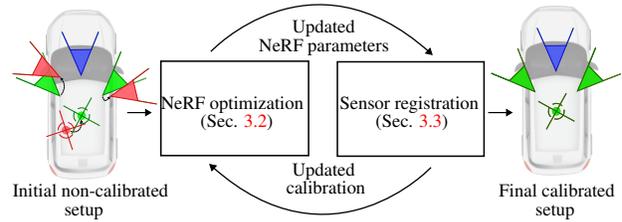
\begin{figure}
    \centering
    \begin{overpic}[width=0.97\linewidth]
    {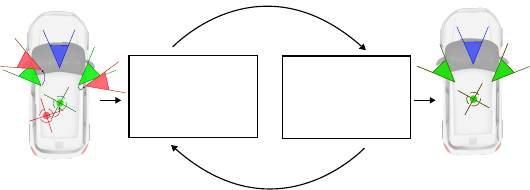}
    \put (24.4,17.5) {\scriptsize{NeRF optimization}}
    \put (30.9,14.5) {\scriptsize{(Sec. \ref{sec/method/step1}})}
    \put (53.8,17.5) {\scriptsize{Sensor registration}}
    \put (59.7,14.5) {\scriptsize{(Sec. \ref{sec/method/step2}})}
    \put (46,30) {\scriptsize{Updated}}
    \put (40.5,27) {\scriptsize{NeRF parameters}}
    \put (46,6.5) {\scriptsize{Updated}}
    \put (45,3.5) {\scriptsize{calibration}}
    \put (0.0,3) {\scriptsize{Initial non-calibrated}}
    \put (8.5,0) {\scriptsize{setup}}
    \put (80,3) {\scriptsize{Final calibrated}}
    \put (86.0,0) {\scriptsize{setup}}
    \end{overpic}
    \vspace{-1mm}
    \caption{\textbf{Method overview.} SOAC is a novel multimodal spatio-temporal calibration method for cameras and LiDAR in the context of autonomous driving. By alternating the training of multiple implicit scenes (Sec.~\ref{sec/method/step1}) and sensors co-registration from these representations (Sec.~\ref{sec/method/step2}), SOAC achieves precise self-supervised calibration from raw data acquired in unconstrained urban environments.}
    \label{fig:teaser}
\vspace{-4mm}
\end{figure}

Target-based calibration methods rely on one or more elements of known dimensions and features purposefully placed in the scene. The most classic target is a checkerboard~\cite{zhang2004extrinsic,geiger2012automatic}, but custom-made planar targets~\cite{guindel2017automatic} or boxes~\cite{pusztai2017accurate} have also been proposed.
These methods usually offer precise and robust calibration compared to targetless approaches. However, requiring hand-placed targets prevents them from being deployed on a large scale and does not enable on-the-fly re-calibration if needed. 
Thus, a more suitable method for mass-produced autonomous driving cars would be targetless. 

Targetless methods do not require manually placed targets and thus can be used on sequences captured without user intervention. This makes them more suitable for large-scale deployment.
These approaches usually rely on shared information (i.e. overlap) between the different sensors, which can be of different modalities. Wang et al.~\cite{wang2012automatic} and Pandey et al.~\cite{pandey2012automatic} propose a correspondence between the reflectivity of the LiDAR scans and the grayscale intensity of the camera images. Other methods propose to find matches of specific features, like edges~\cite{yuan2021pixel} or semantic classes~\cite{kodaira2022sst}.

Following the development of deep learning, methods relying on deep models were introduced to calibrate RGB images and LiDAR scans. These methods have the advantage of being fast and precise, enabling reliable online calibration. Deep learning techniques can leverage regression~\cite{schneider2017regnet,iyer2018calibnet,lv2021lccnet}, flow~\cite{jing2022dxq}, keypoints~\cite{ye2021keypoint} or convolutional features~\cite{fu2023batch} to supervise or regularize the training.
However, as they are supervised methods, they need an accurately calibrated training dataset to be optimized and have issues with cross-domain data due to overfitting to a specific dataset or sensor layout.

Recently, with the arrival of Neural Radiance Fields (NeRF)~\cite{mildenhall2021nerf} for implicit representation of 3D scenes, some works~\cite{herau2023moisst,zhou2023inf,wu2022asyncnerf} propose to take advantage of the fully differentiable structure of the model to achieve self-supervised targetless calibration. Using a NeRF as the common frame for the sensors, these methods are able to densely correlate the captured observation from different sensors in an implicit volumetric space. Yet, by simultaneously learning the information from multiple sensors, the NeRF might overfit regions of the scene only visible from a single sensor without enforcing consistency on the overlapping regions.
This causes the calibration to easily get stuck in a local minimum.

We take inspiration from the aforementioned works by exploiting the fully differentiable properties of the implicit scene representation to achieve spatial and temporal calibration. Different from existing methods~\cite{herau2023moisst,zhou2023inf, wu2022asyncnerf}, we propose to represent the scene by using multiple NeRFs akin to their corresponding sensor and advocate to alternate the optimization target between NeRF training and sensor calibration (i.e. Fig.~\ref{fig:teaser}). Our method avoids overfitting the pose optimization to partial regions of the scene, resulting in a more robust and accurate calibration.

\begin{table}[hbt!]
    \centering
    \resizebox{0.83\columnwidth}{!}
    {
        \renewcommand{\arraystretch}{1.0}
        \setlength{\tabcolsep}{0.028\linewidth}
        \begin{tabular}{@{} l l c c c c c @{}}
         & & \rotatebox{90}{Targetless} & \rotatebox{90}{Cam/Cam} & \rotatebox{90}{Cam/LiDAR} & \rotatebox{90}{Temporal} & \rotatebox{90}{Self-supervised}\\
        \hline
        \multirow{2}{*}{\textbf{Target-based}} & Zhang et al.~\cite{zhang2004extrinsic} & \textcolor{red}{X} & \textcolor{red}{X} & \textcolor{green}{\checkmark} & \textcolor{red}{X} & -\\
        &Geiger et al.~\cite{geiger2012automatic} & \textcolor{red}{X} & \textcolor{red}{X} & \textcolor{green}{\checkmark} & \textcolor{red}{X} & -\\
        \hline
        \multirow{2}{*}{\textbf{Feature-based}} & Pandey et al.~\cite{pandey2012automatic} & \textcolor{green}{\checkmark} & \textcolor{red}{X} & \textcolor{green}{\checkmark} & \textcolor{red}{X} & -\\
        &Park et al.~\cite{park2020spatiotemporal} & \textcolor{green}{\checkmark} & \textcolor{red}{X} & \textcolor{green}{\checkmark} & \textcolor{green}{\checkmark} & -\\
        \hline
        \multirow{2}{*}{\textbf{Deep-learning}} &RegNet~\cite{schneider2017regnet} & \textcolor{green}{\checkmark} & \textcolor{red}{X} & \textcolor{green}{\checkmark} & \textcolor{red}{X} & \textcolor{red}{X}\\
        &LCCNet~\cite{lv2021lccnet} & \textcolor{green}{\checkmark} & \textcolor{red}{X} & \textcolor{green}{\checkmark} & \textcolor{red}{X} & \textcolor{red}{X}\\
        \hline
        \multirow{3}{*}{\textbf{NeRF-based}} &INF~\cite{zhou2023inf} & \textcolor{green}{\checkmark} & \textcolor{red}{X} & \textcolor{green}{\checkmark} & \textcolor{red}{X} & \textcolor{green}{\checkmark}\\
        &MOISST~\cite{herau2023moisst} & \textcolor{green}{\checkmark} & \textcolor{green}{\checkmark} & \textcolor{green}{\checkmark} & \textcolor{green}{\checkmark} & \textcolor{green}{\checkmark}\\
        &SOAC (ours) & \textcolor{green}{\checkmark} & \textcolor{green}{\checkmark} & \textcolor{green}{\checkmark} & \textcolor{green}{\checkmark} & \textcolor{green}{\checkmark}\\
        
        \bottomrule
        \end{tabular}
    }
    \caption{\label{tab:methods_comparison} Comparison of calibration methods.}
    \vspace{-0.45cm}
\end{table}

\section{Related Work} \label{sec/related_works}

With NeRF and the papers improving upon it~\cite{barron2022mip,muller2022instant}, the main focus was on the quality of novel view synthesis in addition to training and rendering speeds. However, since these approaches often validate their claims on carefully curated datasets, it is often assumed that the input poses corresponding to the data are already available and are accurate. However, in real-world situations, some or all the captured frames might be unposed or suffer from inaccuracies, hence, significantly impacting the quality of the final reconstruction result~\cite{lin2021barf}. Therefore, several works later on attempted to tackle this issue through different formulations and adaptations of the overall optimization problem. 

\paragraph{NeRF-based Image Registration.}
\label{sec/related_works/registration}
To register an image with incorrect or no pose, iNeRF~\cite{yen2020inerf} proposes to use an already trained NeRF. It finds the pose that minimizes the photometric difference between the captured image and the rendered result from the model. By focusing on regions of interest, it is able to register unseen images with high precision.
Using this idea as a basis, Loc-NeRF~\cite{maggio2023loc} combines Monte Carlo localization method~\cite{dellaert1999monte} with the use of a pretrained NeRF as a map, to build a real-time global localization method.
CROSSFIRE~\cite{moreau2023crossfire} takes advantage of the NeRF model's flexibility to learn not only the radiance and density information of the map, but also a descriptor field. During the localization process, by iteratively matching the descriptors from the query image and the information given by the NeRF model, this method is able to provide high-precision localization.
Nevertheless, all these methods require training a NeRF from precise camera poses first before being able to localize new query images.

\paragraph{NeRF-based Pose Optimization.}\label{sec/related_works/regression} 
The first method to leverage the fully differentiable nature of NeRF to optimize the input poses through backpropagation is $\text{NeRF-{}-}$~\cite{wang2021nerf}. It proposes to optimize both the NeRF and the input poses by representing them as embeddings
and show higher novel view synthesis quality when trained from noisy poses.
BARF~\cite{lin2021barf} improves upon this idea by adding a coarse-to-fine component to this method. It progressively liberates the frequencies of the input positional encoding to prevent the optimization from getting stuck in a local minimum. SCNeRF~\cite{jeong2021self} adds camera distortion estimation and uses a different 6-vector rotation formulation in the optimization, while SPARF~\cite{truong2023sparf} achieves pose optimization with sparse input views by relying on pixel matching and depth consistency.
While the aforementioned methods need an initial estimate of the camera poses, some recent methods completely remove the need for prior poses.
NoPe-NeRF~\cite{bian2023nope} uses an off-the-shelf monocular depth estimator (i.e.  DPT~\cite{ranftl2021vision}) to regularize relative poses between successive images.
GNeRF~\cite{meng2021gnerf} relies on adversarial learning to coarsely estimate the initial poses before refining them in a second phase. IR-NeRF~\cite{zhang2023pose} improves upon GNeRF by regularizing the implicit pose estimator with the unposed real images, increasing its robustness.
Although the NeRF-based pose optimization methods achieve reasonable scene reconstruction by recovering accurate camera poses, 
they are not suited for autonomous driving data as they do not handle multi-modal observations nor take into account the rigidity constraint between multiple sensors mounted on a vehicle.

\paragraph{NeRF-based Sensor Calibration.}
\label{sec/related_works/calibration}
NeRF-based calibration methods~\cite{herau2023moisst,zhou2023inf, wu2022asyncnerf} take advantage of the rigid constraint between the sensors and the differentiable nature of NeRF to efficiently solve this challenging task.
These methods have the advantage of being targetless and self-supervised, as they do not rely on an annotated training dataset. The idea is to use the NeRF as a common scene representation. Each sensor provides its observations (RGB images, depth measurement, or point clouds), to both train the NeRF to represent the scene and to optimize its own extrinsic calibration parameters to fit the NeRF representation.
In INF~\cite{zhou2023inf}, the goal is to find the extrinsic transformation between a 360° camera and a LiDAR. First, the density network of NeRF is trained using the LiDAR depth data. Then, the whole scene's radiance is trained using images, while simultaneously calibrating the camera. 
This method is limited to the calibration of a single 360° camera and a LiDAR, whereas autonomous driving systems rely on multiple cameras with narrower fields of view. 
AsyncNeRF~\cite{wu2022asyncnerf} calibrates a pair of camera and depth sensors. It takes into account the temporal miscalibration between the sensors, by building a trajectory function. Nevertheless, the time offset is provided as input and not determined through optimization, which limits its utilization for spatio-temporal calibration.
MOISST~\cite{herau2023moisst} proposes to accomplish temporal calibration in addition to extrinsic calibration, and to do so with any number of LiDARs and cameras, by training the NeRF with all the data, while also optimizing the prior extrinsic transformations and time offsets.
By using a single NeRF to fuse the information from all the sensors, we cannot prevent degenerate cases where the estimation of the extrinsic parameters of one sensor diverges and causes the NeRF to learn a wrong scene geometry without correlating multi-sensor observations.
Our method, SOAC, aims to achieve better robustness and calibration performance by leveraging the use of multiple NeRFs to counterbalance such limitations.

\begin{figure*}
\centering
\begin{subfigure}{0.45\textwidth}
    \includegraphics[height=0.55\columnwidth]{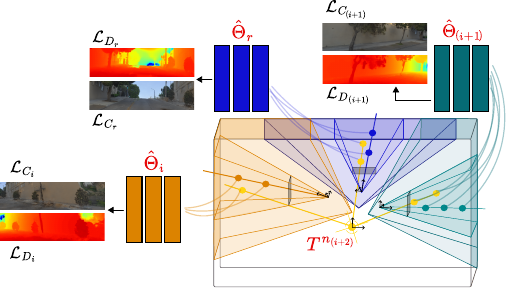} 
    \caption{}
    \label{fig:method/step1}
\end{subfigure}
\hspace{0.03\linewidth}
\vline
\hspace{0.03\linewidth}
\begin{subfigure}{0.45\textwidth}
    \includegraphics[height=0.55\columnwidth]{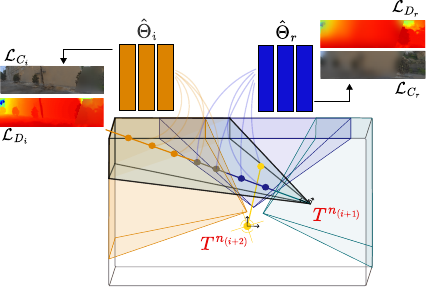} 
    \caption{}
    \label{fig:method/step2}
\end{subfigure}
\vspace{-0.2cm}
	\caption{\textbf{SOAC training strategy.}  \protect\subref{fig:method/step1} Scene representation training (Sec. \ref{sec/method/step1}): The parameters $\textcolor{red}{\hat{\Theta}}$ of each NeRF are trained with the images from their associated cameras and the LiDAR scans. The LiDAR calibration is also optimized through $\textcolor{red}{T^{n_{(i+2)}}}$. \protect\subref{fig:method/step2} Extrinsic and temporal optimization (Sec. \ref{sec/method/step2}): The real frame from the sensor is compared to the predicted frame on the other NeRFs to calculate the losses. The calibration is then optimized with backpropagation through the poses $\textcolor{red}{T^{n_{(i+1)}}}$ and $\textcolor{red}{T^{n_{(i+2)}}}$.}
\label{fig:method}
\vspace{-2mm}
\end{figure*}

\section{Method} \label{sec/method}

Our multi-sensor calibration problem is formulated as follows: given a vehicle trajectory and initial priors of sensor poses mounted on the vehicle%
, we aim to recover the exact spatio-temporal calibration of the sensors on the vehicle. Our method is composed of two optimization steps that are performed sequentially all along the training (cf. Fig.~\ref{fig:teaser}). The first step consists of training multiple implicit scene representations (NeRFs), one by camera, using only the observations from the dedicated sensor. During the second optimization step, we refine the extrinsic and temporal parameters of each sensor using the trained NeRF of all the other sensors in a round-robin manner. The motivation behind this design is to prevent over-fitting, calibration divergence, or implicit model convergence to a poor local minimum when all the observations are fused within the same implicit representation, as in MOISST~\cite{herau2023moisst}.
\subsection{Notations and Background} \label{sec/method/notations}

Without loss of generality, we consider the trajectory of camera $r$ (our reference sensor) as the known trajectory of the vehicle.
We use the same notations introduced in MOISST \cite{herau2023moisst} to describe our method:
\begin{itemize}
    \item $S=\{C,L\}$: the set of sensors composed of at least one or more cameras $C$ and, optionally, one or more LiDARs $L$,
    \item $\{F_i\}$: the set of frames captured by the sensor $i\in S$,
    \item $t^{n_i} \in \mathbf{R^+} $: the timestamp of frame $n_i\in F_i$ relative to the sensor $i\in S$,
    \item $\delta_i \in \mathbf{R}$: the time offset between the reference camera and the sensor $i\in S$ ($\delta_r=0$),
    \item $\prescript{}{w}{T}^i(t)\in \mathbf{R}^{4\times 4}$: the pose of sensor $i\in S$ at time $t$ (the time is relative to sensor $i$'s own clock) in the world reference frame,
    \item $\prescript{}{j}{T}^i \in \mathbf{R}^{4 \times 4}$: the transformation matrix from sensor $i$ to sensor $j$.
\end{itemize}

Our goal is to find the optimal transformations $\hat{\prescript{}{r}{T}^i}$ and time offsets $\hat{\delta_i}$ of the different sensors with respect to the reference camera. The poses of the reference camera $r$ can be obtained by relying on IMU, SLAM~\cite{mur2015orb}, or Structure-from-Motion~\cite{colmap}. Similar to MOISST, we build a continuous trajectory of the reference sensor $r$,  $\mathcal{T}_r$, from the discrete poses of $r$ using linear interpolation for the pose translation and spherical linear interpolation (SLERP~\cite{shoemake1985animating}) for the rotation. This trajectory is expressed as a function of time, that returns the pose of the reference camera $r$ for any given time $t$: $\prescript{}{w}{T}^r(t) = \mathcal{T}_r(t)$.
Using the extrinsic transformations and the time offsets between the other sensors and the reference camera $r$, we can compute the absolute pose of sensor $i$ at specific timestamps with the following equation:
\begin{equation}
     \prescript{}{w}{T}^i(t^{n_i}+\delta_{i}) =\mathcal{T}_r(t^{n_i}+\delta_{i}) \prescript{}{r}{T}^i.
     \label{eq:ext}
\end{equation}
In order to simplify the equations, we designate the absolute pose of sensor $i$ computed from its extrinsic as $T^{n_i} = \prescript{}{w}{T}^i(t^{n_i}+\delta_{i})$.

\paragraph{NeRF model.}
NeRF is a function of parameters $\Theta$ that takes as input rays obtained from a sensor's intrinsic parameters and pose, and generates for each ray color and density information via volumetric rendering. This information can be combined into a color image $\mathcal{R}_I\left({T^{n_i}} \mid \Theta\right)$ and a depth scan $\mathcal{R}_D\left({T^{n_i}} \mid \Theta\right)$ of frame $n_i$ for sensor $i$.

\subsection{Scene Representation Training} \label{sec/method/step1}
For each camera sensor, a dedicated NeRF with parameters $\Theta_i$ is trained using rays that are generated exclusively from camera $i$.
Each NeRF model with parameters $\Theta_i$ will only learn the part of the scene that is observed by its respective camera sensor $i$ (cf. Fig.~\ref{fig:method/step1}).
The color loss for training the scene representation is:
\begin{equation}
    \mathcal{L}_{C} = \sum_{i\in C} \sum_{n_i\in F_i}\left\|\mathcal{R}_I\left({T^{n_i}} \mid \Theta_i\right)-{I}^{n_i}\right\|_{2}^{2}, \label{eq:color_loss_scene}\\
\end{equation}
with ${I}^{n_i}$ the color image $n_i$ of camera $i$. 
The training objective is to estimate the optimal parameters $\hat{\Theta_i}$ for the NeRF models such as:
\begin{equation}
    \left\{ \hat{\Theta}_i \right\}_{i \in C } =\underset{\left\{ \Theta_i \right\}_{i \in C }}{\text{argmin}}(\mathcal{L}_{C}).
    \label{eq:obj_scene}
\end{equation}
\subsection{Extrinsic and Temporal Optimization} \label{sec/method/step2}
During the calibration step, our objective is to optimize the extrinsic transformation matrix $\prescript{}{r}{T}^{i}$ and temporal parameters $\delta_{i}$ by optimizing the poses of camera $i$ using all the NeRFs, except the NeRF of parameters $\Theta_i$ associated to the current camera being calibrated (cf. Fig.~\ref{fig:method/step2}). Using this optimization formulation, we enforce the images captured by each camera to be coherent with the NeRF trained by the other cameras. The camera calibration loss can be written as:
\begin{equation}
    \mathcal{L}_{Cam} = \sum_{j\in C} \sum_{\substack{i\in C \\ i\neq j}} \sum_{n_i\in F_i}\left\|\mathcal{R}_I\left({T^{n_i}} \mid \Theta_j\right)-{I}^{n_i}\right\|_{2}^{2}, \label{eq:color_loss_calib}\\
\end{equation}
and by considering Eq.~\ref{eq:ext}, the optimization objective during the spatio-temporal optimization step is:
\begin{equation}
    \left\{ \hat{\prescript{}{r}{T}^{i}}, \hat{\delta_{i}} \right\}_{i \in  C } =\underset{\left\{ \prescript{}{r}{T}^{i}, \delta_{i} \right\}_{i \in C }}{\text{argmin}}(\mathcal{L}_{Cam}).
    \label{eq:obj_calib}
\end{equation}

\subsection{LiDAR Calibration} \label{sec/method/LiDAR}
As LiDARs only provide geometric information, we cannot register an RGB image to a NeRF which was only trained on LiDAR scans. This means that the registration step (cf. Sec.~\ref{sec/method/step2}) could not be accomplished on a LiDAR-trained NeRF. Instead of dedicating a NeRF for each LiDAR, we simultaneously train the camera NeRFs with all the LiDAR scans, and calibrate the LiDARs against all NeRFs (cf. Fig.~\ref{fig:method}).
Thus, we have for both the NeRF training step and calibration step:
\begin{equation}
\mathcal{L}_{D} = \sum_{j\in C} \sum_{i\in L} \sum_{n_i\in F_i}\lvert\mathcal{R}_D\left({T^{n_i}} \mid \Theta_j\right)-{D}^{n_i}\rvert, \label{eq:depth_loss}
\end{equation}
with ${D}^{n_i}$ the point cloud scan $n_i$ of LiDAR $i$.
When adding the LiDAR loss in the objective Eq.~\ref{eq:obj_scene}, it becomes:
\begin{equation}
    \left\{ \hat{\Theta}_i \right\}_{i \in  C },  \left\{ \hat{\prescript{}{r}{T}^{j}}, \hat{\delta}_i \right\}_{j \in L } =\underset{\left\{ \Theta_i \right\},  \left\{ \prescript{}{r}{T}^{j}, {\delta_{j}} \right\}}{\text{argmin}}(\mathcal{L}_{C} + \mathcal{L}_{D}).
\end{equation}
and Eq.~\ref{eq:obj_calib} becomes:
\begin{equation}
    \left\{ \hat{\prescript{}{r}{T}^{i}}, \hat{\delta}_i \right\}_{i \in S}=\underset{\left\{ \prescript{}{r}{T}^{i}, \delta_{i} \right\}}{\text{argmin}}(\mathcal{L}_{Cam} + \mathcal{L}_{D}).
    \label{eq:final_obj_calib}
\end{equation}

\subsection{Visibility Grid} \label{sec/method/filtergrid}
\begin{figure}[!t]
\centering
\begin{subfigure}{0.37\linewidth}
    \centering
    \includegraphics[width=0.9\textwidth]{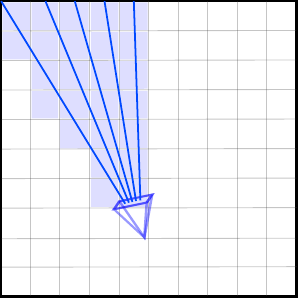} 
    \caption{}
\label{fig:filter_grid/filling}
\end{subfigure}
\hspace{0.05\linewidth}
\begin{subfigure}{0.37\linewidth}
    \centering
    \includegraphics[width=0.9\textwidth]{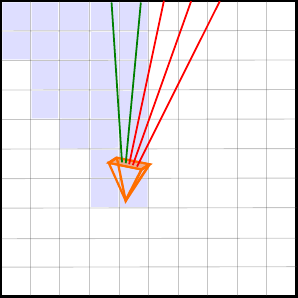} 
    \caption{} 
    \label{fig:filter_grid/filtering}
\end{subfigure}
\vspace{-2mm}
\caption{\textbf{SOAC's visibility grid} (Sec. \ref{sec/method/filtergrid}). \protect\subref{fig:filter_grid/filling} Grid filling: \textcolor{blue}{Rays} from camera \textcolor{blue}{$C_i$} fill the visibility grid linked to Nerf \textcolor{blue}{$\Theta_i$}. \protect\subref{fig:filter_grid/filtering} Ray filtering: For cameras $\textcolor{orange}{C}_{\textcolor{orange}{j, \forall j} \neq \textcolor{blue}{i}}$, rays are \textcolor{ForestGreen}{kept} or \textcolor{red}{filtered} according to visibility from \protect\subref{fig:filter_grid/filling}.}
\label{fig:filter_grid}
\vspace{-2mm}
\end{figure}

\begin{figure}[!htbp]
\centering
\scriptsize
\setlength{\tabcolsep}{0.002\linewidth}
    \begin{tabular}{lccc}
           \rotatebox{90}{~~~Input RGB} & \includegraphics[width=0.31\linewidth]{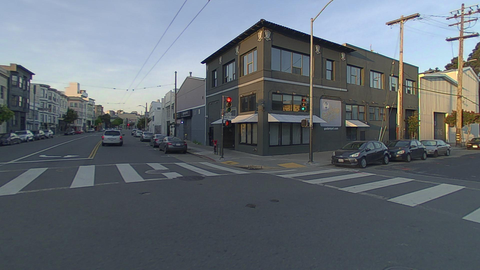} &
         \includegraphics[width=0.31\linewidth]{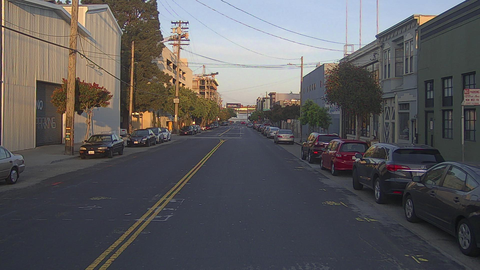} &
         \includegraphics[width=0.31\linewidth]{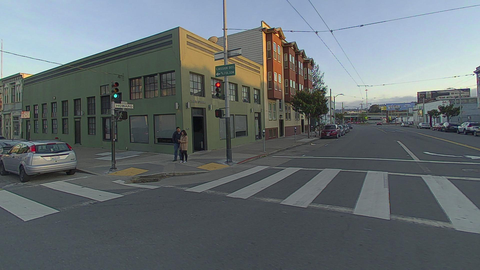}\\

          \rotatebox{90}{~~~Pred. RGB} &\includegraphics[width=0.31\linewidth]{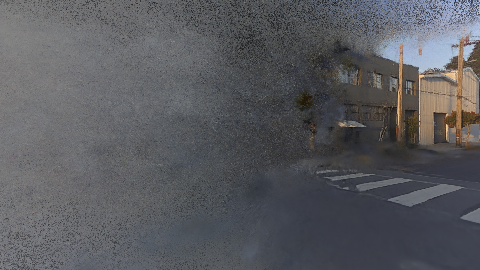} &
         \includegraphics[width=0.31\linewidth]{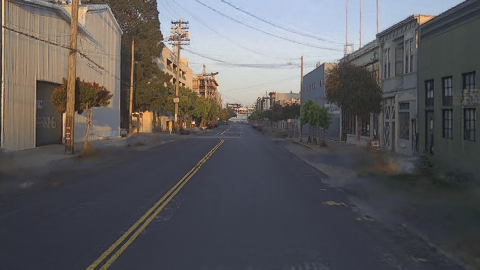} &
         \includegraphics[width=0.31\linewidth]{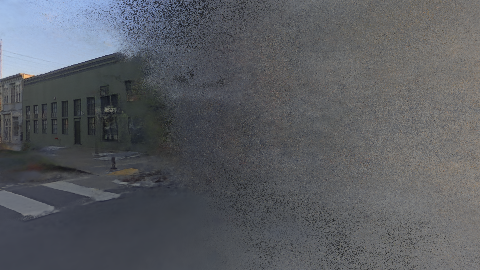}\\

           \rotatebox{90}{~~Pred. depth} &\includegraphics[width=0.31\linewidth]{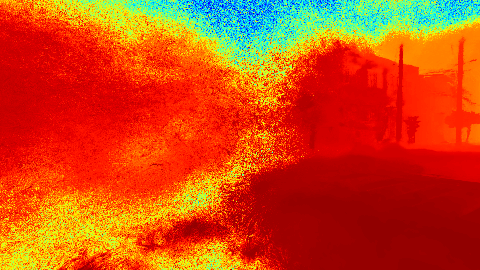} &
         \includegraphics[width=0.31\linewidth]{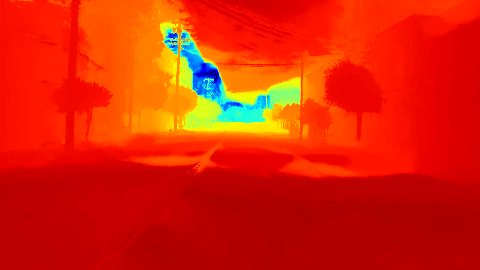} &
         \includegraphics[width=0.31\linewidth]{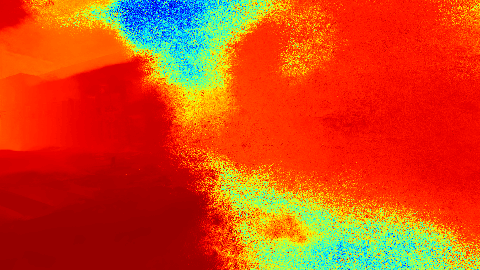}\\

     \rotatebox{90}{~~~Vis. mask} &\includegraphics[width=0.31\linewidth]{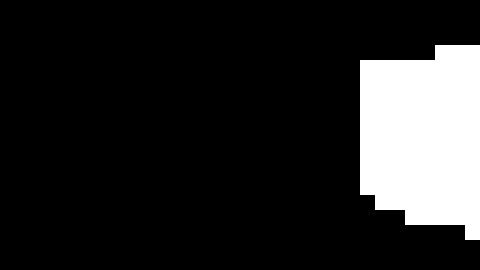} &
         \includegraphics[width=0.31\linewidth]{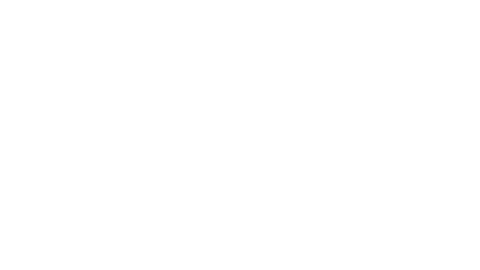} &
         \includegraphics[width=0.31\linewidth]{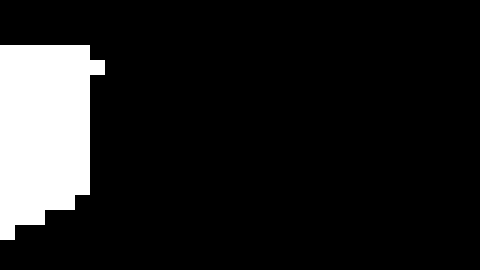}\\

    \end{tabular}
\vspace{-0.2cm}
\caption{\textbf{Visualization of the visibility grid} (Sec.~\ref{sec/method/filtergrid}). Predictions done with the NeRF trained by front camera on a Pandaset~\cite{xiao2021pandaset} sequence.}
\label{fig:result_example}
\vspace{-2mm}
\end{figure}

In a multi-sensor setup, the NeRF representation exploits the overlap between sensors w.r.t. the whole sequence rather than a particular frame as for traditional targetless methods. However, the portions of the scene observed from the different sensors might not entirely overlap. This can lead to noisy reconstruction in the NeRF model if inference is performed at the unobserved regions (cf. Fig.~\ref{fig:result_example}). To overcome this problem, NeRF2NeRF~\cite{goli2023nerf2nerf} performs pairwise registration of two NeRF models produced from different viewpoints by aligning the partially overlapping geometry of the two models. In a similar sense, we aim to consider the overlapping geometry from our different NeRFs that have been learned separately from each camera.\\
To achieve this, a boolean visibility grid for each NeRF model is reconstructed by considering the rays belonging to its akin sensor (see Fig.~\ref{fig:filter_grid/filling}) during the scene representation step (Sec.~\ref{sec/method/step1}). During the calibration step (Sec.~\ref{sec/method/step2}) we exploit this visibility grid to only consider rays that overlap with trained regions on each NeRF used for registration (cf. Fig.~\ref{fig:filter_grid/filtering}). The grids are reinitialized every few epochs to account for the new poses resulting from the calibration refinements. 

\subsection{Optimization Details} \label{sec/method/details}
Overall, the training process can be summarized as follows: during each training step, a mini-batch of rays is first used in the scene representation training step (Sec.~\ref{fig:method/step1}). Rays of each camera train their specific NeRF and fill the respective visibility grids (Sec.~\ref{sec/method/filtergrid}). The LiDAR rays train all the NeRFs and are used to optimize the LiDAR calibration parameters after being filtered by the visibility grids.
In a subsequent step, the same mini-batch is passed to the extrinsic and temporal optimization. Rays are filtered through the visibility grids before being fed to the NeRFs as explained in Sec.~\ref{sec/method/step2}. Calibration losses (Eq.~\ref{eq:final_obj_calib}) are computed and the gradient is backpropagated to optimize the calibration parameters.
Once this is done, we continue the training with the next mini-batch.

\paragraph{NeRF delaying.}
In our system, all the sensors, except the reference camera, have incorrect calibration. As such, the NeRF trained with the reference camera is the most adequate for calibration at the beginning. That is why we introduce a delaying schedule for the other NeRFs based on the overlap with the reference camera; more details about this policy are provided in the supplementary materials. 

\paragraph{Correction bounding.}
As we consider the extrinsic and temporal calibration on a car, we can suppose that the translation error should not be off by more than the car's size. We can also consider that the sensors should not have a time offset too high, even without the help of an external synchronizing system. Thus, by using an offset and scaled sigmoid function on the output of the embeddings for the translation and temporal correction, we can confine the learned correction, avoiding divergence and increasing the stability and robustness of the calibration.

\begin{figure}[!htbp]
\centering
\setlength{\tabcolsep}{0.001\linewidth}
    \begin{tabular}{cc}
        KITTI-360 & Nuscenes\\
         \includegraphics[height=.47\linewidth]{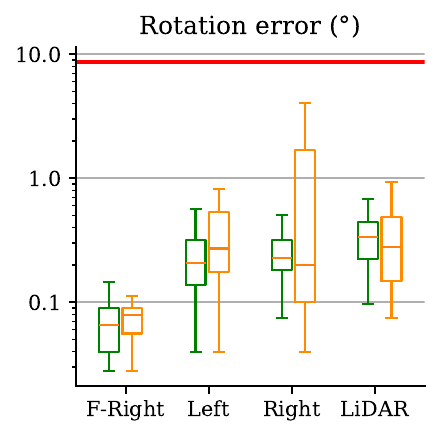} &  \includegraphics[height=.48\linewidth]{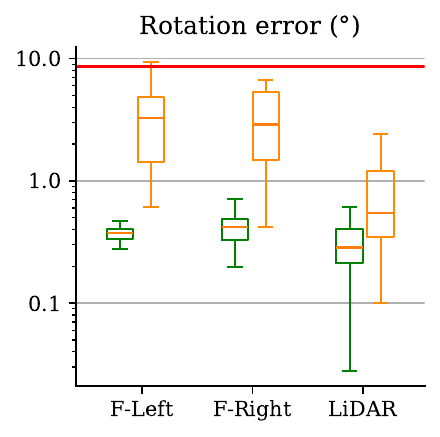}\\  \includegraphics[height=.47\linewidth]{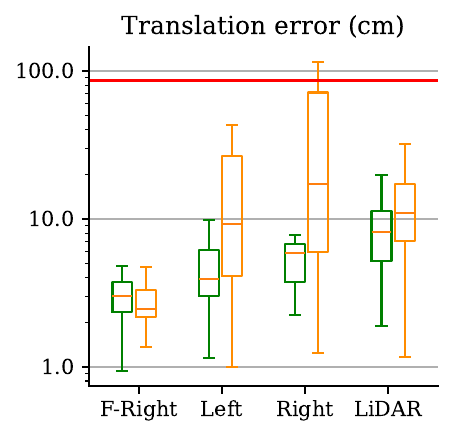}&
         \includegraphics[height=.47\linewidth]{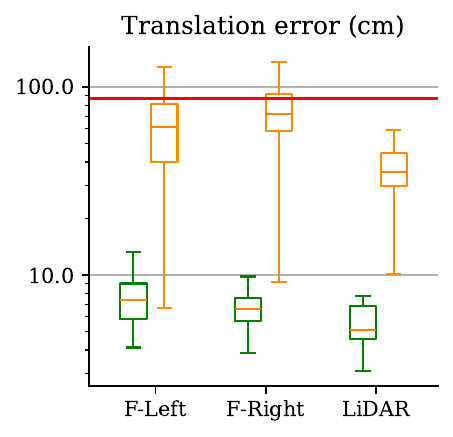}\\ \includegraphics[height=.48\linewidth]{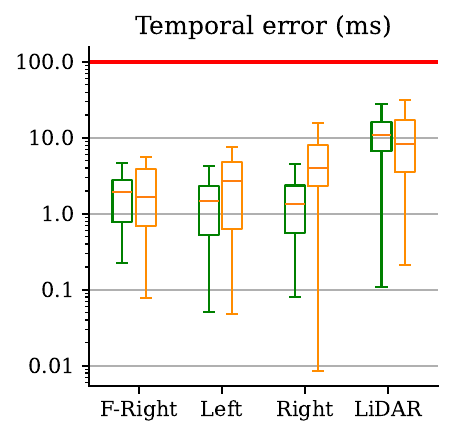} & \includegraphics[height=.47\linewidth]{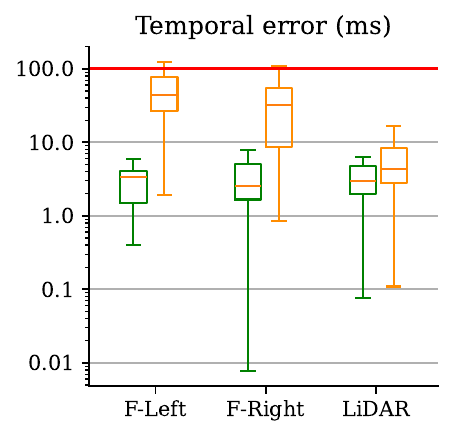}

    \end{tabular}
\vspace{-0.3cm}
\caption{Results for \textcolor{ForestGreen}{SOAC} and \textcolor{orange}{MOISST}~\cite{herau2023moisst} as box plots with log scale on KITTI-360~\cite{liao2022kitti} and Nuscenes~\cite{caesar2020nuscenes} sequences. The \textcolor{red}{red lines} show the initial error (best viewed in color).}
\label{fig:results}
\vspace{-2mm}
\end{figure}

\section{Experiments} \label{sec/experiments}
\subsection{Setup} \label{sec/experiments/setup}
\paragraph{Datasets.}
We perform experiments on three popular autonomous driving datasets: KITTI-360~\cite{liao2022kitti}, nuScenes~\cite{caesar2020nuscenes} and Pandaset~\cite{xiao2021pandaset}. 
For KITTI-360, we use the two front cameras, the two side cameras and the Velodyne LiDAR for our experiments. For nuScenes and Pandaset, we use the front camera, the two front diagonal cameras, and the LiDAR. Undistorted LiDAR scans are considered for all datasets.
We assign the front-left camera of KITTI-360, and the front cameras of nuScenes and Pandaset to be the reference sensor.
More details on selected sequences and dataset parameters are provided in the supplementary materials.
\begin{figure*}[!hbtp]
\centering
    \begin{tabular}{ccc}
         \includegraphics[width=.24\linewidth]{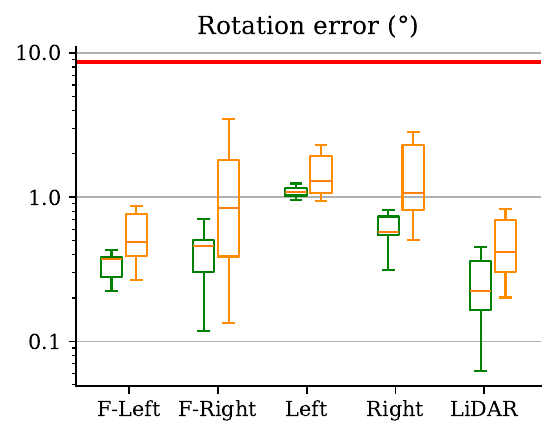} &  \includegraphics[width=.24\linewidth]{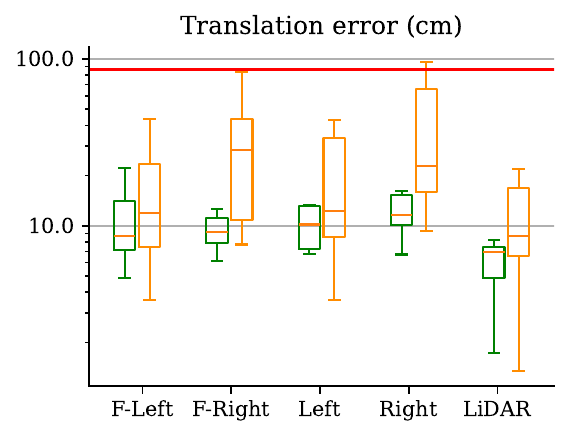} &
         \includegraphics[width=.24\linewidth]{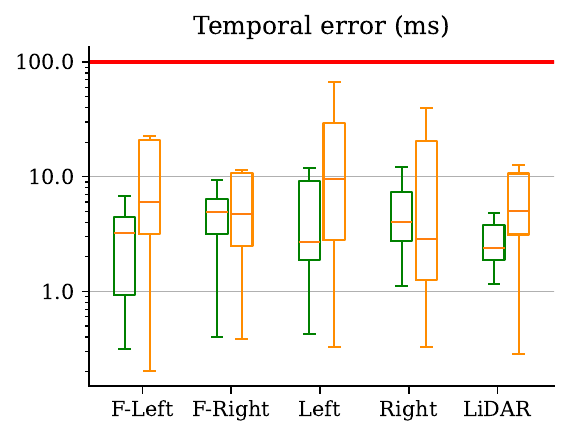}
    \end{tabular}
\vspace{-3mm}
\caption{Results on nuScenes~\cite{caesar2020nuscenes} with 5 cameras for \textcolor{ForestGreen}{SOAC} and \textcolor{orange}{SOAC w/o NeRF delaying} as box plots with log scale, the \textcolor{red}{red lines} show the initial error (best viewed in color).}
\label{fig:5cams}
\vspace{-2mm}
\end{figure*}

\paragraph{Baseline.}
We select MOISST~\cite{herau2023moisst} as our baseline, as it also aims to solve targetless, multi-modal, and spatiotemporal calibration. We refer to the supplementary for details about the re-implementation of the method.
For the LiDAR/Camera calibration task, we compare against LCCNet~\cite{lv2021lccnet} by using the code and the pre-trained weights from the official repository\footnote{https://github.com/IIPCVLAB/LCCNet}, and with Pandey et al.~\cite{pandey2012automatic} using the official implementation provided by authors\footnote{https://robots.engin.umich.edu/SoftwareData/InfoExtrinsicCalib}. For SOAC, We replicate MOISST NeRF architecture and apply the same supervision and regularization losses. We refer to the supplementary for more implementation details.

\subsection{Results} \label{sec/experiments/results}
\paragraph{Spatial and temporal calibration.}
We run both SOAC and MOISST on 4 KITTI-360 sequences and 3 nuScenes sequences.
For SOAC, KITTI-360 images are downscaled by 4, while a downscale factor of 6 is applied for nuScenes. For MOISST, we do not downscale the KITTI-360 images and apply a downscale factor of 2 for nuScenes as we found that the method performs better with high-resolution images.
Each test is run with an initial noise of 50 cm translation error and 5° rotation error on each axis, as well as 100 ms of time offset. We use 10 different seeds to randomly sign the error noises applied and compute the statistics over these 10 runs. Following common practices \cite{taylor2016motion, rehder2014spatio}, we show results on Fig.~\ref{fig:results} by employing box plots\footnote{The boxes show the first quartile $Q_1$, median, third quartile $Q_3$. The whiskers use 1.5 IQR (Interquartile range) above and below the box and stop at a value within the results.}
As can be seen, SOAC achieves better calibration results on KITTI-360 with an overall error (average over median for each sensor) of 0.21°, 5.24 cm and 3.95 ms for rotation, translation and time offset, respectively. In contrast, MOISST obtains errors about 10 times higher (i.e. 2.24°, 56.34 cm and 27.07 ms) for the same setup. Detailed quantitative results by sequence are given in the supplementary materials.

\begin{table}[bt!]
    \centering
    \resizebox{1.0\columnwidth}{!}
    {
        \renewcommand{\arraystretch}{1.0}
        \setlength{\tabcolsep}{0.015\linewidth}
        \begin{tabular}{l cc | cc }
        & \multicolumn{2}{c|}{KITTI-360~\cite{liao2022kitti}} & \multicolumn{2}{c}{Pandaset~\cite{xiao2021pandaset}} \\
        \cmidrule(lr){2-5}
         & Rotation (°) & Translation (cm) & Rotation (°) & Translation (cm) \\
        \midrule
        Pandey et al.~\cite{pandey2012automatic} & $11.8 \pm 5.4$ & $143 \pm 109$  &  $15.4 \pm 0.8$ & $139 \pm 17.5$ \\
        LCCNet~\cite{lv2021lccnet} & $1.9 \pm 0.1$ & $95.8 \pm 7.7$  &  $14.3 \pm 3.4$ & $370 \pm 11.6$ \\
        MOISST~\cite{herau2023moisst} & $\mathbf{0.2 \pm 0.1}$ & $10.0 \pm 9.8$ & $2.8 \pm 2.3$ & $56.4 \pm 17.2$ \\
        SOAC~(ours) & $0.3 \pm 0.2$ & $\mathbf{7.8 \pm 3.5}$  & $\mathbf{1.3 \pm 0.8}$ & $\mathbf{29.4 \pm 13.6}$\\
    
        \end{tabular}
    }
    \caption{\label{tab:lidar_camera_avg} LiDAR/Camera calibration results.}
    \vspace{-0.35cm}
\end{table}

\paragraph{LiDAR/Camera calibration.}
For the task of LiDAR/Camera calibration, the same initial rotation and translation error setup from previous experiments is applied, but without considering any temporal error. We compare our method against LCCNet~\cite{lv2021lccnet} and Pandey et al.~\cite{pandey2012automatic}. The provided weights for LCCNet were pre-trained on the KITTI odometry dataset~\cite{geiger2012we}. For KITTI-360, We predict the calibration between the front-left camera and the LiDAR. For Pandaset, we predict the calibration between the front camera and the 360° LiDAR. Results are shown in Tab.~\ref{tab:lidar_camera_avg}. The performance of LCCNet, is very poor in comparison to SOAC, especially for the translation (results per sequence are provided in the supplementary). As LLCNet is a supervised method, we observe that it is setup-specific, and a slight change in the LiDAR/Camera configuration greatly reduces the performance. This was also highlighted by Fu et al.~\cite{fu2023batch} when using the front-right camera for calibration on the KITTI odometry dataset.
For Pandey et al.~\cite{pandey2012automatic}, we were unable to obtain convincing calibration results on the sequences. We argue that feature-based targetless methods are not designed for “in-the-wild” calibration, and sequences need to be acquired in a specific manner to obtain proper results (i.e. indoor, structured environment, dense LiDAR).
\begin{figure}[!t]
\vspace{-3mm}
\centering
\setlength{\tabcolsep}{0.001\linewidth}
    \begin{tabular}{cc}
         \includegraphics[height=.44\linewidth]{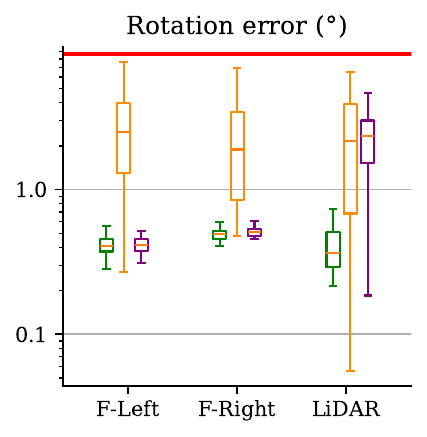} &  \includegraphics[height=.44\linewidth]{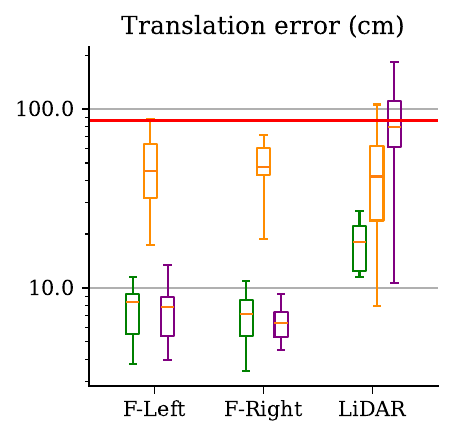}
    \end{tabular}
\vspace{-3mm}
\caption{Results on Pandaset~\cite{xiao2021pandaset} for \textcolor{ForestGreen}{SOAC}, \textcolor{orange}{MOISST}~\cite{herau2023moisst} and \textcolor{violet}{SOAC w/o semantic filtering} as box plots with log scale, the \textcolor{red}{red lines} show the initial error (best viewed in color).}
\label{fig:pandaset}
\vspace{-3mm}
\end{figure}

\paragraph{Calibration in dynamic environments.}
For the evaluation in dynamic environments, we select 3 Pandaset sequences with the presence of dynamic elements (e.g. cars, pedestrians). When calibrating on dynamic scenes, the moving elements are not handled by the NeRF model. Therefore, a simple and efficient way of removing these elements is to filter the dynamic classes with semantic segmentation. This results in losing some useful information for calibration (i.e. parked vehicles). Nevertheless, if the rest of the scene provides sufficient overlap, proper calibration can be obtained. We apply an analogous setup to KITTI-360 and nuScenes, except for the removal of temporal calibration and the initial time offset (cf. Sec.~\ref{sec/experiments/limits} on Time-space compensation). We downscale the image by a factor of 4 for SOAC and 2 for MOISST. We use semantic segmentation computed by Mask2Former~\cite{cheng2022masked} to remove all classes that can be considered dynamic for both methods and test SOAC with and w/out semantic filtering. Results are shown in Fig.~\ref{fig:pandaset}. It can be observed that by applying semantic filtering, %
calibration performance on SOAC can be greatly improved on the LiDAR with a median error of 0.41° and 7.79 cm on rotation and translation, respectively, in comparison to results w/out filtering (2.36° / 79.17 cm). It can be also seen that SOAC performs much better than MOISST on the overall calibration of all the sensors with a mean error an of 0.42° / 11.18 cm vs. 2.18° / 44.73 cm for MOISST. 

\paragraph{Complete camera rig calibration.}
To evaluate SOAC performances with a nearly complete 360° camera rig, we add two additional side cameras on the nuScenes sequences. We run both with and without the NeRFs delaying scheduling as explained in Sec.~\ref{sec/method/details}. In Fig.~\ref{fig:5cams} we can see the impact of not delaying the NeRFs, as the accuracy and stability of the calibration plummet.
\begin{figure}[!t]
\centering
\includegraphics[width=0.8\linewidth]{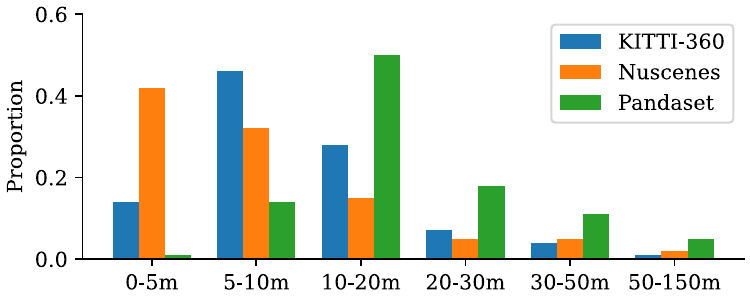} 
\vspace{-2mm}
\caption{LiDAR ray length distribution of the sequences used in our calibration experiments.}
\label{fig:lidar_length}
\vspace{-1mm}
\end{figure}

\begin{figure}[!t]
\centering
\scriptsize
\setlength{\tabcolsep}{0.002\linewidth}
\begin{tabular}{lcc}
     \rotatebox{90}{~~~~~~~~~~~~~~GT} & \includegraphics[width=0.58\linewidth, trim={5cm 0 15cm 0}, clip]{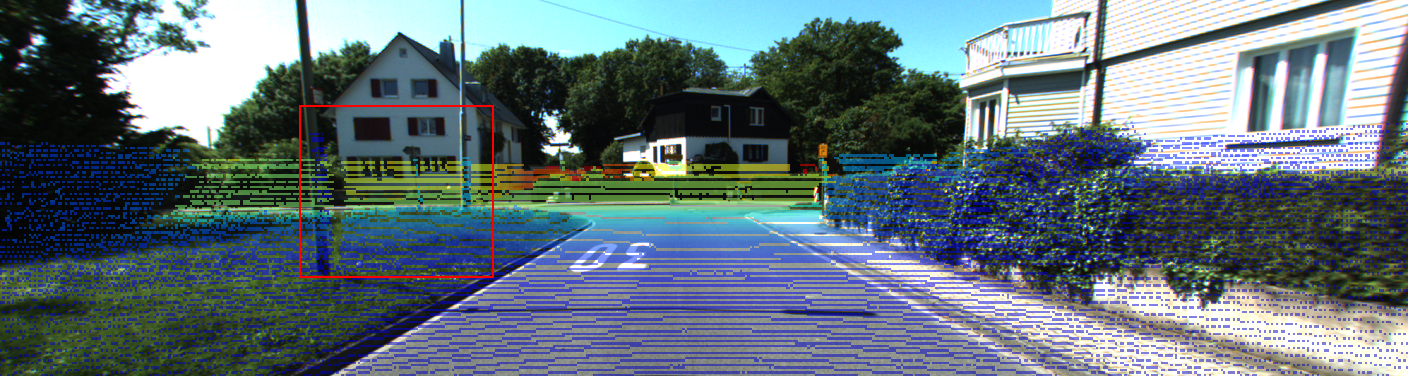}
     & \includegraphics[width=0.293\linewidth]{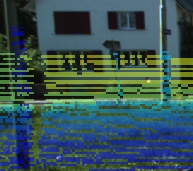}\\
     \rotatebox{90}{~~~~~~~~LCCNet~\cite{lv2021lccnet}} & \includegraphics[width=0.58\linewidth, trim={5cm 0 15cm 0}, clip]{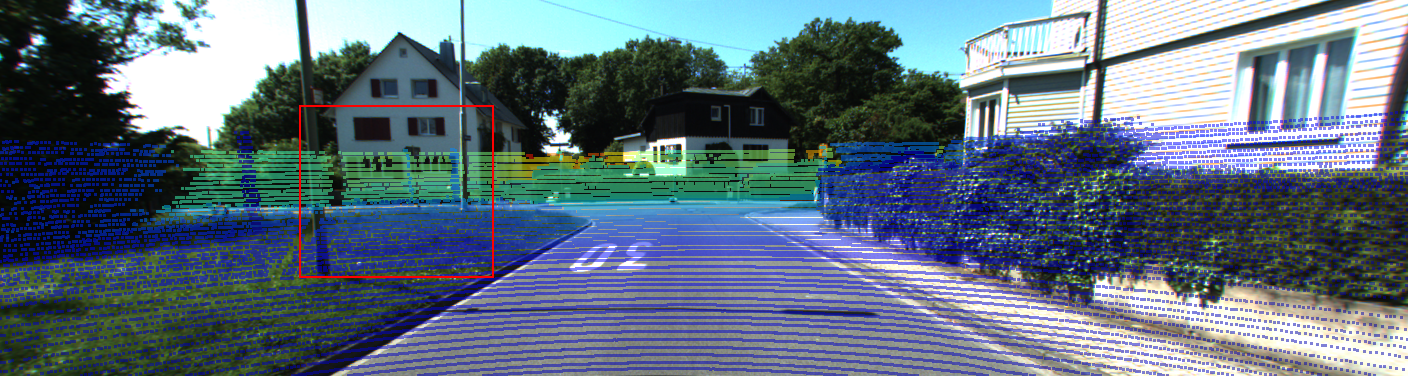}
     & \includegraphics[width=0.295\linewidth]{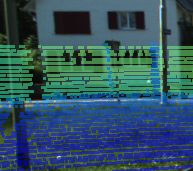}\\
     \rotatebox{90}{~~~~~~~~SOAC~(ours)} & \includegraphics[width=0.58\linewidth, trim={5cm 0 15cm 0}, clip]{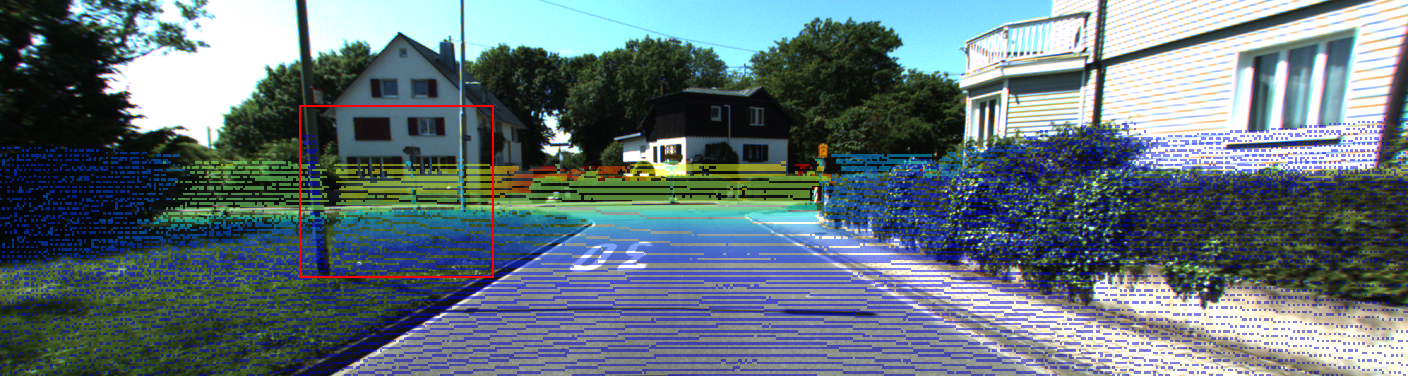}
     & \includegraphics[width=0.295\linewidth]{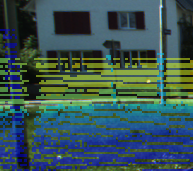}
\end{tabular}
\vspace{-2mm}
\caption{Qualitative LiDAR/front camera calibration results on KITTI-360~\cite{liao2022kitti} dataset.}
\label{fig:qualitative_kitti}
\vspace{-5mm}
\end{figure}

\begin{figure*}[!htbp]
\centering
\scriptsize
\setlength{\tabcolsep}{0.002\linewidth}
    \begin{tabular}{lccc}  
        \rotatebox{90}{~~~~~~~~~~~~GT} & 
        \includegraphics[width=0.31\linewidth, trim={0 2cm 0 7cm}, clip]{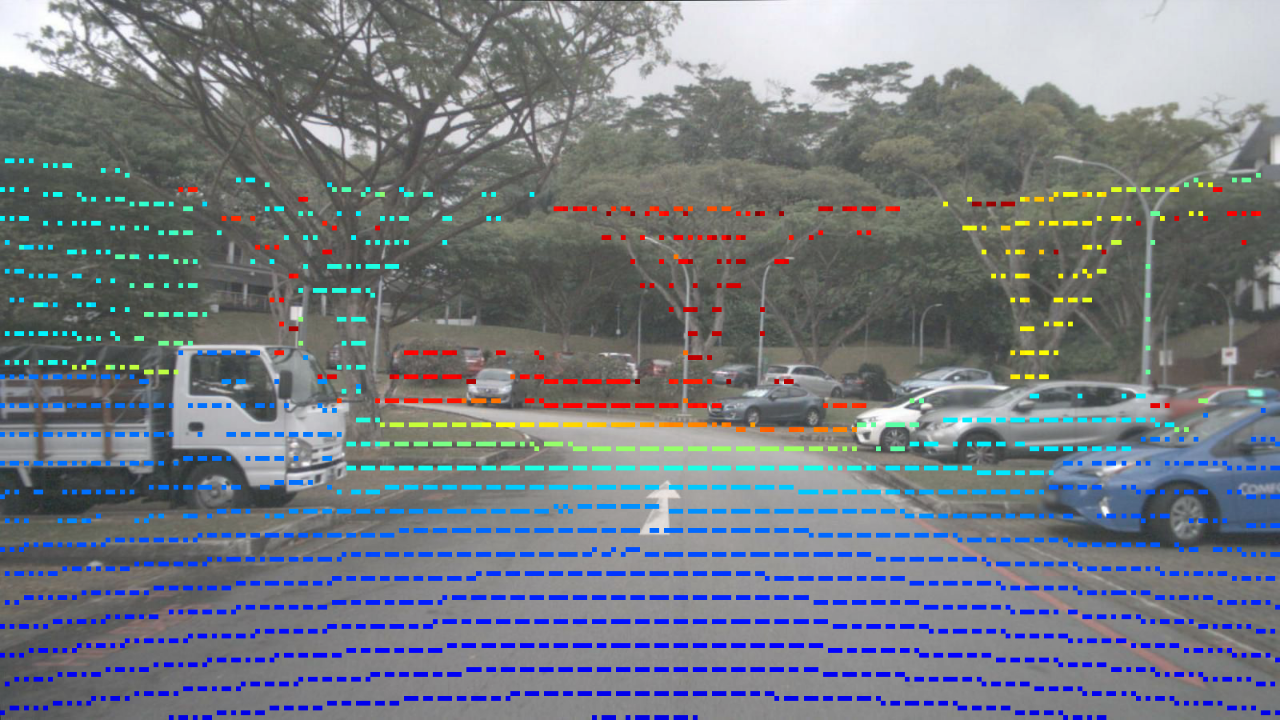} &
        \includegraphics[width=0.31\linewidth, trim={0 2cm 0 7cm}, clip]{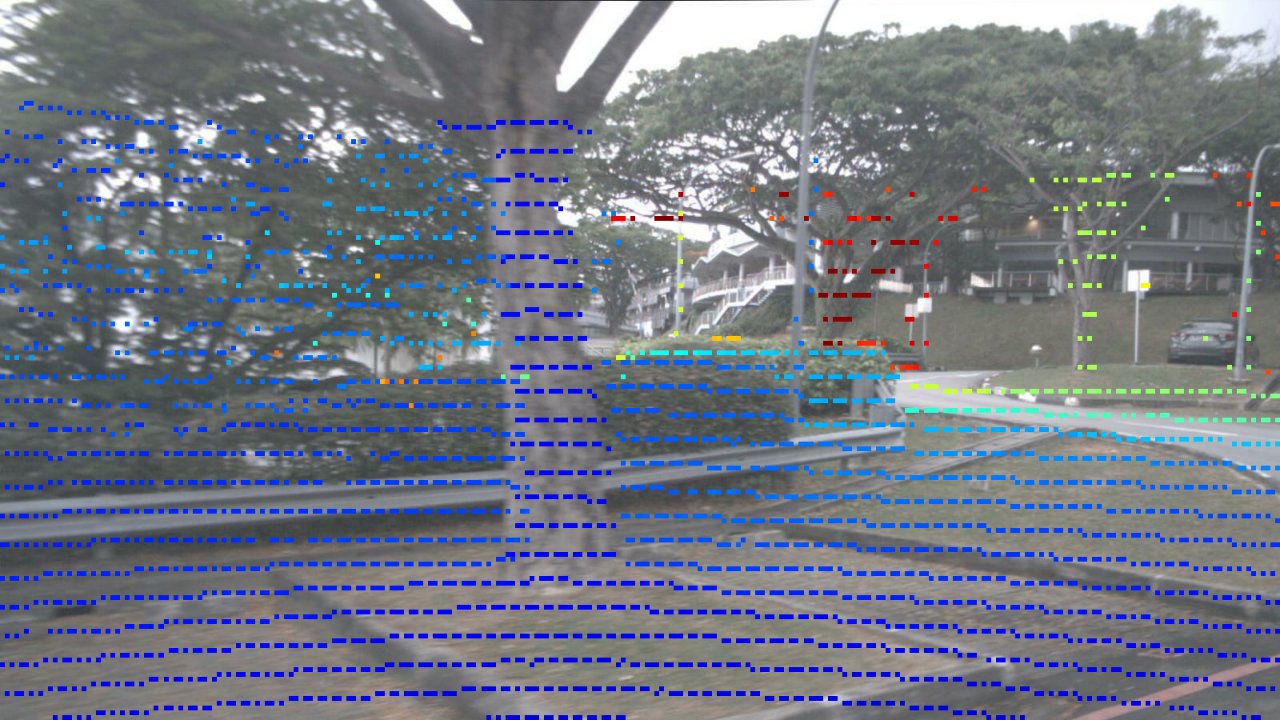} &
        \includegraphics[width=0.31\linewidth, trim={0 2cm 0 7cm}, clip]{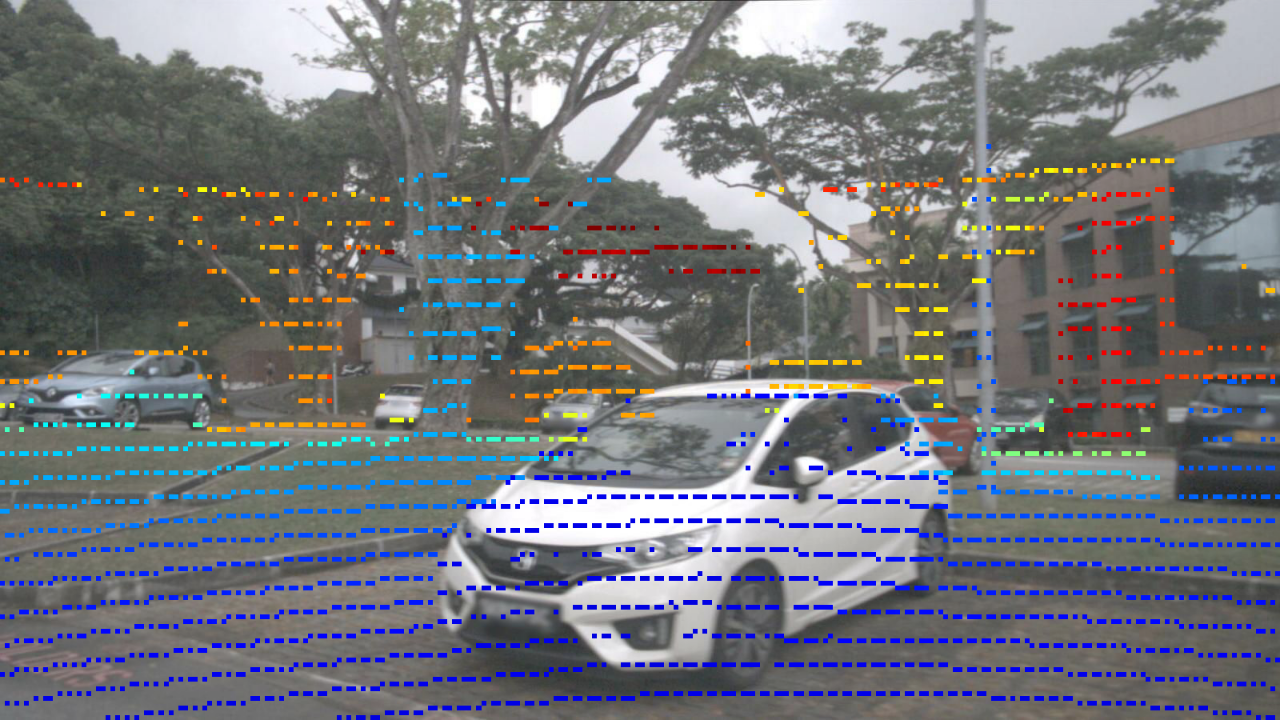}\\

        \rotatebox{90}{~~~~~~MOISST\cite{herau2023moisst}} &
        \includegraphics[width=0.31\linewidth, trim={0 2cm 0 7cm}, clip]{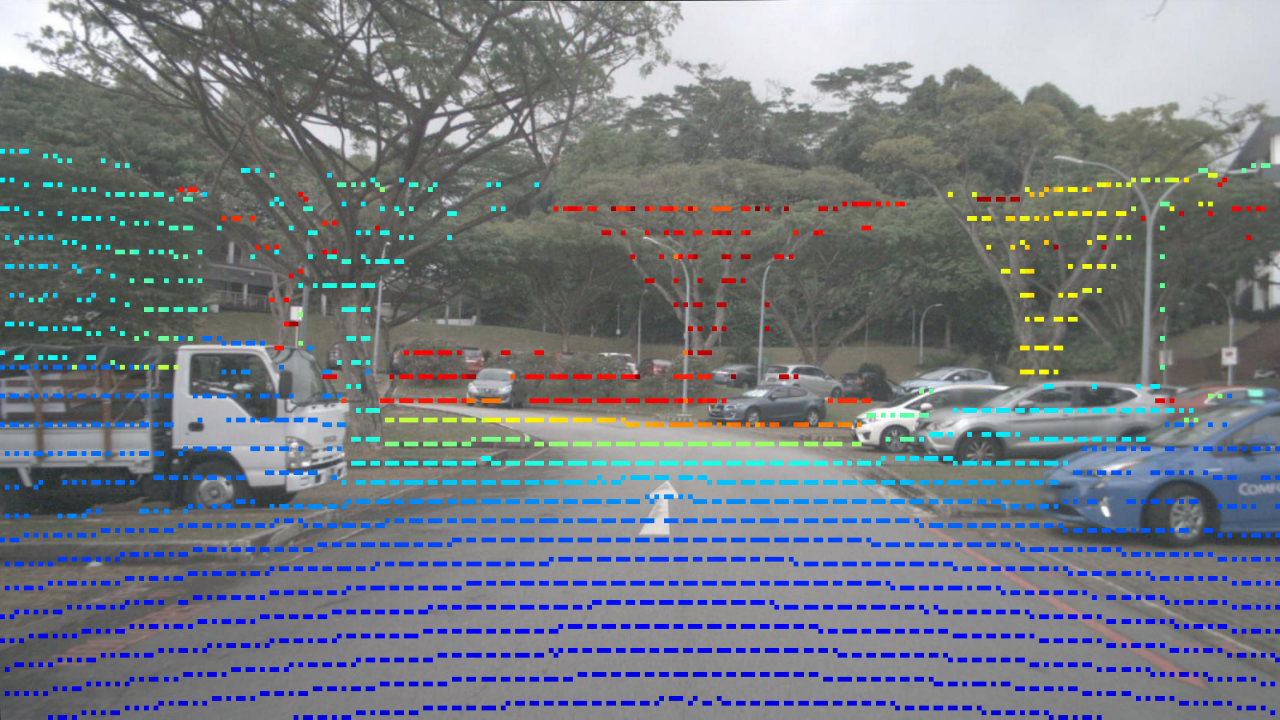} &
        \includegraphics[width=0.31\linewidth, trim={0 2cm 0 7cm}, clip]{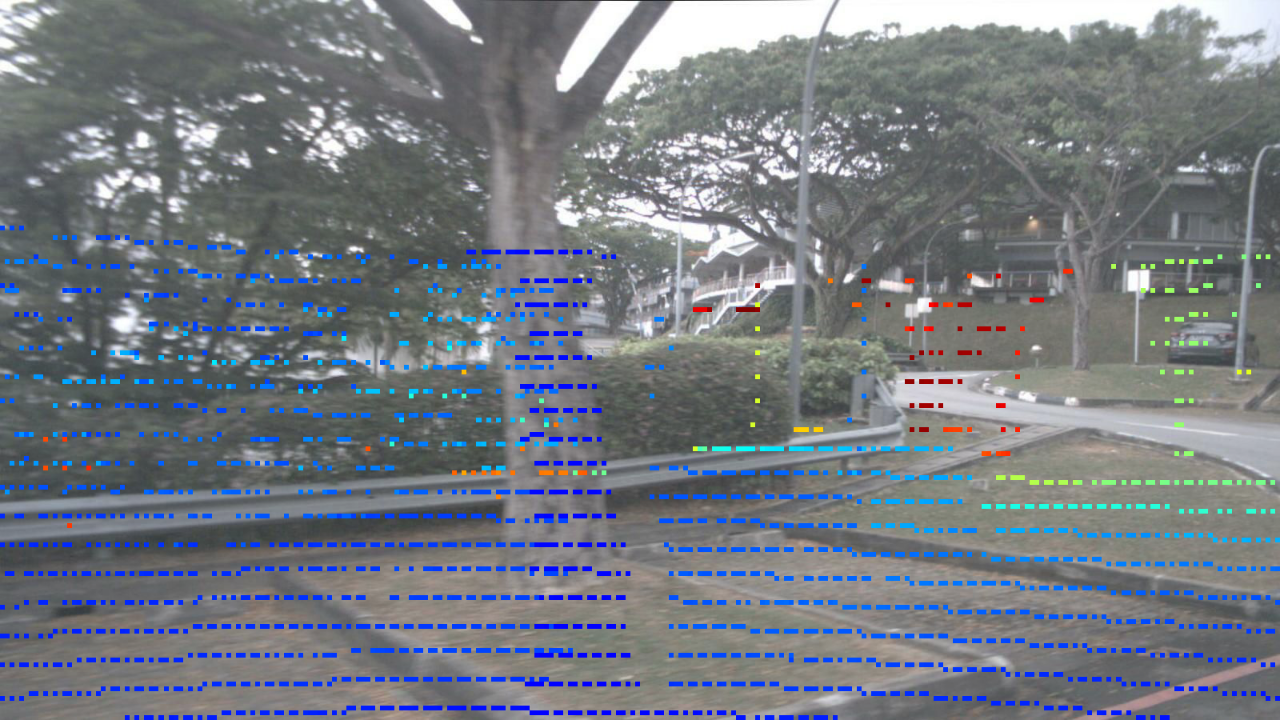} &
        \includegraphics[width=0.31\linewidth, trim={0 2cm 0 7cm}, clip]{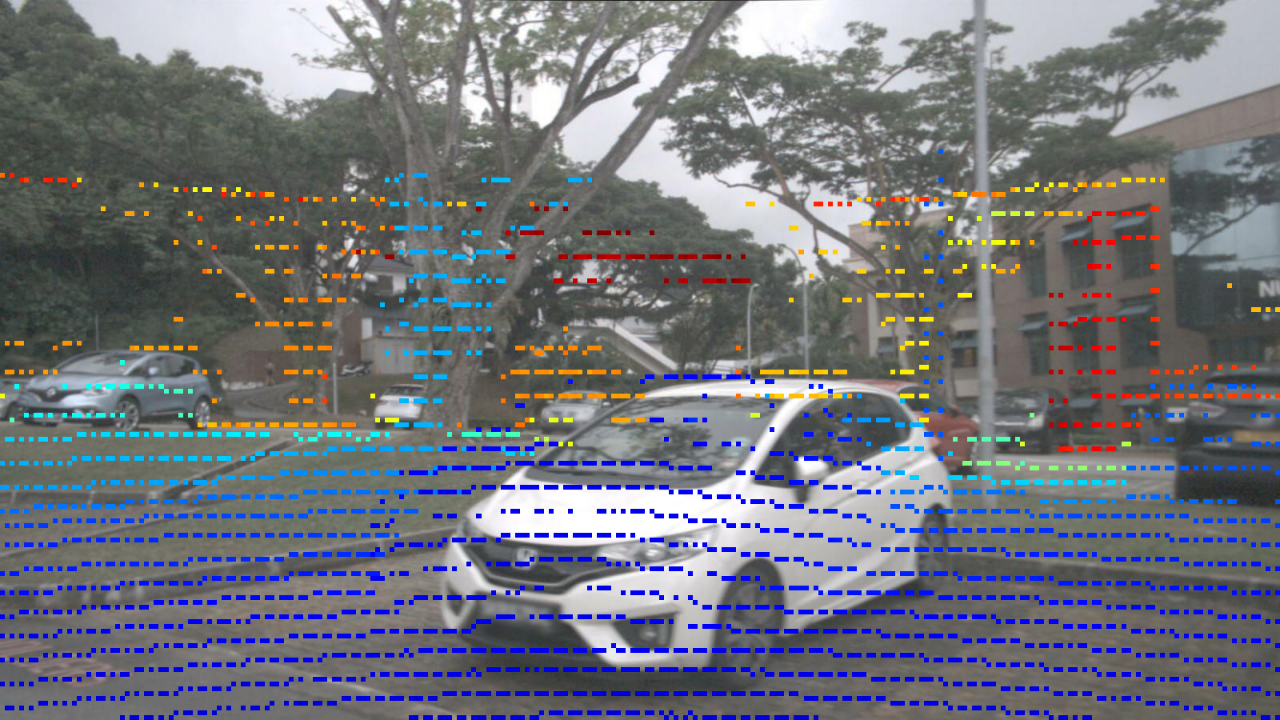}\\

        \rotatebox{90}{~~~~~~SOAC~(ours)} &
        \includegraphics[width=0.31\linewidth, trim={0 2cm 0 7cm}, clip]{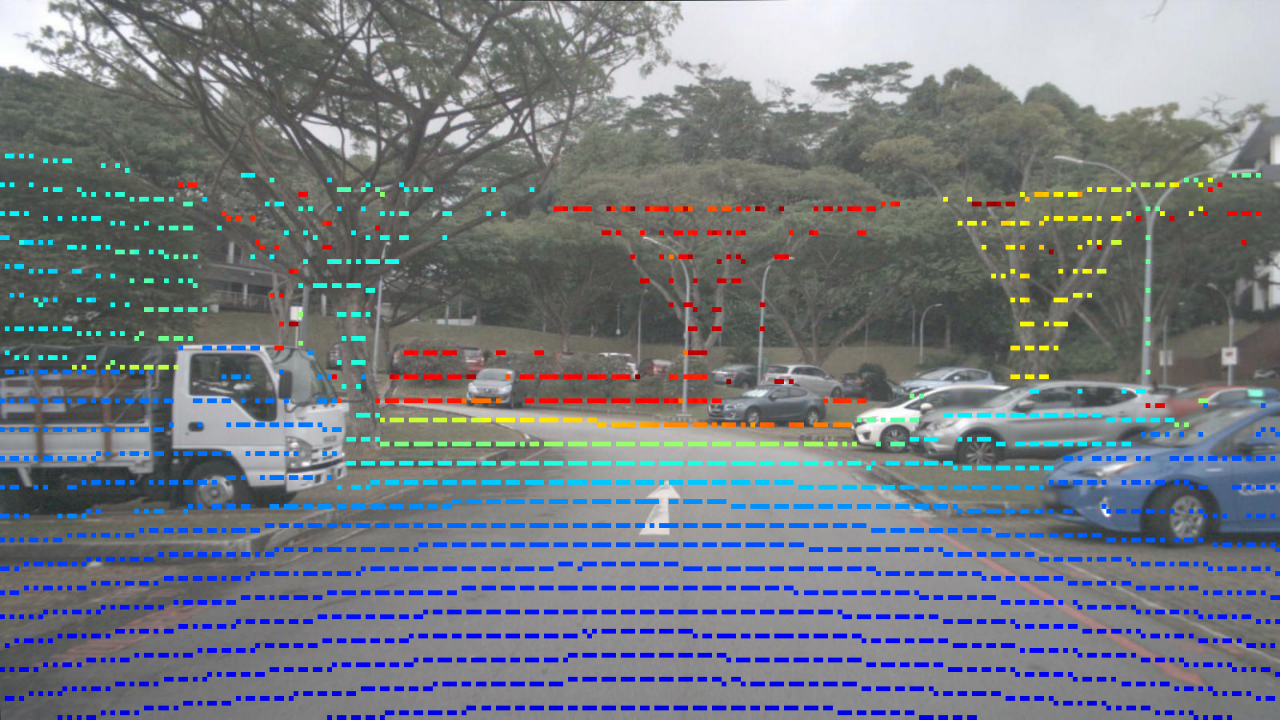} &
        \includegraphics[width=0.31\linewidth, trim={0 2cm 0 7cm}, clip]{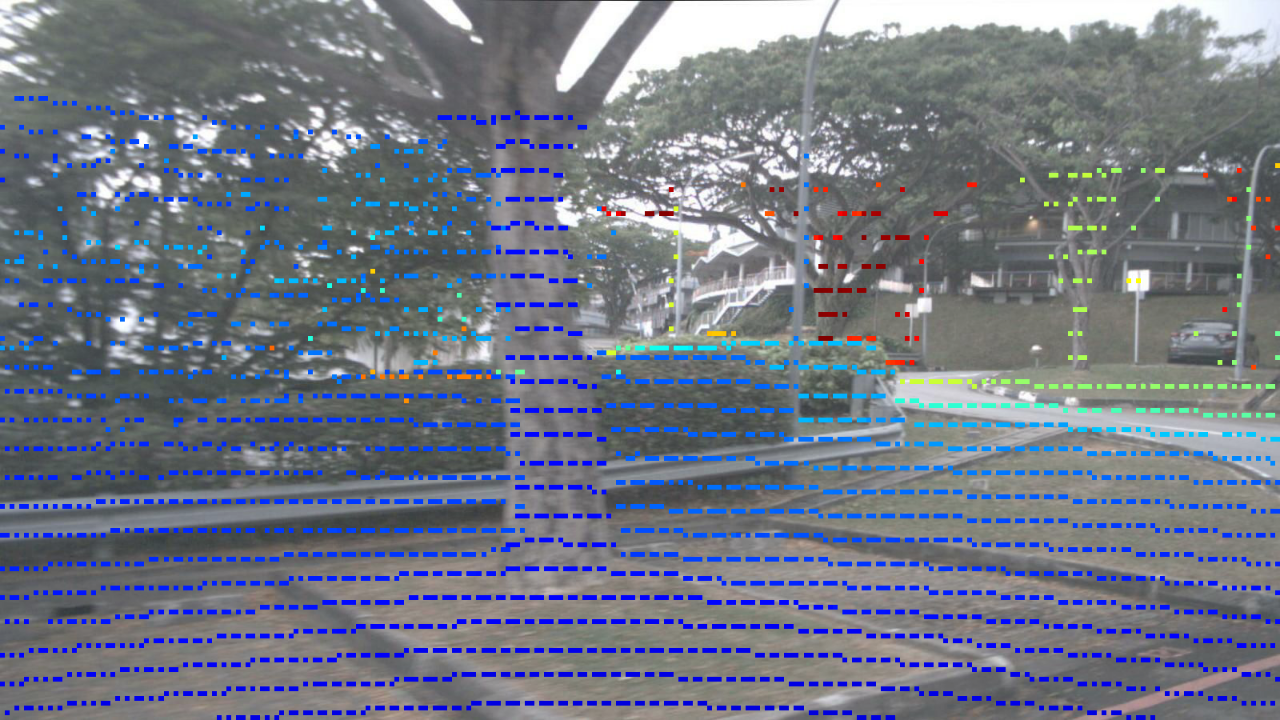} &
        \includegraphics[width=0.31\linewidth, trim={0 2cm 0 7cm}, clip]{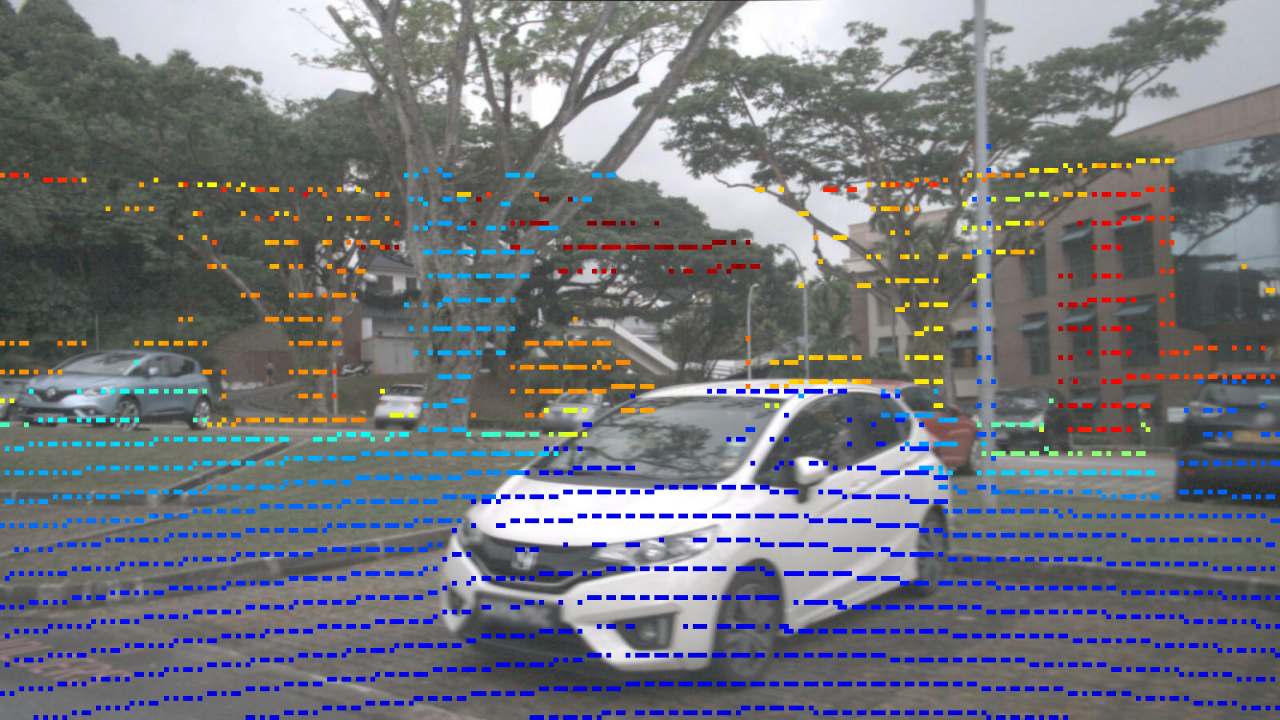} 

    \end{tabular}
\vspace{-3mm}
\caption{Qualitative LiDAR/Camera reprojection results on nuScenes~\cite{caesar2020nuscenes} dataset.}
\label{fig:reproj_nuscenes}
\vspace{-1mm}
\end{figure*}

\paragraph{Qualitative results.}
We show the reprojection of the LiDAR on the images using the calibration obtained from different methods. On KITTI-360 (cf. Fig.~\ref{fig:qualitative_kitti}) we can see that LCCNet does not provide a satisfying result and that SOAC is able to provide a visually comparable alignment to the ground truth calibration. On nuScenes (cf. Fig.~\ref{fig:reproj_nuscenes}), the calibration from SOAC provides a better alignment than MOISST, assessing the quantitative results of Fig.~\ref{fig:results}. More qualitative results are given in the supplementary, along with ablation studies on visibility grids (cf. Sec.~\ref{sec/method/filtergrid}) and correction bounding (cf. Sec.~\ref{sec/method/details}).

\subsection{Limitations} \label{sec/experiments/limits}

\begin{table}[t!]
    \centering
    \resizebox{1.0\columnwidth}{!}
    {
        \renewcommand{\arraystretch}{1.3}
        \setlength{\tabcolsep}{0.025\linewidth}
        \begin{tabular}{@{}l l c c c c c c@{}}
        \multicolumn{2}{c}{Errors} & Cam Front & Cam Left & Cam Right & LiDAR \\
        \toprule
        \multirow{2}{*}{Translation (cm)} & Extrinsic & $47.9$ & $67.8$ & $70.4$ & $50.9$ \\
        & Poses & $3.5$ & $2.6$ & $26.2$ & $19.1$ \\
        \hline
        \multirow{2}{*}{Rotation (°)} & Extrinsic & $0.13$ & $0.18$ & $0.23$ & $0.58$ \\
        & Poses  & $1.63$ & $1.12$ & $1.35$ & $0.60$ \\
        \hline
        \multicolumn{2}{c}{Time offset (ms)} & $39.18$ & $58.16$ & $40.74$ & $39.38$ \\
        \bottomrule
        \end{tabular}
    }
    \caption{\label{tab:compensation} \textbf{SOAC space-time compensation on a sequence from KITTI-360~\cite{liao2022kitti}.} Mean absolute poses of sensors are correct whereas the spatio-temporal calibration computed by the method is erroneous.}
    \vspace{-0.35cm}
\end{table}

\noindent\textbf{Time-space compensation.}
When simultaneously calibrating spatially and temporally, there are cases where the disentanglement is impossible. In a sequence where the vehicle is driving in a straight line at a constant speed, there is an infinite number of solutions that can provide the correct poses. In Tab.~\ref{tab:compensation}, we show the calibration results on a straight line with constant speed from KITTI-360. We can see the pose error is fairly low, but the extrinsic and temporal calibration is incorrect. This means that there is a need to select scenes with speed variation in order to reduce to a single possible solution. As most Pandaset sequences are in a straight line at a constant speed, we decided to not do temporal calibration on them.

\noindent\textbf{Scene structure.}
When the scenes are more open and/or larger, the projected rays will have to travel a longer distance before reaching the scene's structures.
Considering LiDAR to camera calibration, the rotation error has a linearly increasing impact according to the ray distance when reprojected to the camera frame, while the translation error's impact is independent of the ray distance. Thus, we tend to lose precision on the translation as the ray gets longer.
When analyzing the LiDAR rays length distribution of the datasets in Fig.~\ref{fig:lidar_length}, we observe that the LiDAR rays on Pandaset are longer, meaning that the scenes are larger and open, and the structures are farther than on KITTI-360 and nuScenes.
This explains most likely the decrease in calibration performances for the LiDAR extrinsic translation parameters on Pandaset (median error of 18.1 cm) compared to KITTI-360 (median error of 8.2 cm) or nuScenes (median error of 5.1 cm). 

\noindent\textbf{Training time.}
As we train one NeRF per camera, and register all the other sensors on each NeRF, the training time increases exponentially with the number of cameras. For instance, on nuScenes one epoch takes approximately 1 minute 45 seconds for 3 cameras and 8 minutes for 5 cameras using the same GPU. This reduces the scalability of our method, but this phenomenon is mitigated by the fact that we use much smaller images than MOISST to reach better performance. We refer to the supplementary for more in-depth details on the efficiency of our method wrt. image size compared to MOISST. 

\section{Conclusion}

In this paper, we presented SOAC, a targetless and self-supervised method for spatial and temporal calibration. This approach is able to simultaneously calibrate multiple sensors of different modalities, by leveraging the use of multiple camera-specific implicit scene representations, and taking into account the overlap between the sensors. Our approach is fully automatic by relying on gradient descent for the optimization process, and surpasses similar methods previously introduced.
The reliance on a reference sensor with known trajectory, and the need of near structures for a precise calibration, are restrictions that could open to future research to alleviate them.

{
    \small
    \bibliographystyle{ieeenat_fullname}
    \bibliography{cvpr_biblio}
}
\newpage
\label{sec/supplementary}
\appendix
\section{Technical Details} 
\subsection{Datasets}

\paragraph{KITTI-360~\cite{liao2022kitti}:} Sequences are selected and cropped by considering vehicle speed variations to remove time-space compensation issues as described in main article Sec.~\textcolor{red}{4.3}. Once sequences are cropped, one out of two frames are kept for all sensors to obtain a total of 40 frames per sequence. This decision was made to match the same length as the NVS benchmark sequences present on the dataset. The details from each sequence are summarized in Tab.~\ref{tab:itti_sequences}.

\begin{table}[hbt!]
    \centering
    \resizebox{0.95\columnwidth}{!}
    {
        \renewcommand{\arraystretch}{1.0}
        \setlength{\tabcolsep}{0.028\linewidth}
        \begin{tabular}{@{}c c c c c @{}}
        Sequence & KITTI-360 run & Starting frame & Ending frame &\\
        \toprule
        1 & 0009 & 980 & 1058\\
        2 & 0009 & 2854 & 2932\\
        3 & 0010 & 3390 & 3468\\
        4 & 0002 & 4722 & 4800\\
        Straight line & 0009 & 220 & 298\\
        \bottomrule
        \end{tabular}
    }
    \caption{\label{tab:itti_sequences} Selected frames for each KITTI-360~\cite{liao2022kitti} sequence.}
    \vspace{-0.45cm}
\end{table}

\paragraph{nuScenes~\cite{caesar2020nuscenes}:} Since nuScnes poses are provided only in $\mathrm{SE(2)}$, they cannot be used directly for our method. Instead, we use KISS-ICP~\cite{vizzo2023kiss} to get a good estimate of the LiDAR poses. Extrinsic calibration provided by the dataset is then used to obtain the poses for all cameras. We select the sequences 916, 410 and 417 for our experiments, as they are more suitable for the calibration (closer structures, more speed variation). All LiDAR scans are used during calibration as the LiDAR is sparser than the one in KITTI-360, while one out of two images is subsampled to reduce training time.

\paragraph{Pandaset~\cite{xiao2021pandaset}:} Since extrinsic parameters are not provided by the dataset, they are estimated using the global poses of all sensors at several frames by calculating the transformation between the frames with the same timestamp from each sensor. Sequences 33, 40 and 53 are used for our experiments as they have more close structures. We apply the same subsampling strategy as for nuScenes.

\subsection{Architecture and Losses}

For our NeRF network architecture, we use the same model as MOISST~\cite{herau2023moisst} which is inspired by the \texttt{nerfacto} model of Nerfstudio\footnote{https://docs.nerf.studio/en/latest/nerfology/methods/nerfacto.html} open source project.
It uses the combination of two papers. The first one is the proposal network from MipNeRF-360~\cite{barron2022mip} with two proposal networks for the coarse density estimation and a final NeRF for the radiance and the fine density, improving the geometry of the scene, the rendering quality and reducing the training time. The second one is the hash grid introduced by instant-NGP~\cite{muller2022instant} to replace the deterministic positional encoding, which also accelerates the training.
Following the nerfacto implementation, 128 points (instead of 256) per ray are sampled for the first proposal model, 96 points for the second one, and 48 points for the final NeRF model, which outputs our results.

On top of $\mathcal{L}_{C}$, $\mathcal{L}_{Cam}$ and $\mathcal{L}_{D}$, two losses for geometric consistency, also used by MOISST, are added: a structural dissimilarity (DSSIM) loss  $\mathcal{L}_{SSIM}$~\cite{ssim}, and a depth smoothness loss $\mathcal{L}_{DS}$ from RegNeRF~\cite{niemeyer2022regnerf}.

\begin{figure*}[!htbp]
\centering
\begin{tabular}{lll}
 \includegraphics[height=.27\linewidth]{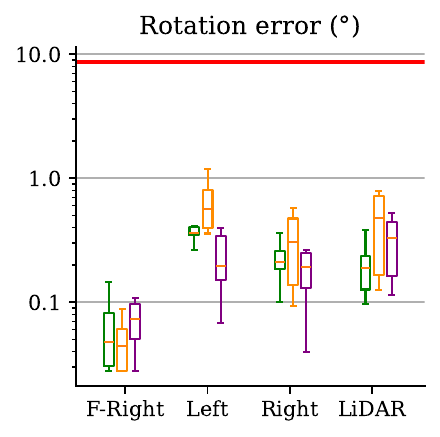}&  \includegraphics[height=.27\linewidth]{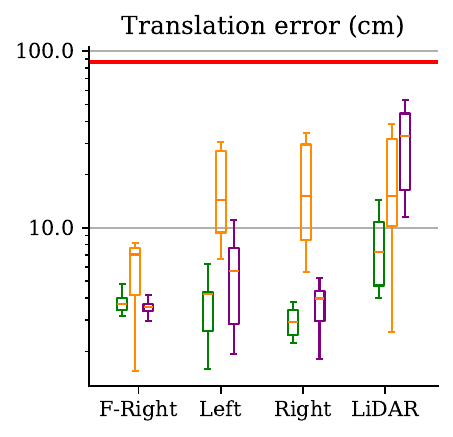}& \includegraphics[height=.27\linewidth]{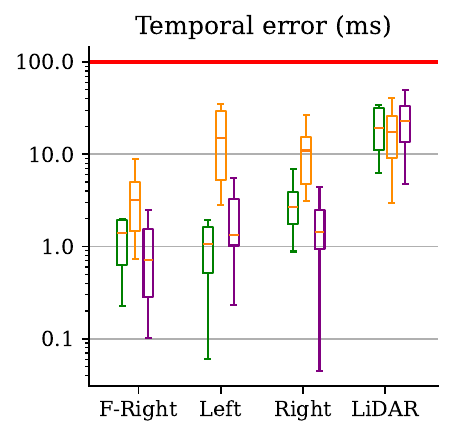}
\end{tabular}
\caption{Ablation results on KITTI-360~\cite{liao2022kitti} sequence 4: for \textcolor{ForestGreen}{SOAC}, \textcolor{orange}{SOAC w/o Sigmoid} and \textcolor{violet}{SOAC w/o visibility grid} as box plots with log scale, the \textcolor{red}{red lines} show the initial error (Best viewed in color).}
\label{fig:ablation}
\end{figure*}

\subsection{Hyperparameters}

\begin{table}[hbt!]
    \centering
    \resizebox{0.6\columnwidth}{!}
    {
        \renewcommand{\arraystretch}{1.0}
        \setlength{\tabcolsep}{0.028\linewidth}
        \begin{tabular}{@{}l c c c c @{}}
        Hyperparameter & Value &\\
        \toprule
        Number of epochs & 20 \\
        Initial calibration lr & 1e-3 \\
        Final calibration lr & 1e-4 \\
        Visibility grid size & 20 \\
        Batch size & 100 \\
        Patch size & [15, 15]\\
        $\mathcal{L}_{C}$ coef & 1\\
        $\mathcal{L}_{Cam}$ coef & 1\\
        $\mathcal{L}_{SSIM}$ coef & 0.1\\
        $\mathcal{L}_{D}$ coef & 1\\
        $\mathcal{L}_{DS}$ coef & 1e-4\\
        Translation bounding & 2 meters\\
        Temporal bounding & 500 ms\\
        \bottomrule
        \end{tabular}
    }
    \caption{\label{tab:hyperparameters} SOAC hyperparameters used for the training.}
\end{table}

\begin{table}[hbt!]
    \centering
    \resizebox{1.0\columnwidth}{!}
    {
        \renewcommand{\arraystretch}{1.0}
        \setlength{\tabcolsep}{0.028\linewidth}
        \begin{tabular}{@{}l c c c c @{}}
        Sensor & KITTI-360~\cite{liao2022kitti} & nuScenes~\cite{caesar2020nuscenes} & Pandaset~\cite{xiao2021pandaset} &\\
        \toprule
        Diagonal cams& - & 1 & 3 \\
        Side cams& 1 & 9 & - \\
        LiDAR& 6 & 5 & 8 \\
        
        \bottomrule
        \end{tabular}
    }
    \caption{\label{tab:hyperparameters} SOAC hyperparameters used for the training.}
     \vspace{-0.45cm}
\end{table}

In Tab.~\ref{tab:hyperparameters} are indicated the hyperparameters used for the training of SOAC, and in Tab.~\ref{tab:dataset_hyperparameters} are the NeRF delaying epochs depending on the dataset. Delaying the NeRF proves advantageous in scenarios characterized by a multitude of sensors, some of which exhibit minimal overlap with the reference sensor throughout the sequence. This approach facilitates the accurate propagation of calibration information during training from sensors presenting significant overlap with the reference sensor to those with lesser or no overlap at all. Basically, larger overlaps and larger quantities of data reduce the number of necessary delay epochs.
The number of epochs for training MOISST is reduced to 20, as improvement was not observed with more. 
The spatial and temporal optimization learning rate is fine-tuned to 5e-4.

\section{Additional ablations}
\paragraph{Correction bounding.}
The addition of the sigmoid for bounding the translation and temporal corrections allows better stability and robustness as shown in Fig.~\ref{fig:ablation} on which a huge decrease in calibration accuracy can be noticed when removing the sigmoid.
\paragraph{Visibility grid.}
Removing the visibility grids deteriorates the performance of the LiDAR calibration rotation and translation as shown in Fig.~\ref{fig:ablation}. 

\begin{figure*}[!htbp]
\centering
\setlength{\tabcolsep}{0.001\linewidth}
    \begin{tabular}{cccc}

        \small{Sequence 1} & \small{Sequence 2} & \small{Sequence 3} & \small{Sequence 4}\\

         \includegraphics[height=.21\linewidth]{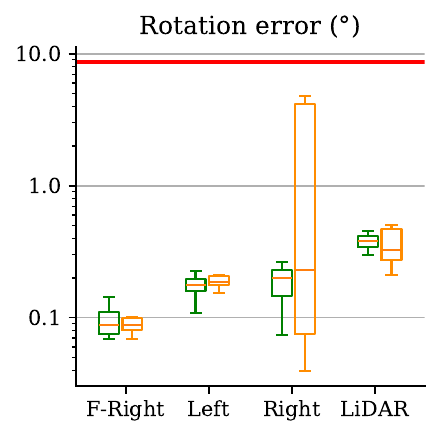} & 
         \includegraphics[height=.21\linewidth]{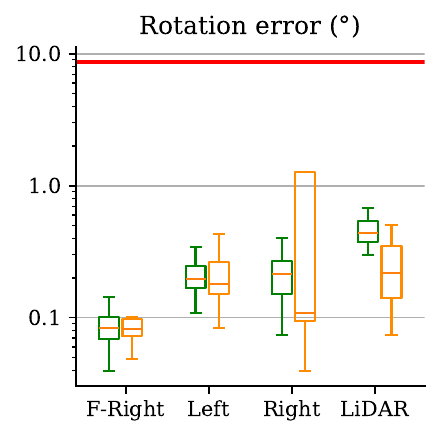} &
         \includegraphics[height=.21\linewidth]{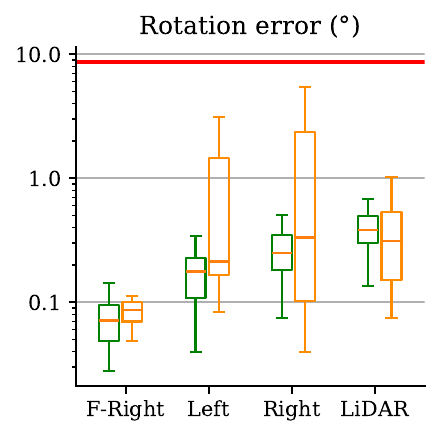} &
         \includegraphics[height=.21\linewidth]{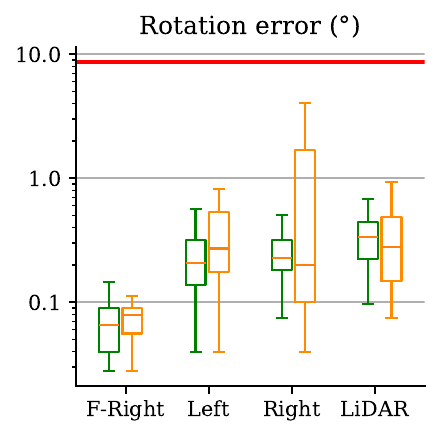} \\ 
         \includegraphics[height=.21\linewidth]{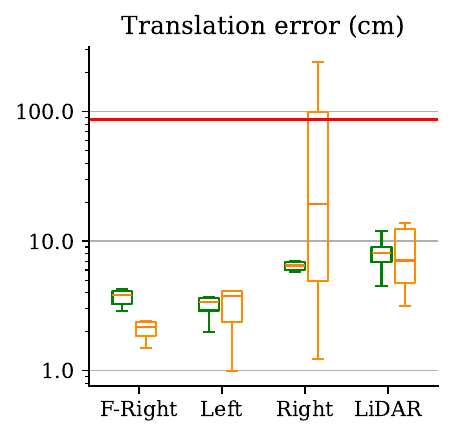} & 
         \includegraphics[height=.21\linewidth]{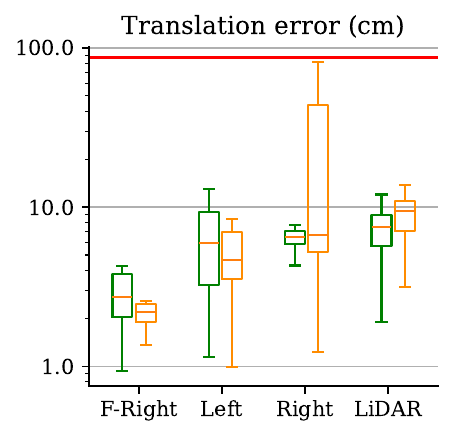} &
         \includegraphics[height=.21\linewidth]{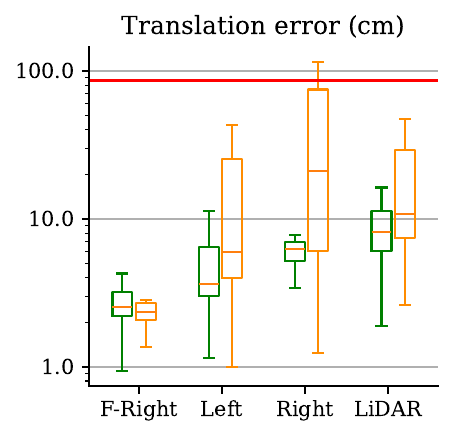} &
         \includegraphics[height=.21\linewidth]{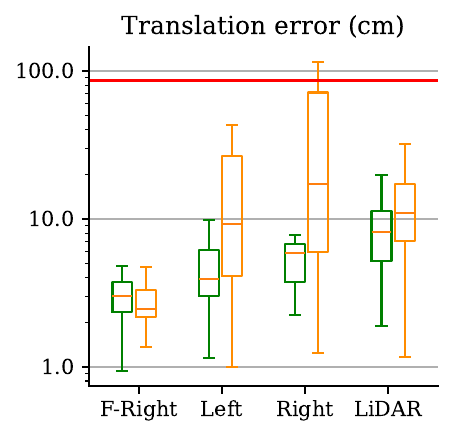} \\
         \includegraphics[height=.21\linewidth]{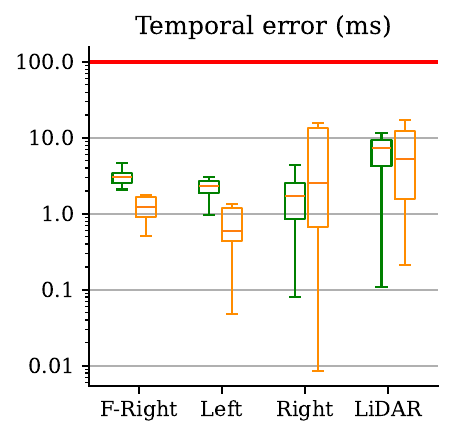} & 
         \includegraphics[height=.21\linewidth]{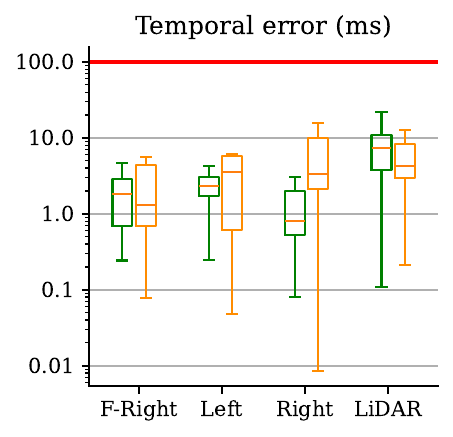} &
         \includegraphics[height=.21\linewidth]{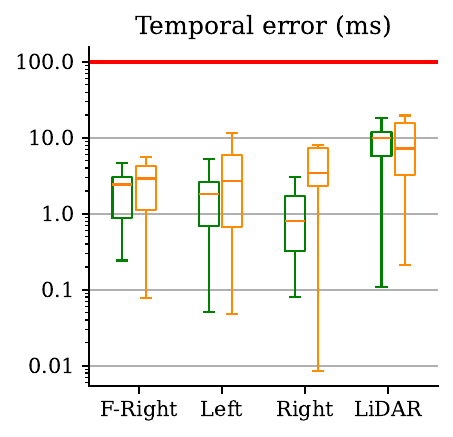} &
         \includegraphics[height=.21\linewidth]{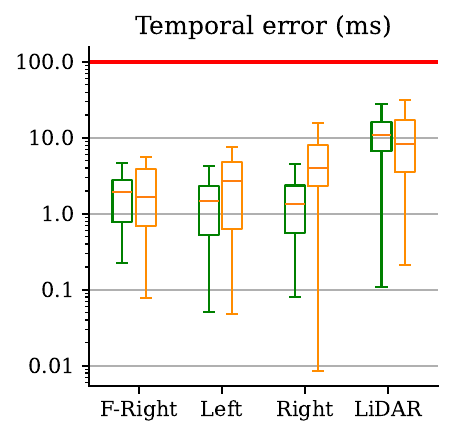} 
    \end{tabular}
\caption{Results for KITTI-360~\cite{liao2022kitti} per sequence for \textcolor{ForestGreen}{SOAC} and \textcolor{orange}{MOISST}~\cite{herau2023moisst} as box plots with log scale. The \textcolor{red}{red lines} show the initial error (best viewed in color).}
\label{fig:kitti_per_seq}
\end{figure*}

\begin{figure*}[!htbp]
\centering
\setlength{\tabcolsep}{0.001\linewidth}
    \begin{tabular}{ccc}

        \small{Sequence 33} & \small{Sequence 40} & \small{Sequence 53}\\

         \includegraphics[height=.21\linewidth]{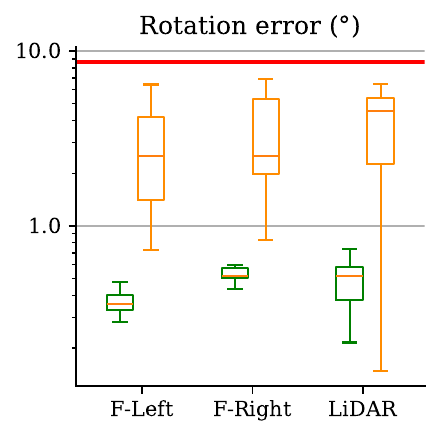} & 
         \includegraphics[height=.21\linewidth]{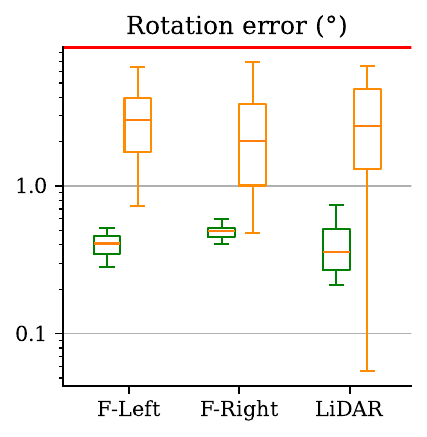} &
         \includegraphics[height=.21\linewidth]{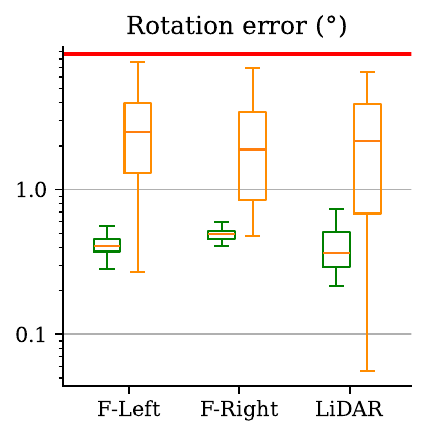}\\ 
         \includegraphics[height=.21\linewidth]{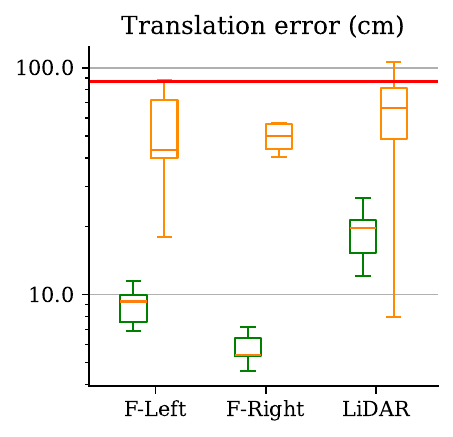} & 
         \includegraphics[height=.21\linewidth]{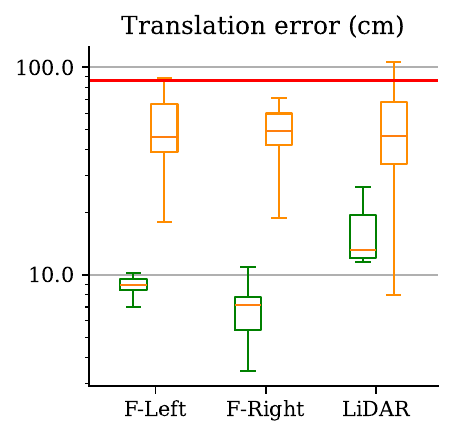} &
         \includegraphics[height=.21\linewidth]{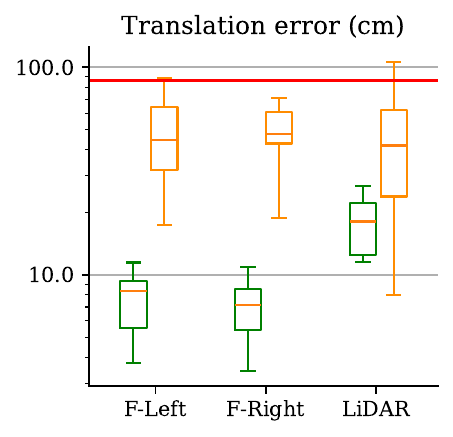}
    \end{tabular}
\caption{Results for Pandaset~\cite{pandey2012automatic} per sequence for \textcolor{ForestGreen}{SOAC} and \textcolor{orange}{MOISST}~\cite{herau2023moisst} as box plots with log scale. The \textcolor{red}{red lines} show the initial error (best viewed in color).}
\label{fig:Pandaset_per_seq}
\end{figure*}

\begin{figure*}[!htbp]
\centering
\setlength{\tabcolsep}{0.001\linewidth}
    \begin{tabular}{ccc}

        \small{Sequence 916} & \small{Sequence 410} & \small{Sequence 417}\\

         \includegraphics[height=.21\linewidth]{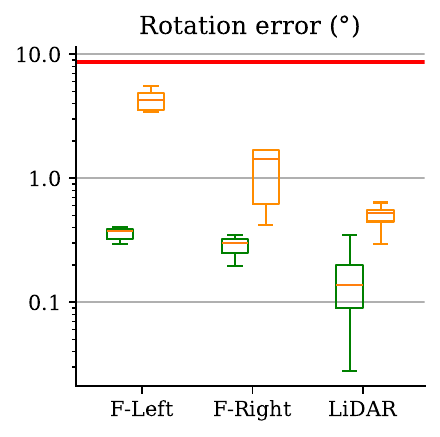} & 
         \includegraphics[height=.21\linewidth]{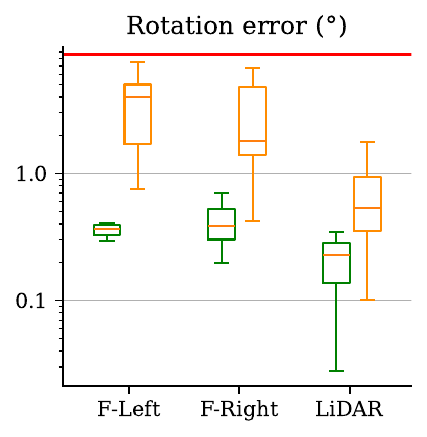} &
         \includegraphics[height=.21\linewidth]{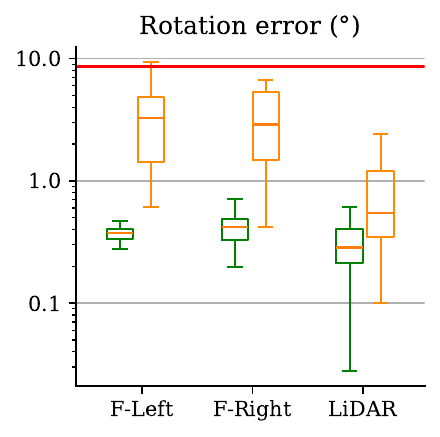}\\ 
         \includegraphics[height=.21\linewidth]{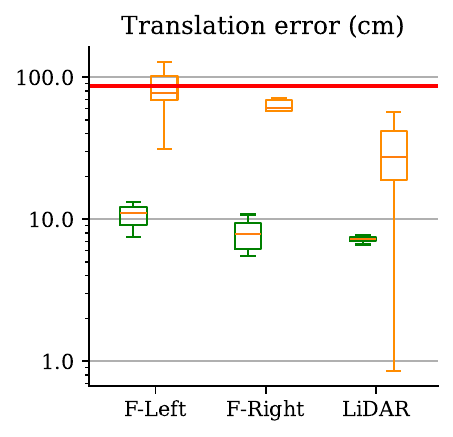} & 
         \includegraphics[height=.21\linewidth]{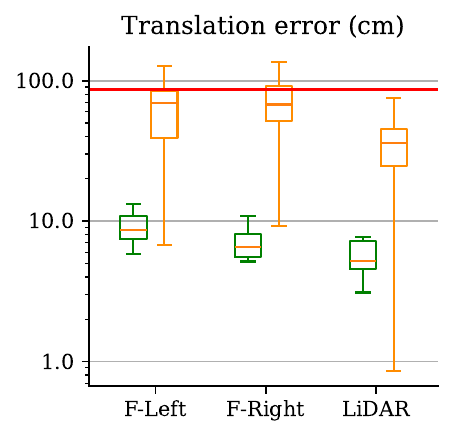} &
         \includegraphics[height=.21\linewidth]{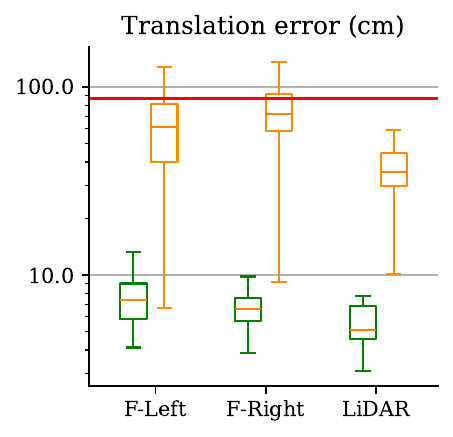}\\
         \includegraphics[height=.21\linewidth]{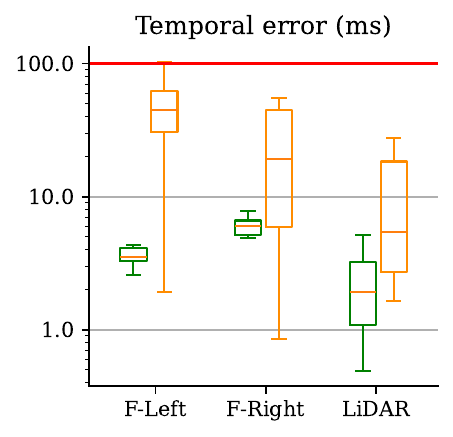} & 
         \includegraphics[height=.21\linewidth]{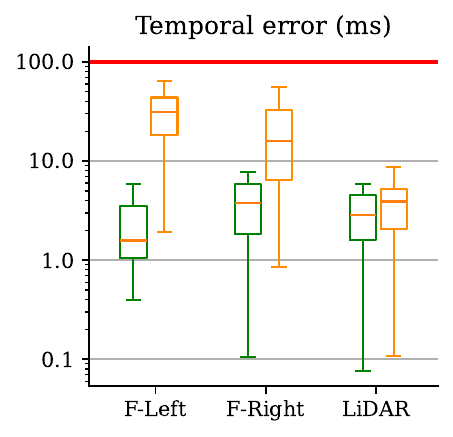} &
         \includegraphics[height=.21\linewidth]{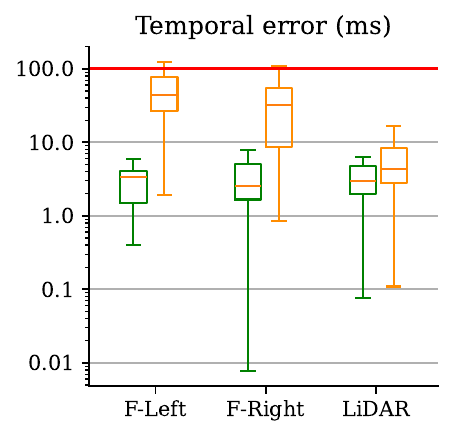}
    \end{tabular}
\caption{Results for nuScenes~\cite{caesar2020nuscenes} per sequence for \textcolor{ForestGreen}{SOAC} and \textcolor{orange}{MOISST}~\cite{herau2023moisst} as box plots with log scale. The \textcolor{red}{red lines} show the initial error (best viewed in color).}
\label{fig:nuscenes_per_seq}
\end{figure*}

\section{Training time}

\begin{table}[t!]
    \centering
    \resizebox{0.9\columnwidth}{!}
    {
        \renewcommand{\arraystretch}{1.0}
        \setlength{\tabcolsep}{0.028\linewidth}
        \begin{tabular}{@{}l c c c c c @{}}
        Dataset & MOISST~\cite{herau2023moisst} & SOAC \\
        \toprule
        KITTI-360~\cite{liao2022kitti} & $\sim$ 2 h 30 min & $\sim$ 1 h 30 min \\
        Nuscenes~\cite{caesar2020nuscenes} (3 cams) & $\sim$ 2 h 30 min & $\sim$ 1 h \\
        Nuscenes~\cite{caesar2020nuscenes} (5 cams) & $\sim$ 4 h 30 min & $\sim$ 2 h 30 min \\
        Pandaset~\cite{xiao2021pandaset} & $\sim$ 1 h 30 min & $\sim$ 1 h 20 min \\
        \bottomrule
        \end{tabular}
    }
    \caption{\label{tab:training_time} Training time comparison on different sequences.}
\end{table}

\begin{table}[t!]
    \centering
    \resizebox{1.0\columnwidth}{!}
    {
        \renewcommand{\arraystretch}{1.0}
        \setlength{\tabcolsep}{0.02\linewidth}
        \begin{tabular}{cc | cc | cc}
            \multicolumn{2}{c|}{~~Downscale} & \multicolumn{2}{c|}{Calibration error(°/cm/ms)} & \multicolumn{2}{c}{Training time (min)} \\
            &factor & SOAC & MOISST~\cite{herau2023moisst} & SOAC & MOISST~\cite{herau2023moisst}\\
            \midrule
             &1 & 0.2/4.6/3.9 & 0.1/5.3/1.3   & 605 & 163\\
             &2 & 0.3/4.6/2.5 & 0.3/24.1/5.3   & 181 & 42\\
             &4 & 0.2/4.6/1.7 & 1.1/41.4/13.1  & 85 & 17\\
             &8 &  0.4/12.3/8.2  & 2.6/56.6/28.6 & 53 & 12\\
        
        \end{tabular}
    }
    \caption{\label{tab:downscale_perf} Training time and calibration accuracy for varying downscale factor on KITTI-360~\cite{liao2022kitti} sequence 1 seed 0}
\end{table}

We report the mean training times with both SOAC and MOISST for the sequences from each tested dataset in Tab.~\ref{tab:training_time}. For all the experiments, we used a GPU of similar performance to an RTX 3090.
The shown results are with the downscaled images as described in the paper. SOAC is able to provide better calibration than MOISST with shorter training time, even if multiple NeRFs are used, as it can use much smaller images.
To measure the impact of the image downscale factor in relation to each method's training time, we train both methods at different downscale factors and report results on Tab.~\ref{tab:downscale_perf}. As it can be observed, MOISST accuracy is considerably harmed by using lower-resolution images in comparison to SOAC. 
Furthermore, SOAC achieves high accuracy even with large downscale factors on the images (i.e. downscaling the image resolution by 4 shows no drop in accuracy for SOAC while being 8 times faster. In comparison, MOISST presents a severe drop in performance when downscaling). This enables SOAC to achieve more efficient training times given its ability to exploit lower-resolution images.

\section{Quantitative results}
Specific box plot results are provided for each sequence. The results for KITTI-360 are in Fig.~\ref{fig:kitti_per_seq}, the results for Nuscenes in Fig.~\ref{fig:nuscenes_per_seq}, and the results for Pandaset in Fig.~\ref{fig:Pandaset_per_seq}.

On KITTI-360, MOISST seems to provide results on par with SOAC on the Front-right camera and the LiDAR. However, on the side cameras, there is a significant difference in the stability of the calibration.
On Nuscenes and Pandaset, SOAC is much more precise and stable than MOISST all across the board.

\section{Qualitative results}
\begin{figure*}[!htbp]
\centering
\scriptsize
\setlength{\tabcolsep}{0.002\linewidth}
    \begin{tabular}{lccc}  
        \rotatebox{90}{~~~~~~~~~~~~GT} & 
        \includegraphics[width=0.31\linewidth, trim={0 2cm 0 7cm}, clip]{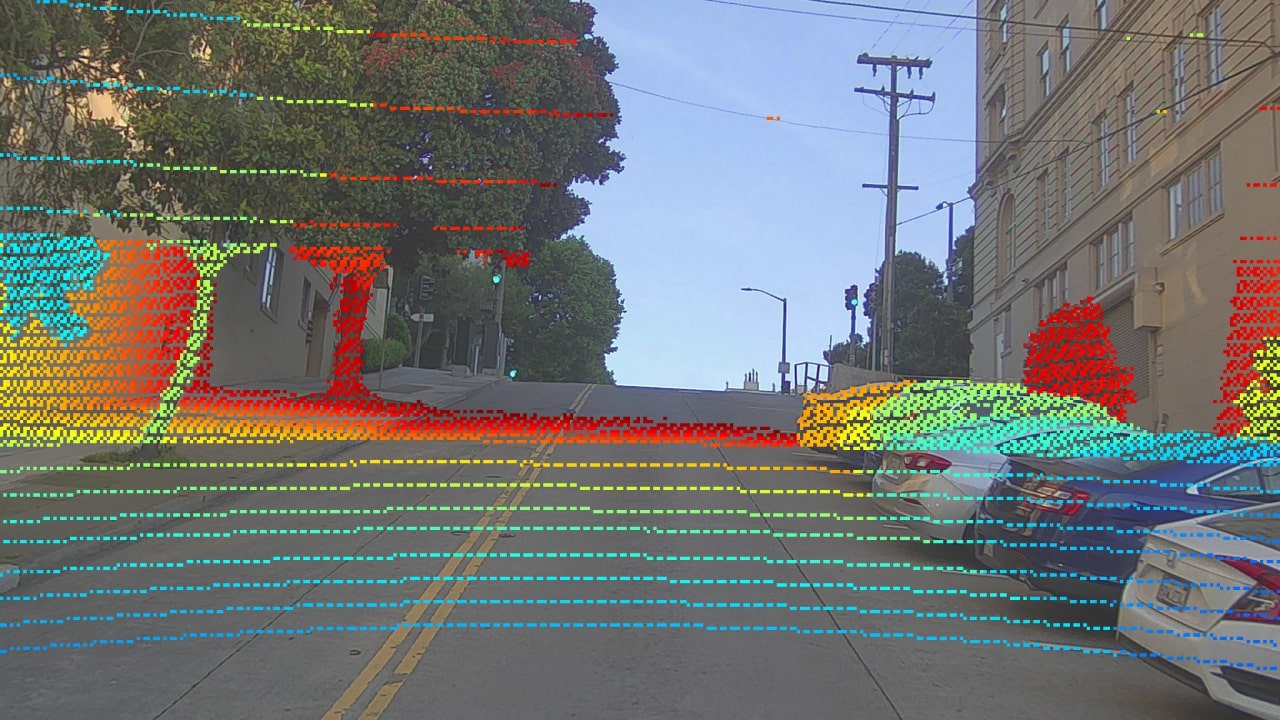} &
        \includegraphics[width=0.31\linewidth, trim={0 2cm 0 7cm}, clip]{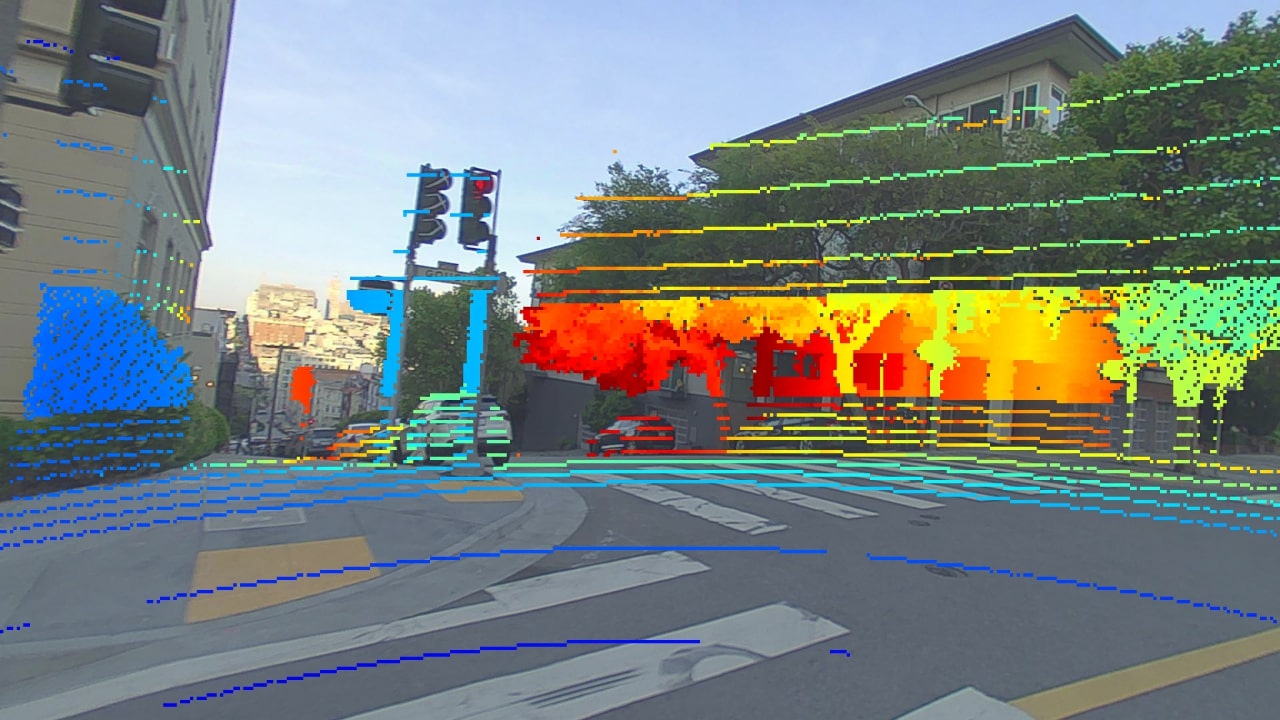} &
        \includegraphics[width=0.31\linewidth, trim={0 2cm 0 7cm}, clip]{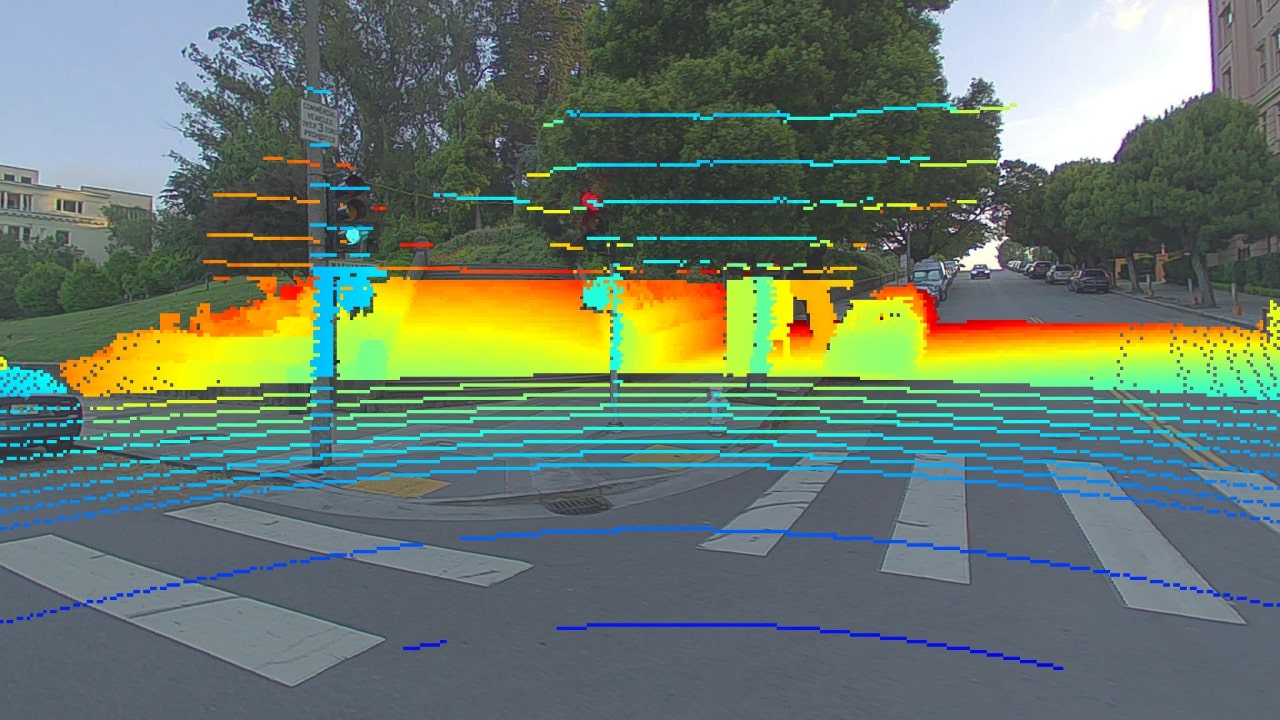}\\
        \rotatebox{90}{~~~~~~MOISST~\cite{herau2023moisst}} &
        \includegraphics[width=0.31\linewidth, trim={0 2cm 0 7cm}, clip]{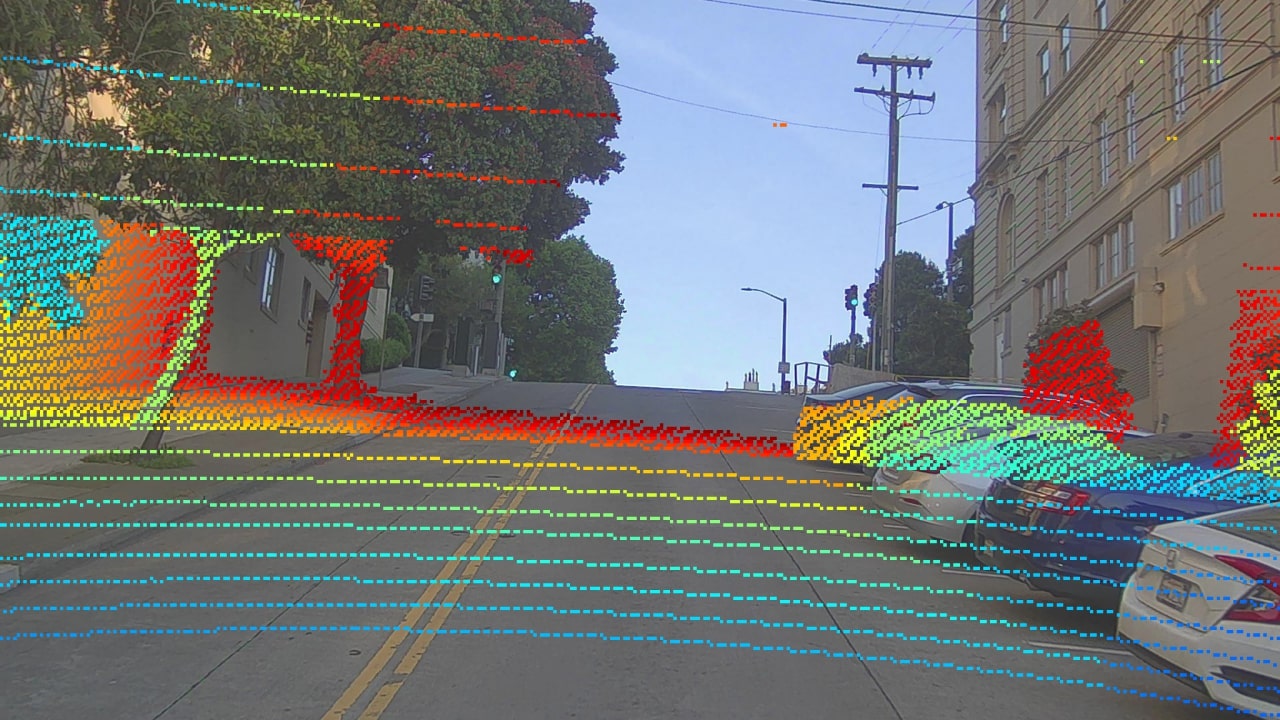} &
        \includegraphics[width=0.31\linewidth, trim={0 2cm 0 7cm}, clip]{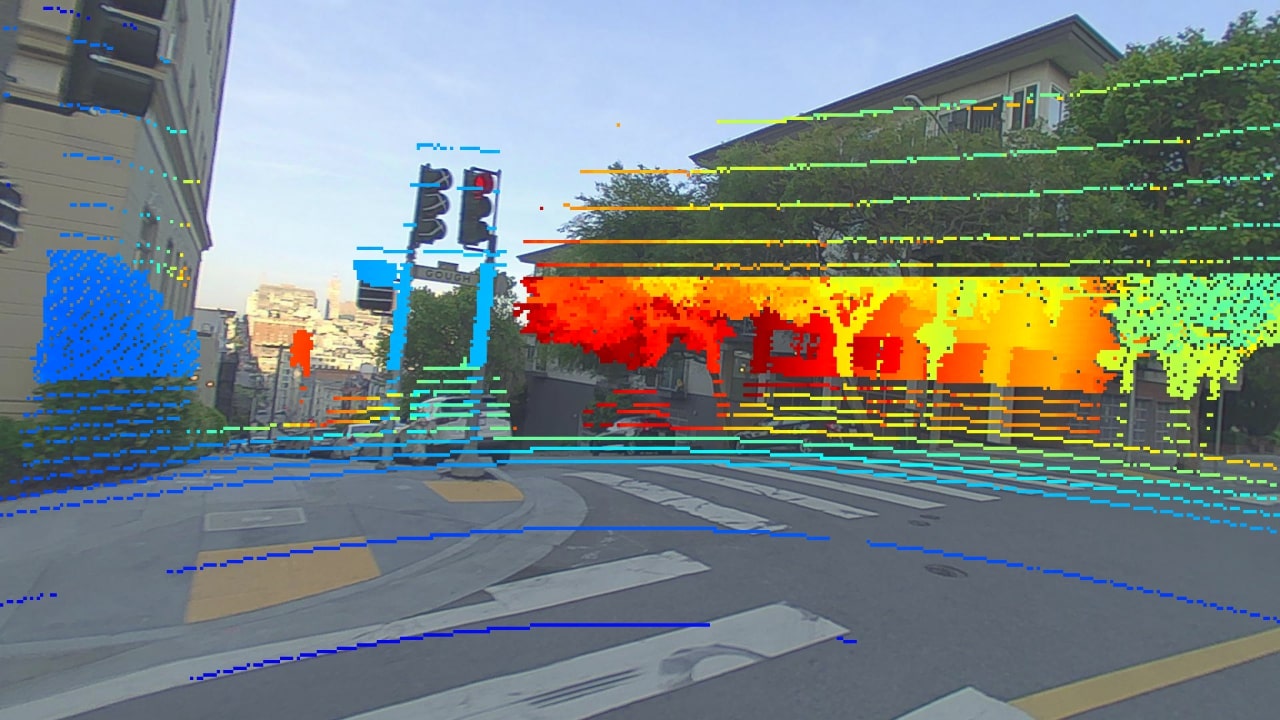} &
        \includegraphics[width=0.31\linewidth, trim={0 2cm 0 7cm}, clip]{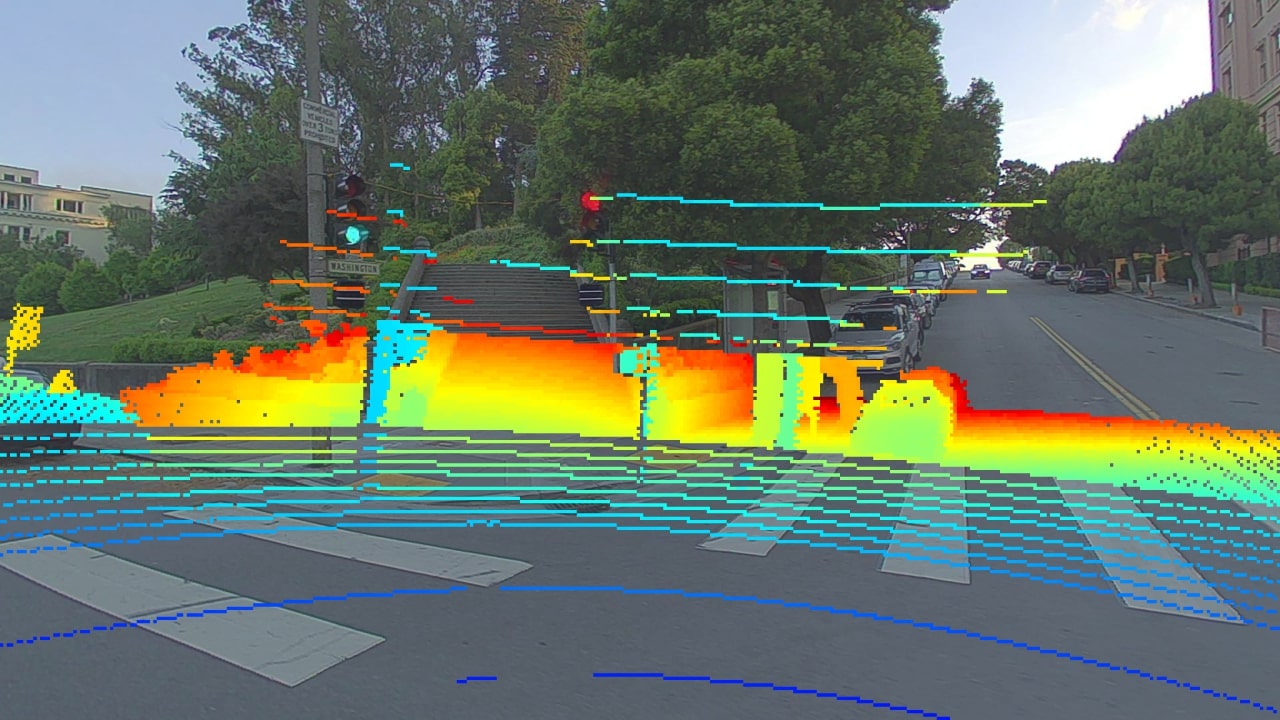}\\
        \rotatebox{90}{~~~~~SOAC~(ours)} &
        \includegraphics[width=0.31\linewidth, trim={0 2cm 0 7cm}, clip]{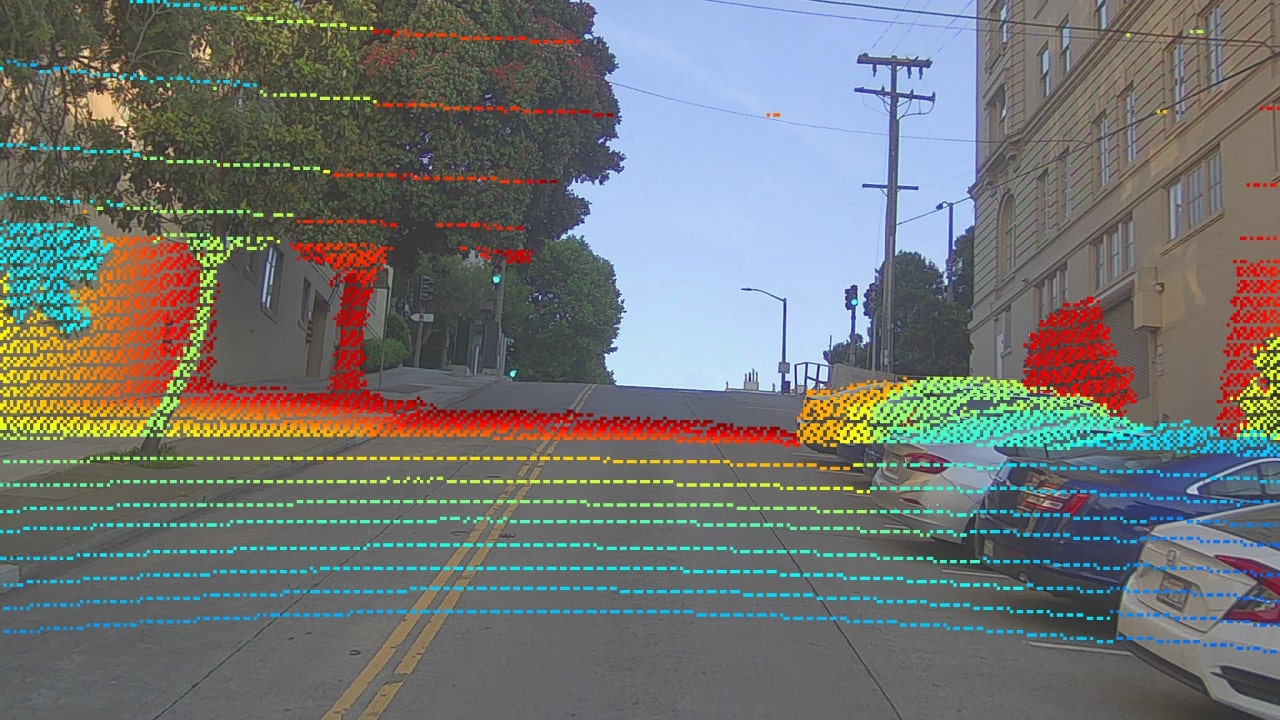} &
        \includegraphics[width=0.31\linewidth, trim={0 2cm 0 7cm}, clip]{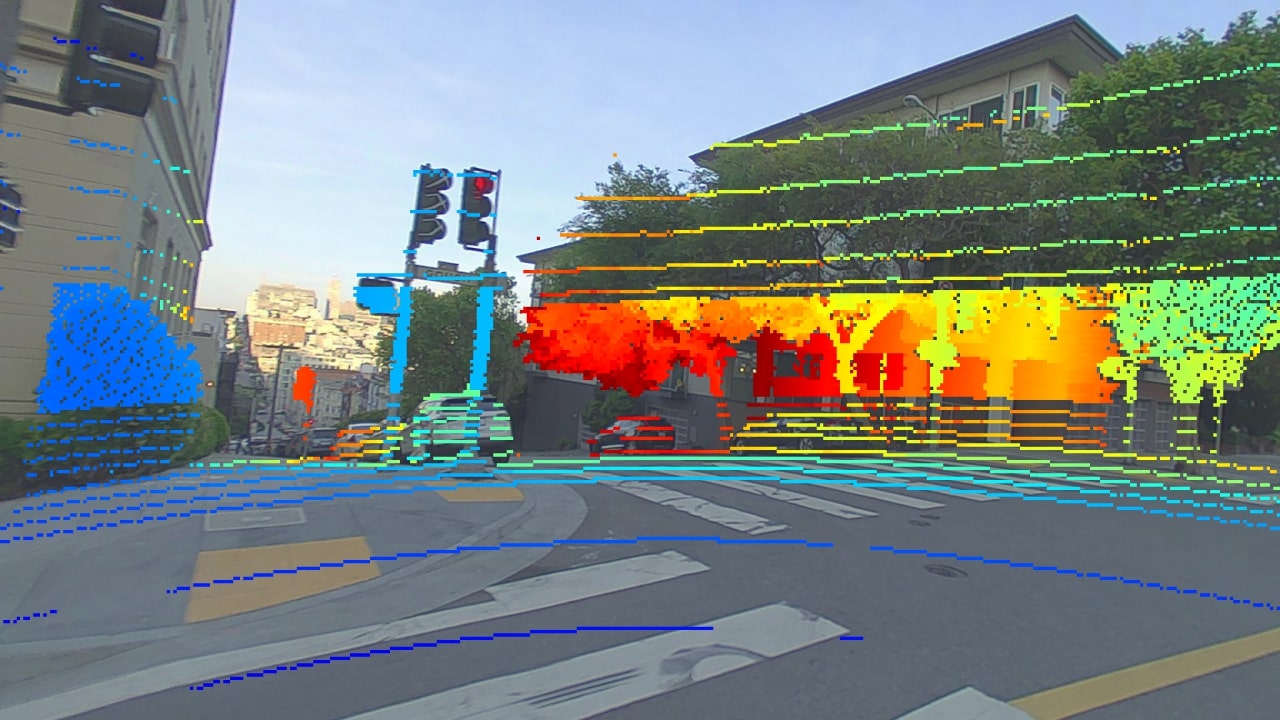} &
        \includegraphics[width=0.31\linewidth, trim={0 2cm 0 7cm}, clip]{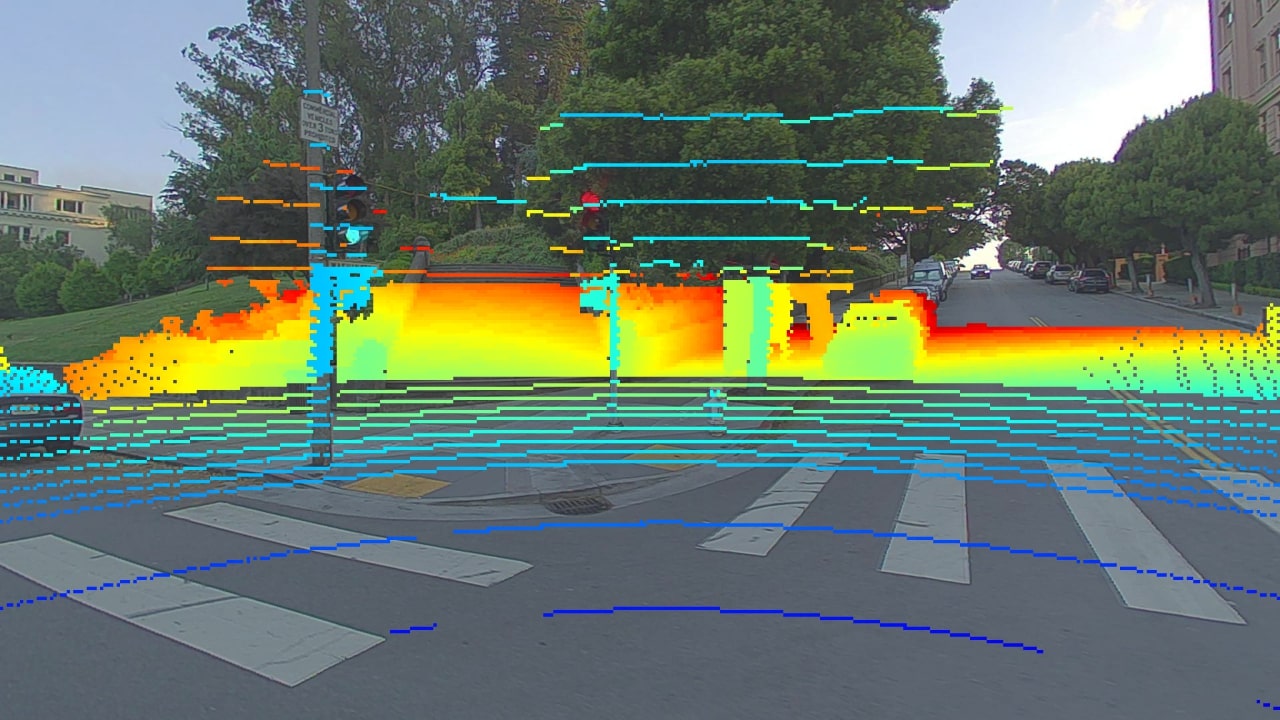}   

    \end{tabular}
\caption{Qualitative LiDAR/Camera reprojection results on Pandaset~\cite{xiao2021pandaset} dataset.}
\vspace{+1cm}
\label{fig:reproj_pandaset_supp}
\end{figure*}

\begin{figure*}[!htbp]
\centering
\scriptsize
\setlength{\tabcolsep}{0.002\linewidth}
    \begin{tabular}{lccc}  
        \rotatebox{90}{~~~~~~~~~~~~GT} & 
        \includegraphics[width=0.31\linewidth, trim={0 2cm 0 7cm}, clip]{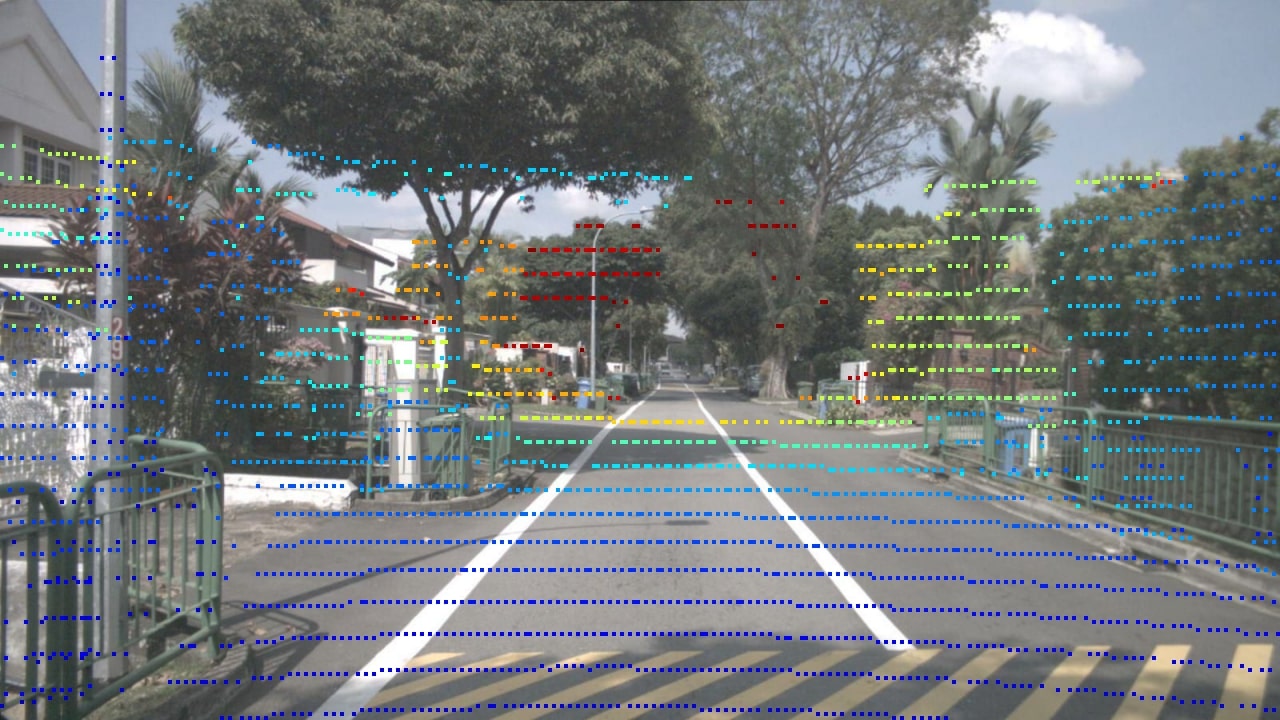} &
        \includegraphics[width=0.31\linewidth, trim={0 2cm 0 7cm}, clip]{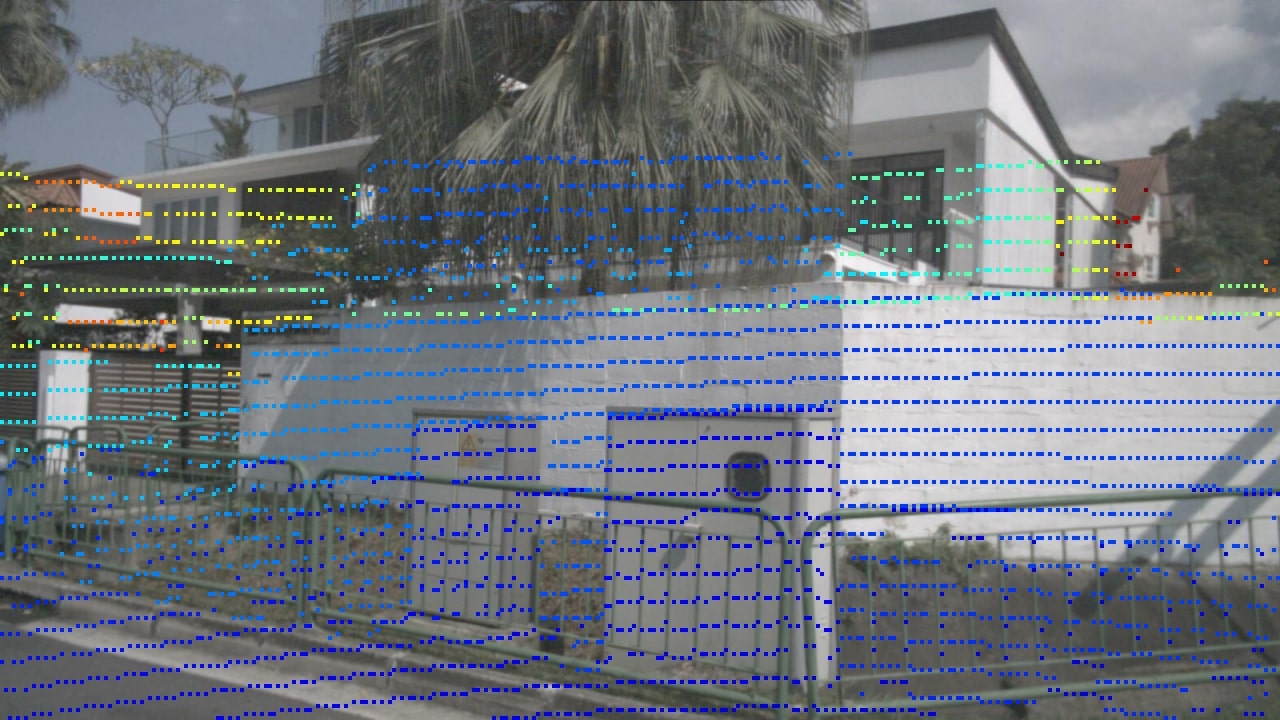} &
        \includegraphics[width=0.31\linewidth, trim={0 2cm 0 7cm}, clip]{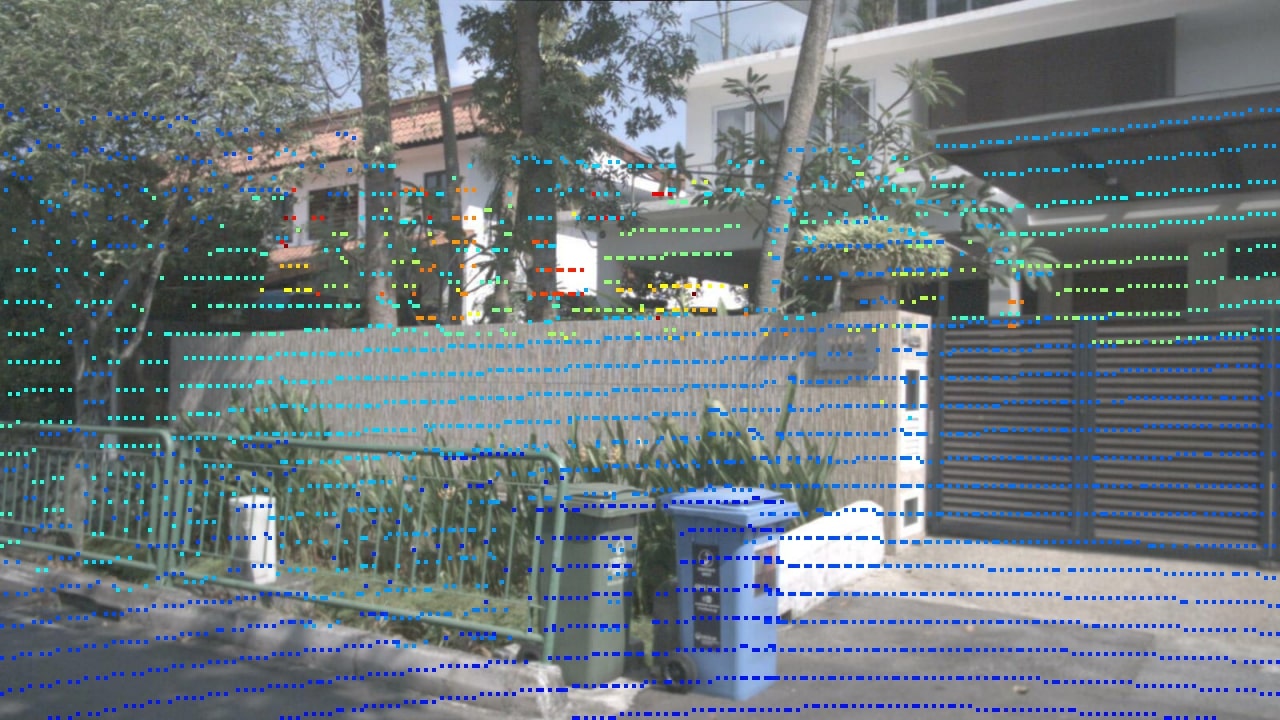}\\
   
        \rotatebox{90}{~~~~~~MOISST~\cite{herau2023moisst}} &
        \includegraphics[width=0.31\linewidth, trim={0 2cm 0 7cm}, clip]{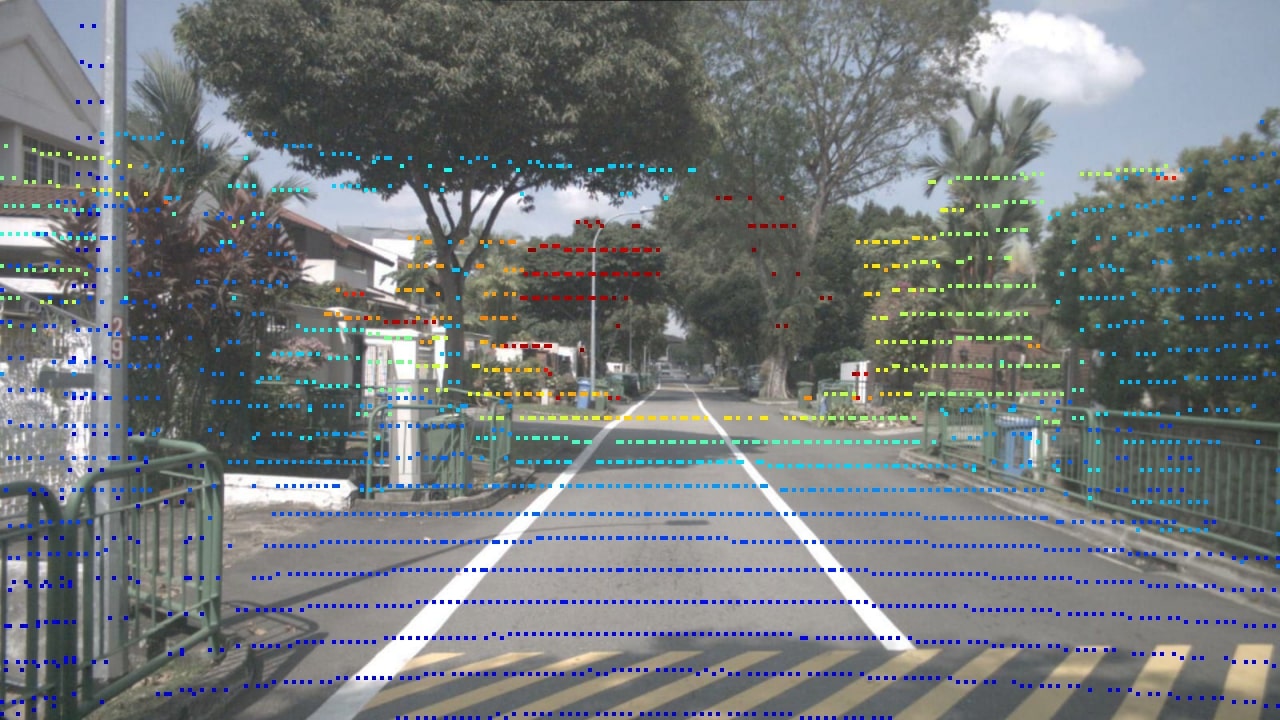} &
        \includegraphics[width=0.31\linewidth, trim={0 2cm 0 7cm}, clip]{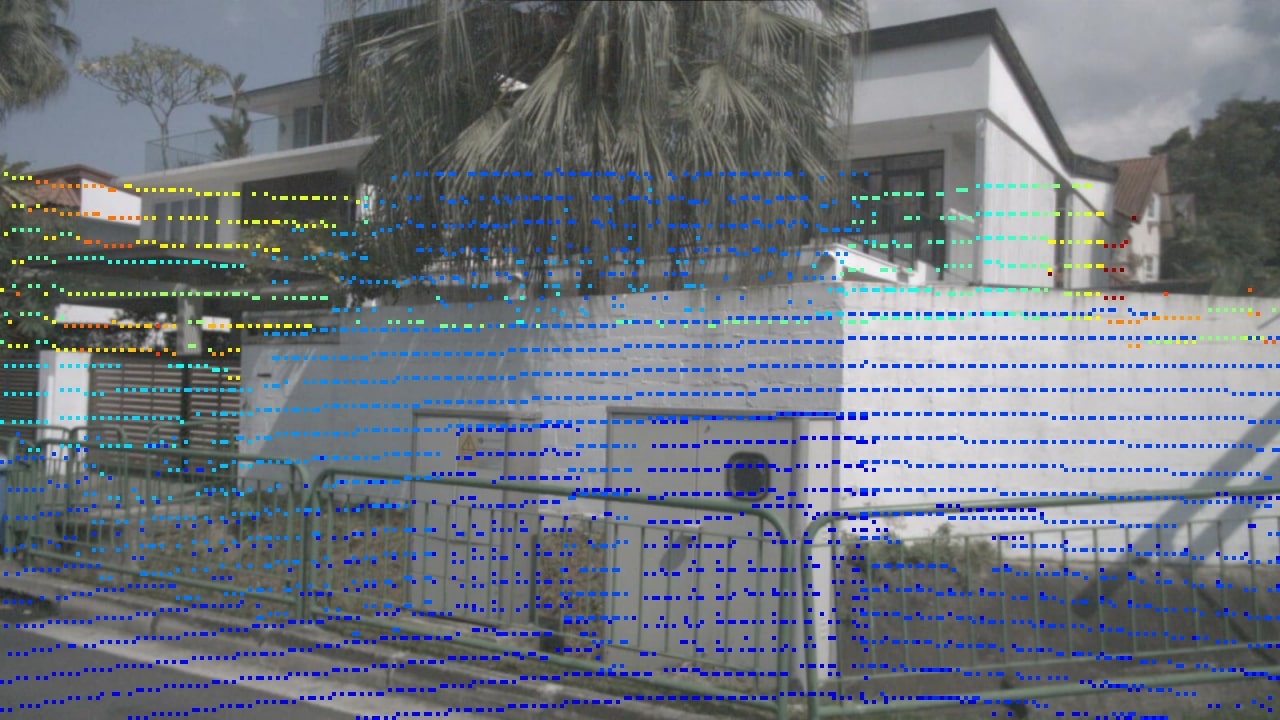} &
        \includegraphics[width=0.31\linewidth, trim={0 2cm 0 7cm}, clip]{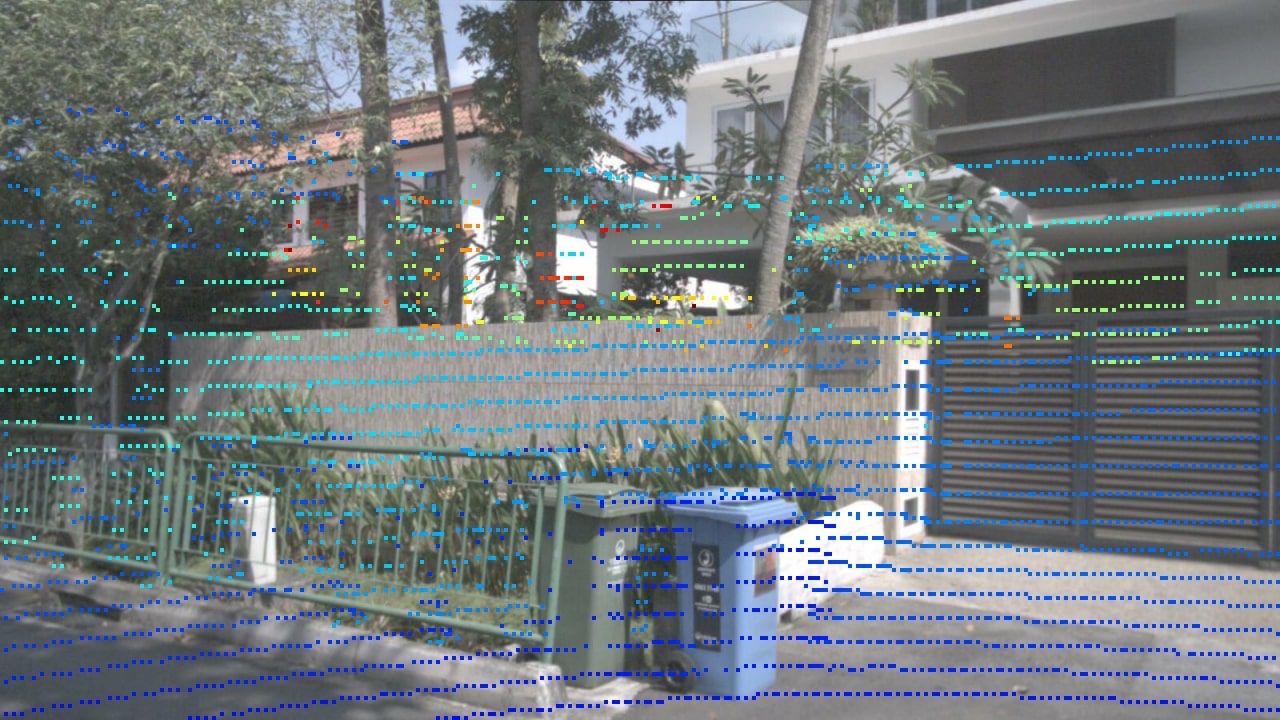}\\

        \rotatebox{90}{~~~~~SOAC~(ours)} &
        \includegraphics[width=0.31\linewidth, trim={0 2cm 0 7cm}, clip]{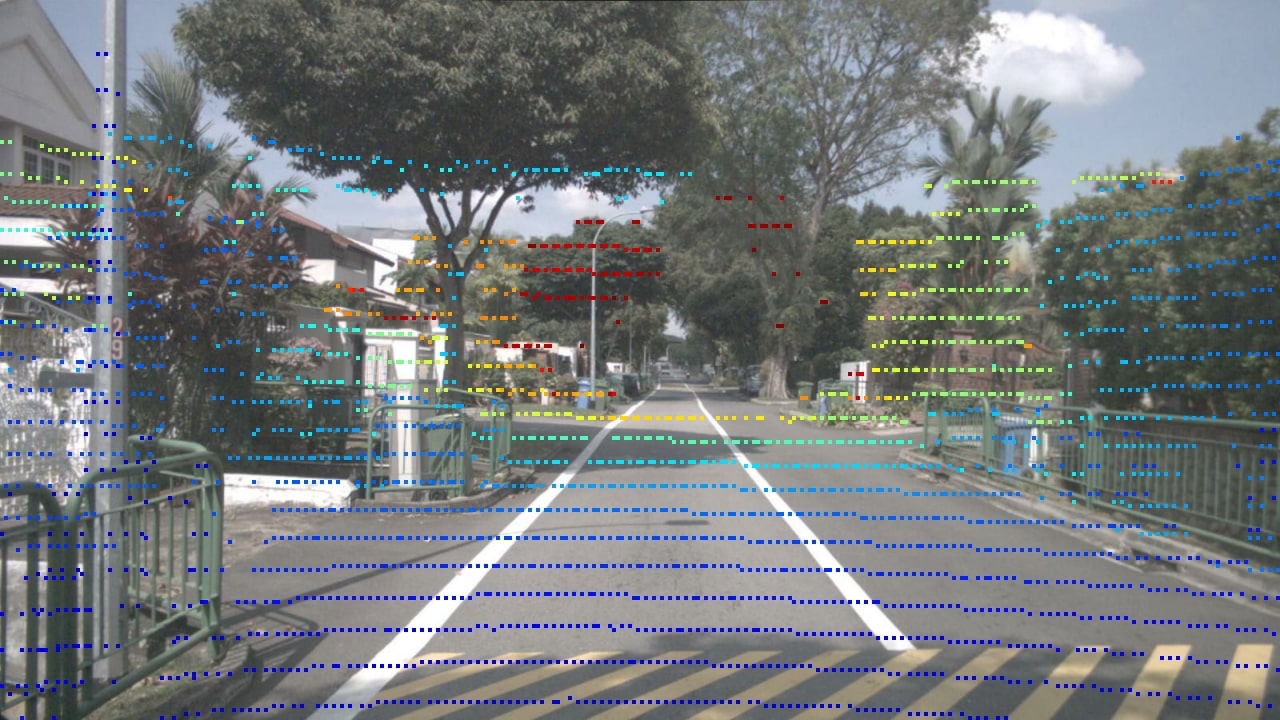} &
        \includegraphics[width=0.31\linewidth, trim={0 2cm 0 7cm}, clip]{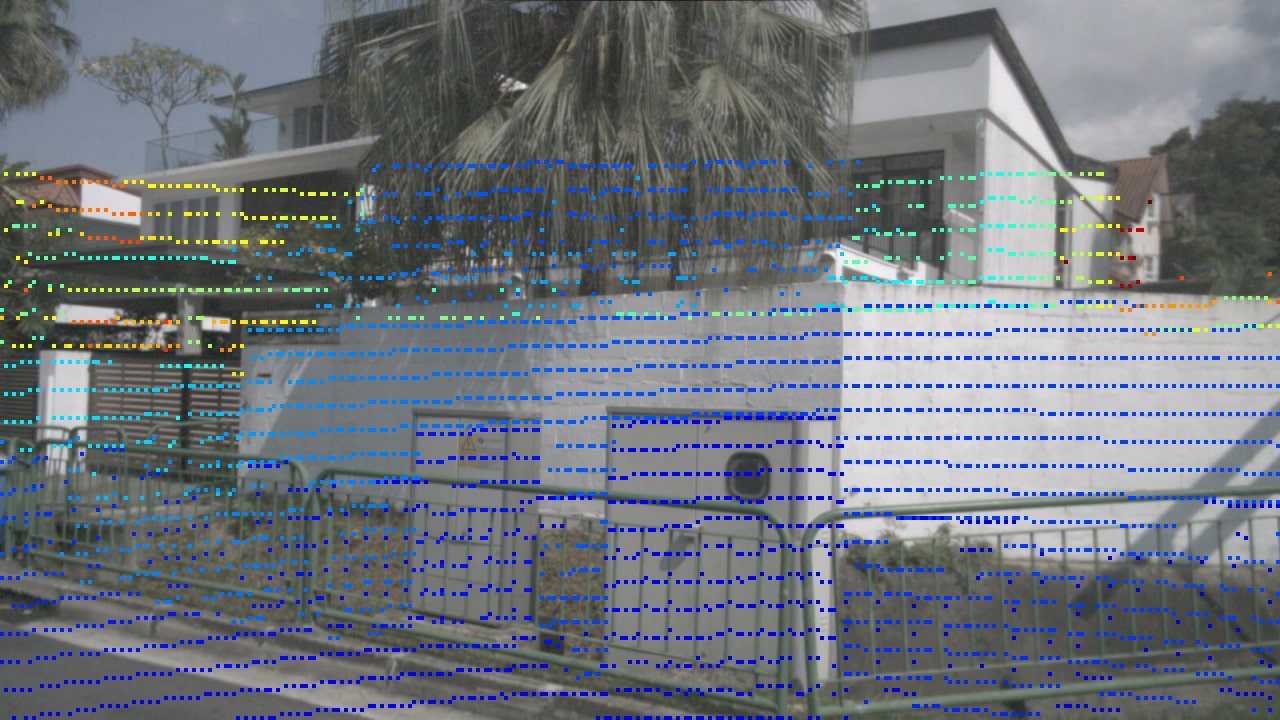} &
        \includegraphics[width=0.31\linewidth, trim={0 2cm 0 7cm}, clip]{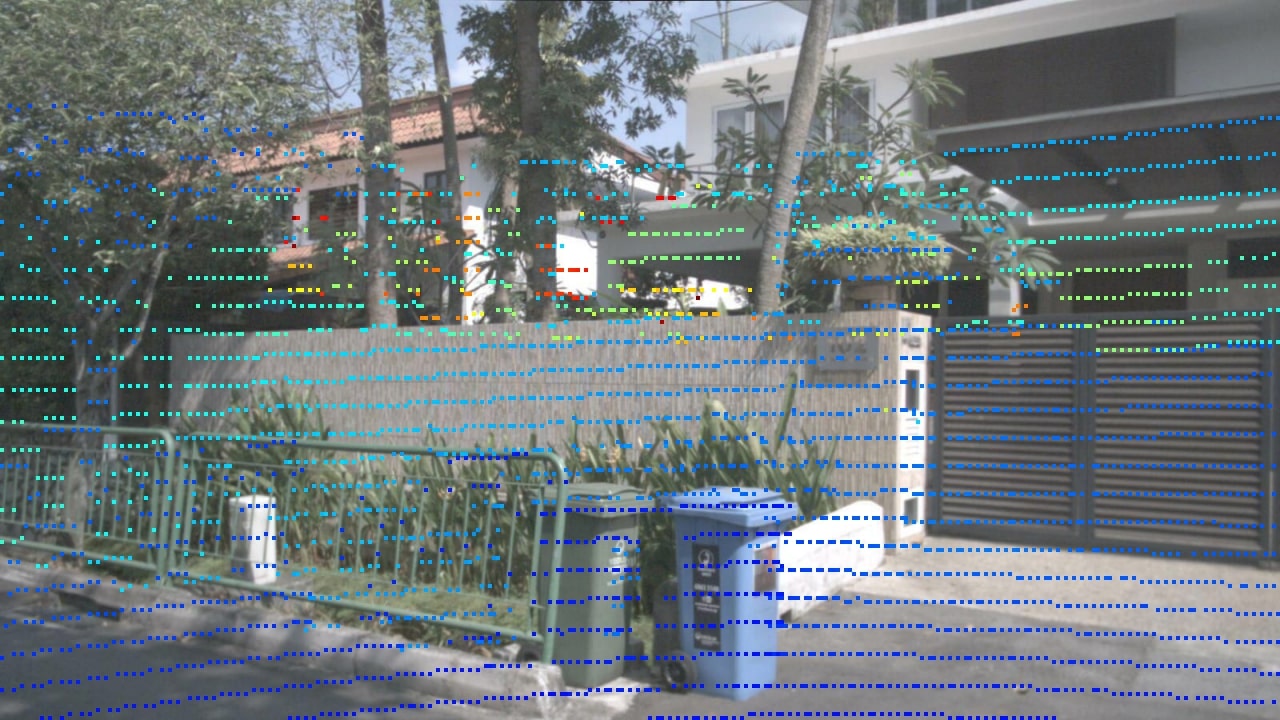}
        \\
        \midrule
        
        \rotatebox{90}{~~~~~~~~~~~~GT} & 
        \includegraphics[width=0.31\linewidth, trim={0 2cm 0 7cm}, clip]{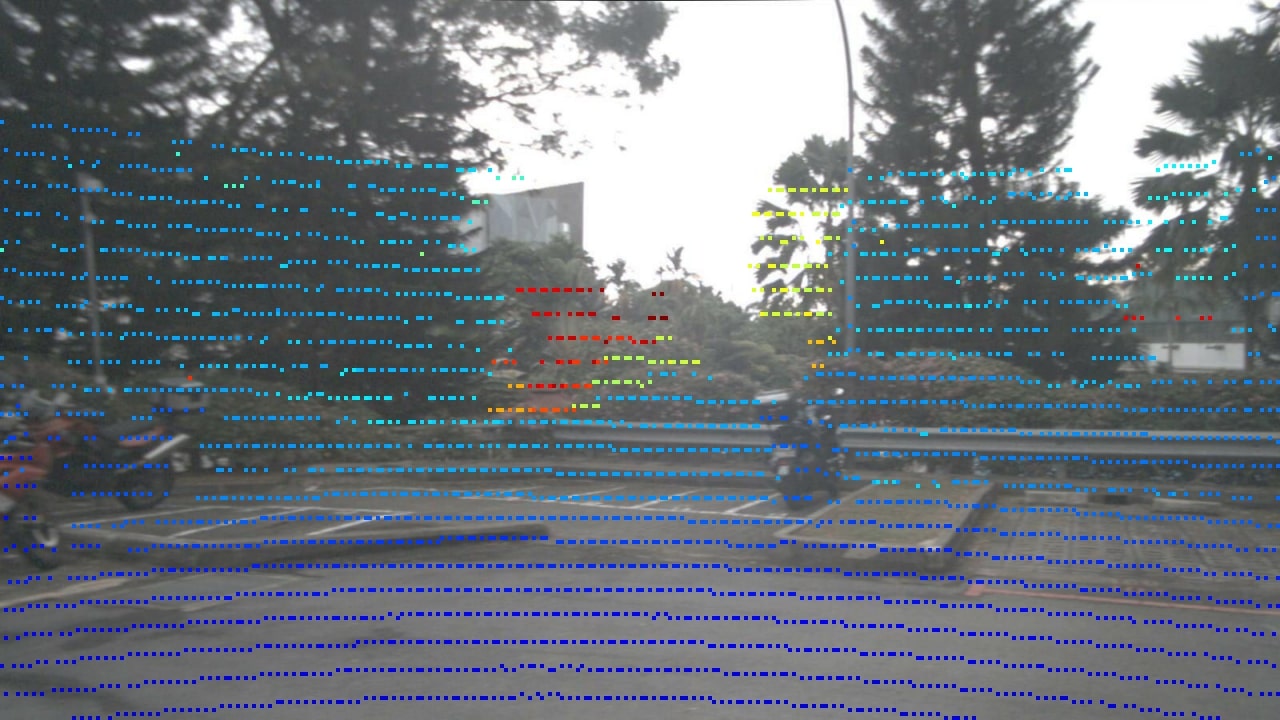} &
        \includegraphics[width=0.31\linewidth, trim={0 2cm 0 7cm}, clip]{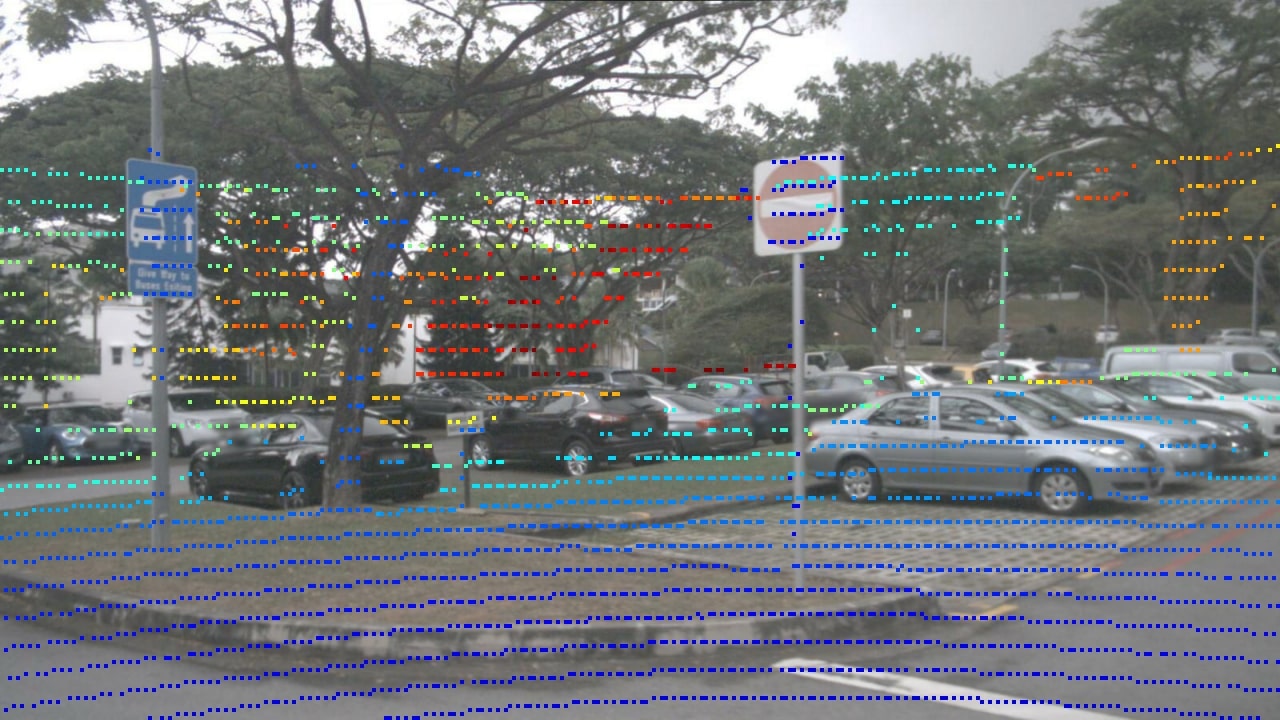} &
        \includegraphics[width=0.31\linewidth, trim={0 2cm 0 7cm}, clip]{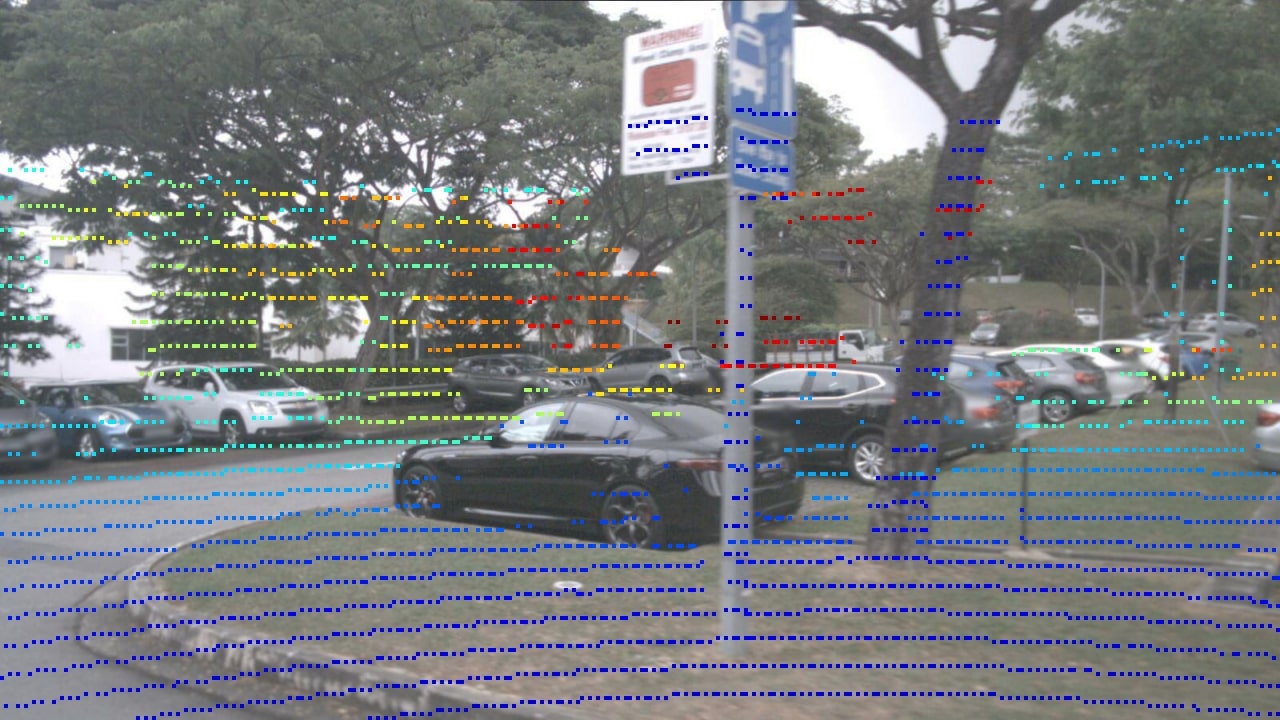}\\
        \rotatebox{90}{~~~~~~MOISST~\cite{herau2023moisst}} &
        \includegraphics[width=0.31\linewidth, trim={0 2cm 0 7cm}, clip]{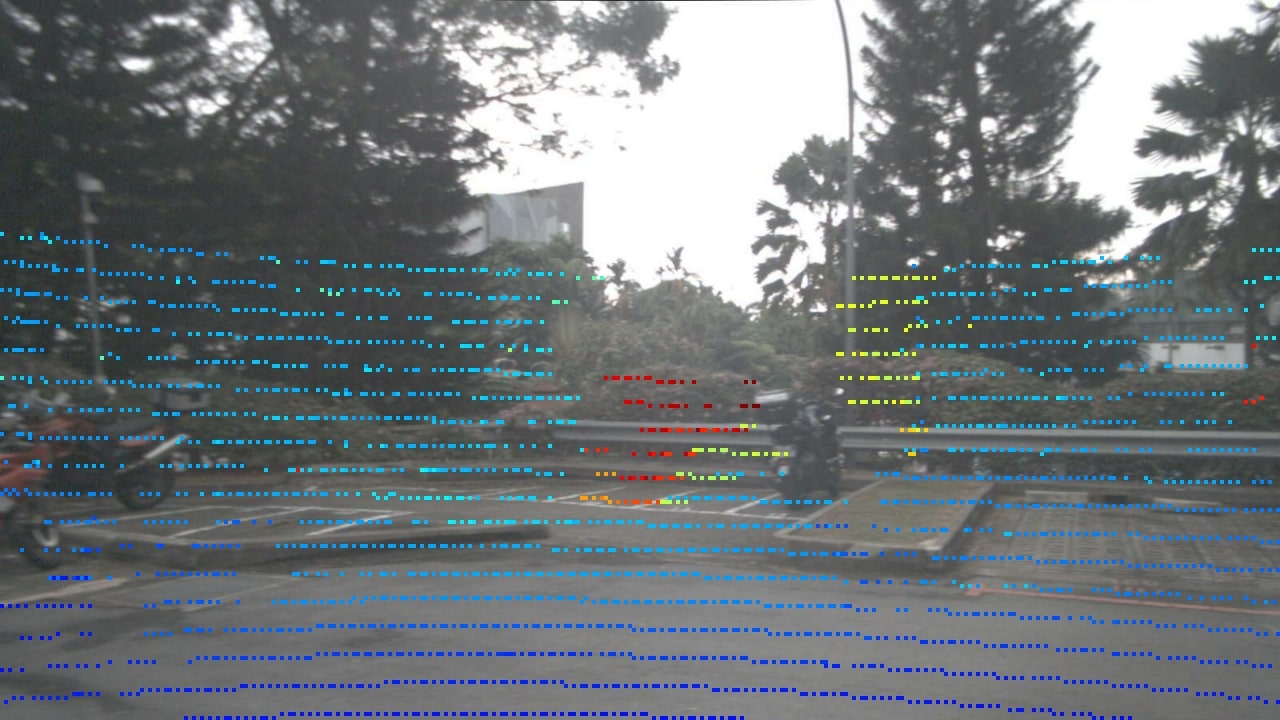} &
        \includegraphics[width=0.31\linewidth, trim={0 2cm 0 7cm}, clip]{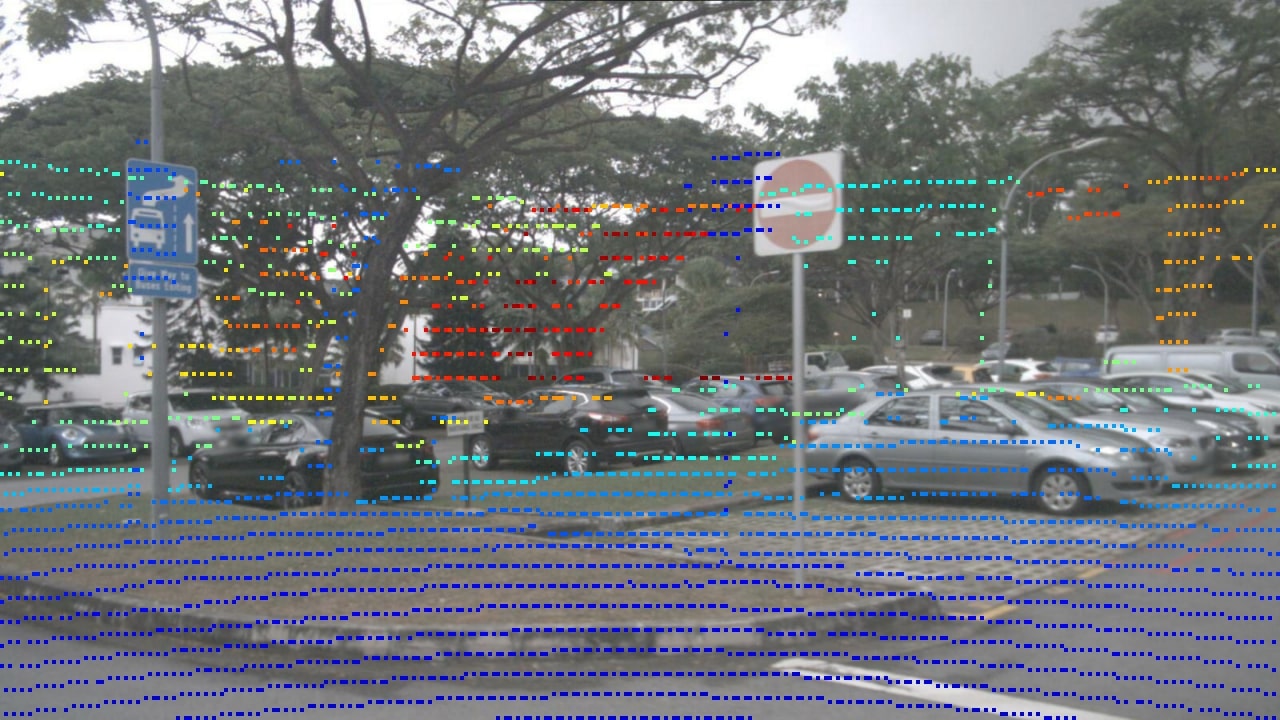} &
        \includegraphics[width=0.31\linewidth, trim={0 2cm 0 7cm}, clip]{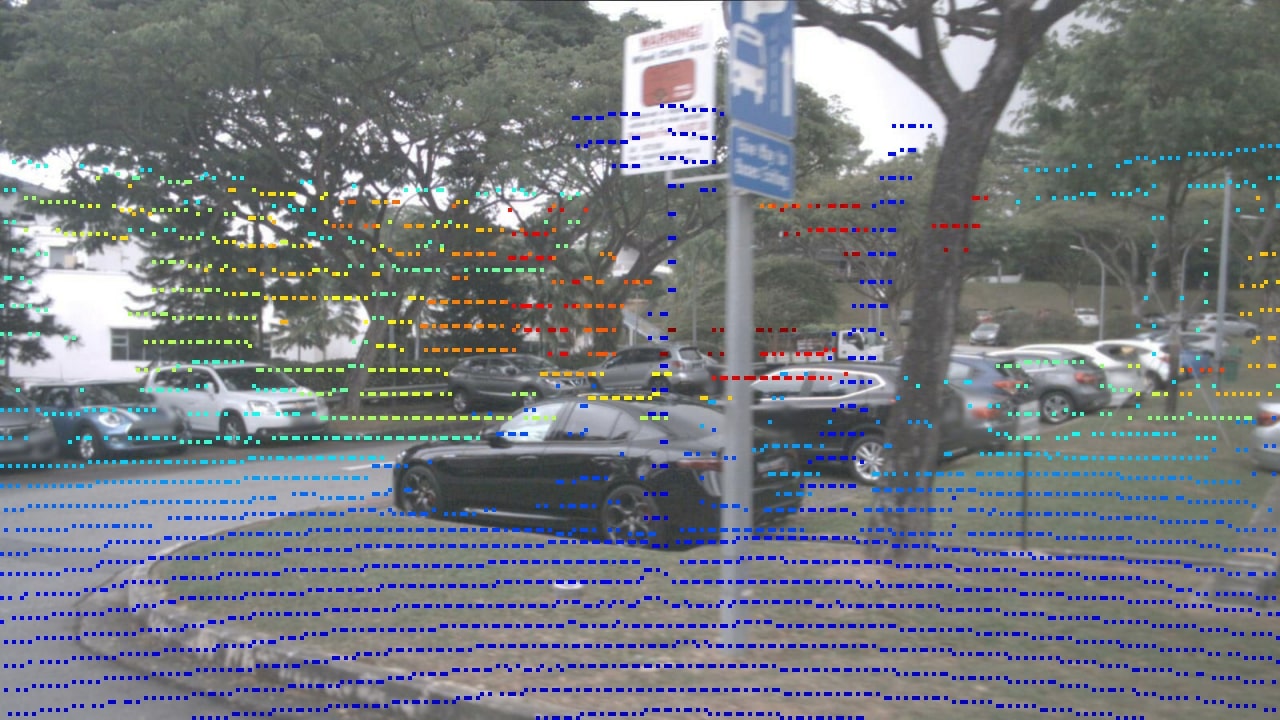}\\
                \rotatebox{90}{~~~~~SOAC~(ours)}  &
        \includegraphics[width=0.31\linewidth, trim={0 2cm 0 7cm}, clip]{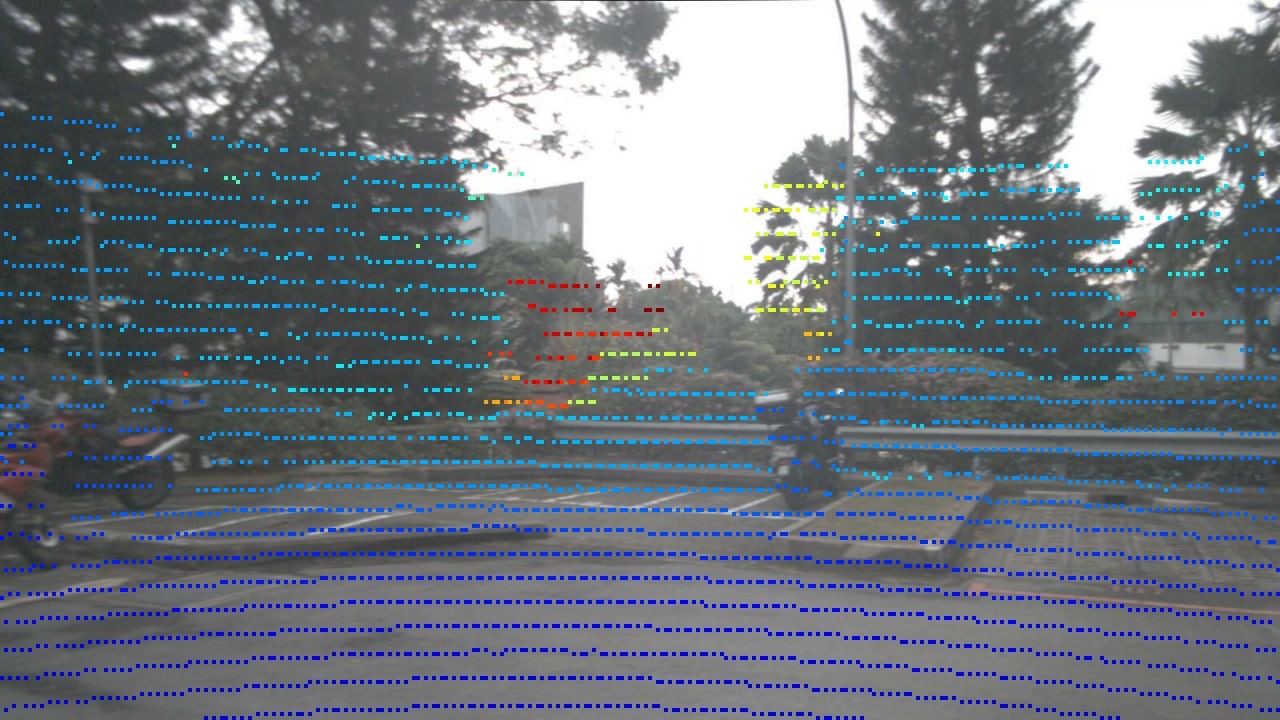} &
        \includegraphics[width=0.31\linewidth, trim={0 2cm 0 7cm}, clip]{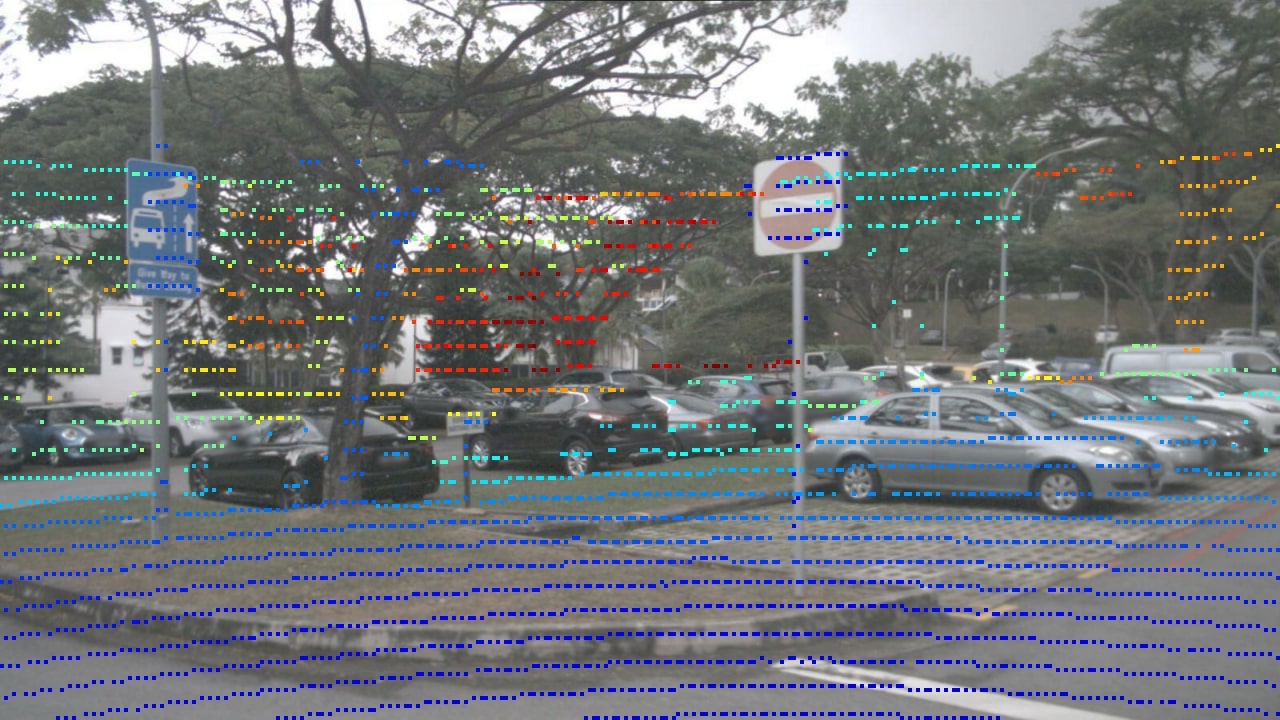} &
        \includegraphics[width=0.31\linewidth, trim={0 2cm 0 7cm}, clip]{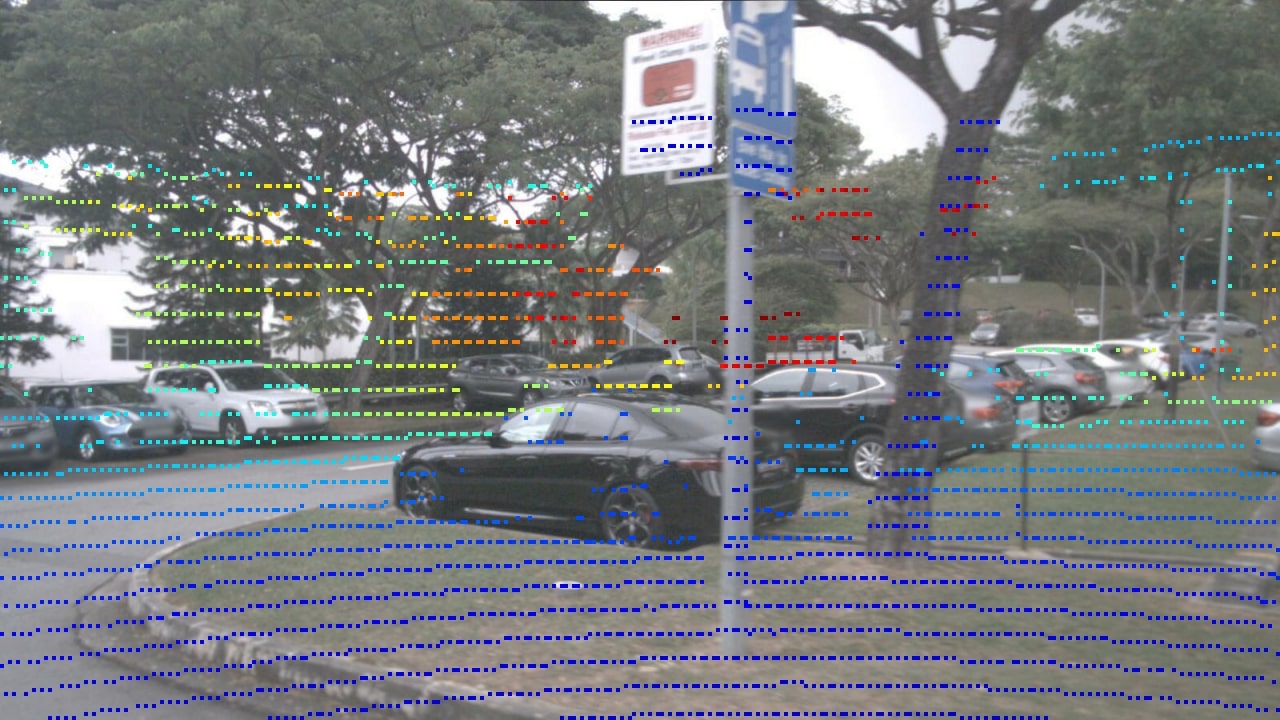}

    \end{tabular}
\caption{More qualitative LiDAR/Camera reprojection results on nuScenes~\cite{caesar2020nuscenes} dataset.}
\label{fig:reproj_nuscenes_supp}
\end{figure*}

In Fig.~\ref{fig:reproj_nuscenes_supp} and Fig.~\ref{fig:reproj_pandaset_supp} are shown LiDAR/Camera projection on nuScenes and Pandaset sequences. The calibration optimized by SOAC provides substantially better alignment than the one from MOISST. 

\begin{figure*}[!htbp]
\centering
\scriptsize
\setlength{\tabcolsep}{0.002\linewidth}
    \begin{tabular}{cccccc}
        \multicolumn{6}{c}{Input images}\\

         \includegraphics[width=0.16\linewidth]{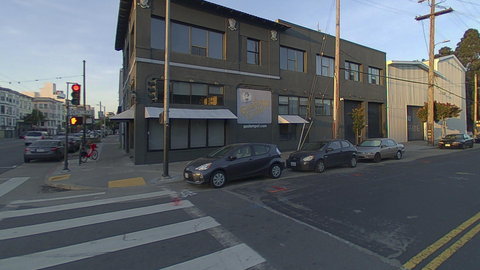} & 
         \includegraphics[width=0.16\linewidth]{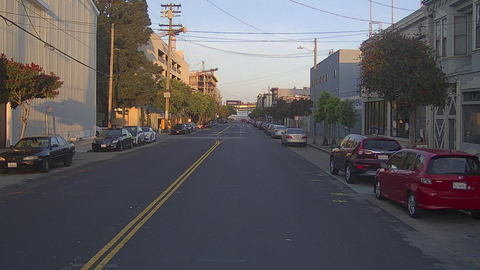} &
         \includegraphics[width=0.16\linewidth]{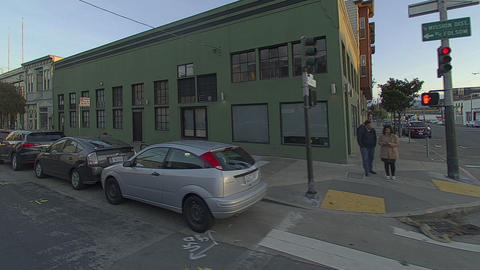}&
         \includegraphics[width=0.16\linewidth]{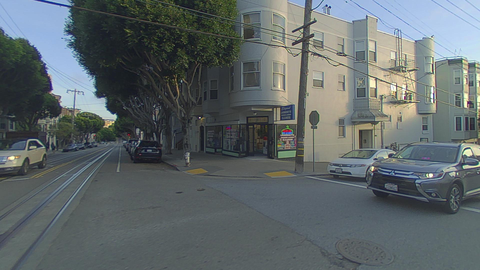} & 
         \includegraphics[width=0.16\linewidth]{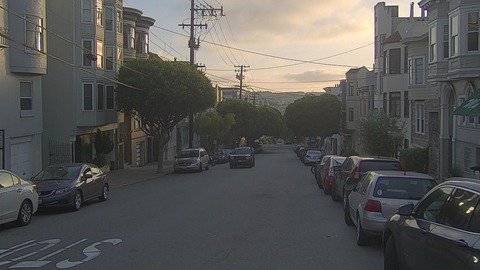} &
         \includegraphics[width=0.16\linewidth]{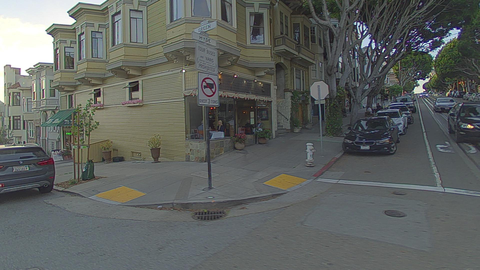}\\
         
        \multicolumn{6}{c}{Predictions from NeRF trained with front camera}\\ 
          \includegraphics[width=0.16\linewidth]{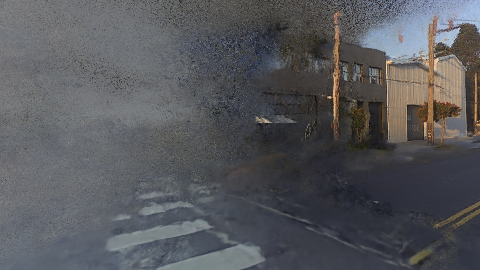} &
         \includegraphics[width=0.16\linewidth]{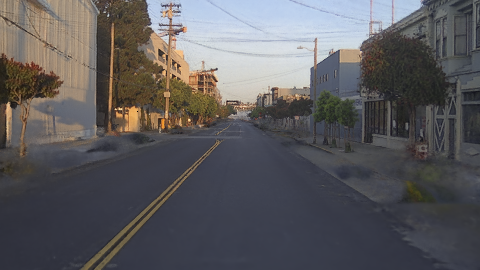} &
         \includegraphics[width=0.16\linewidth]{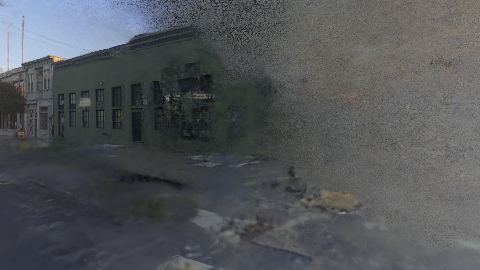}&
        \includegraphics[width=0.16\linewidth]{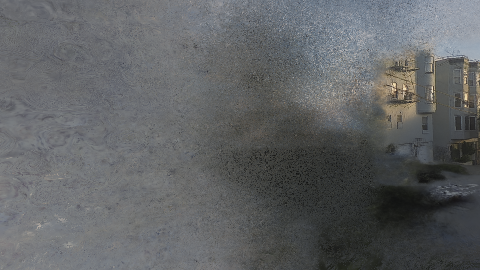} &
         \includegraphics[width=0.16\linewidth]{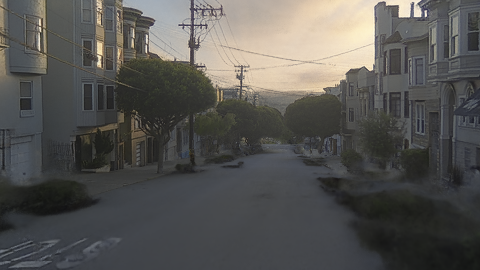} &
         \includegraphics[width=0.16\linewidth]{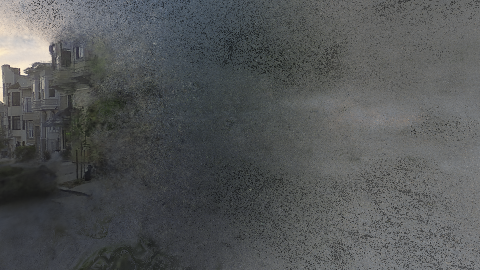}\\

          \includegraphics[width=0.16\linewidth]{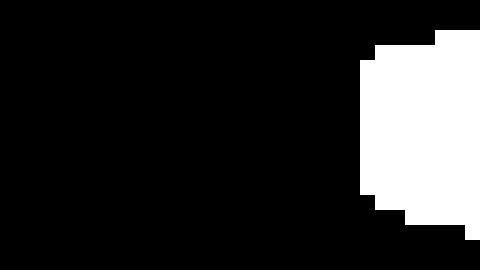} &
         \includegraphics[width=0.16\linewidth]{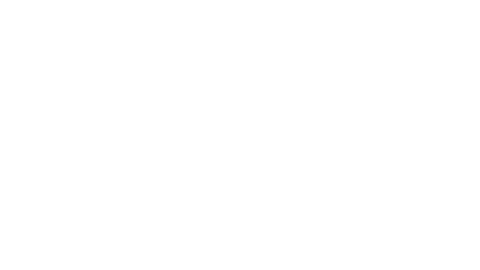} &
         \includegraphics[width=0.16\linewidth]{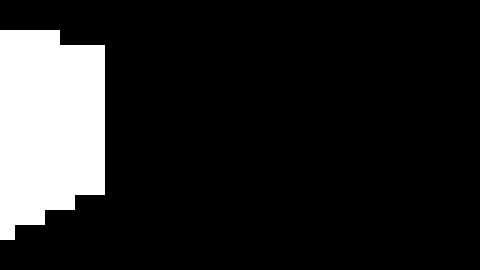}&
          \includegraphics[width=0.16\linewidth]{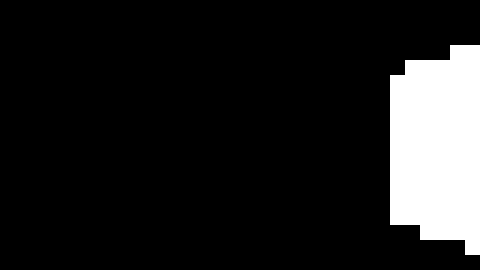} &
         \includegraphics[width=0.16\linewidth]{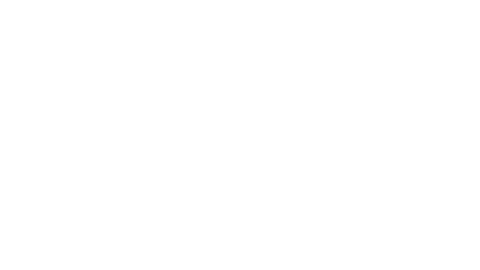} &
         \includegraphics[width=0.16\linewidth]{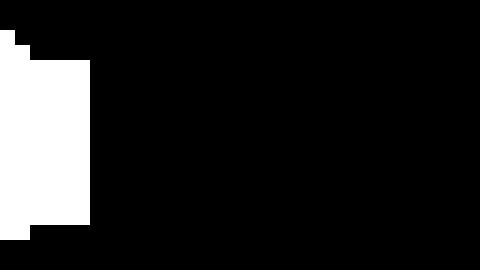}\\

          \multicolumn{6}{c}{Predictions from NeRF trained with left camera} \\
          \includegraphics[width=0.16\linewidth]{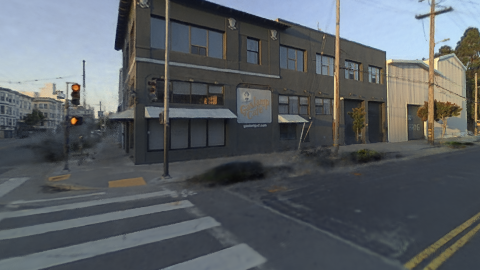} &
         \includegraphics[width=0.16\linewidth]{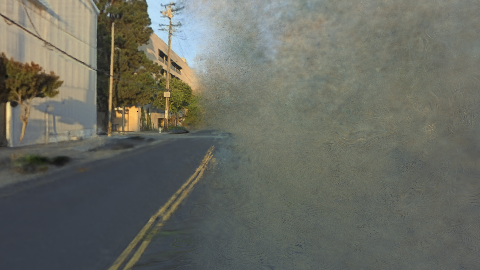} &
         \includegraphics[width=0.16\linewidth]{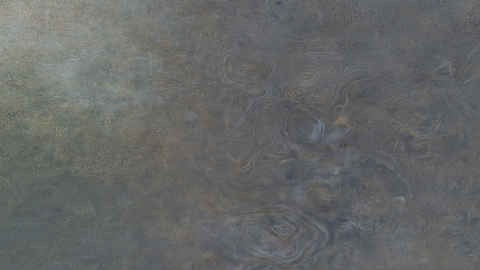} &
          \includegraphics[width=0.16\linewidth]{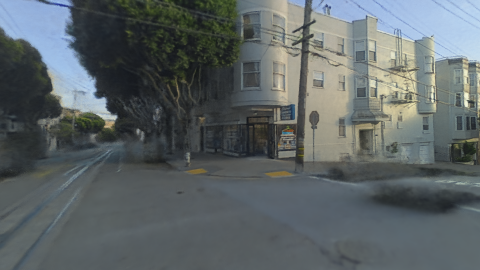} &
         \includegraphics[width=0.16\linewidth]{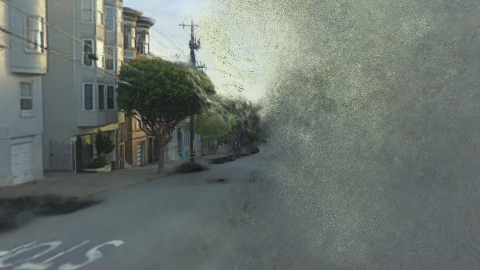} &
         \includegraphics[width=0.16\linewidth]{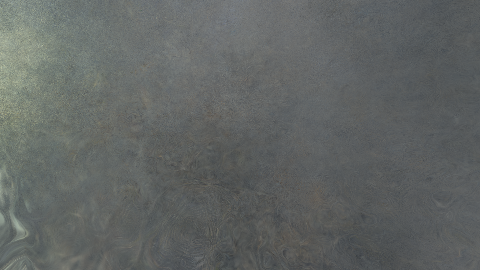}\\

          \includegraphics[width=0.16\linewidth]{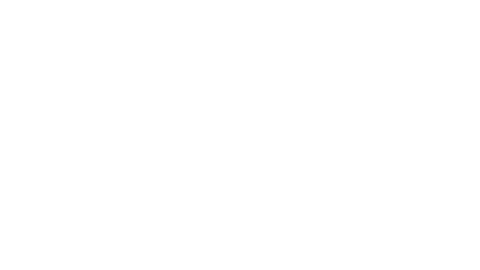} &
         \includegraphics[width=0.16\linewidth]{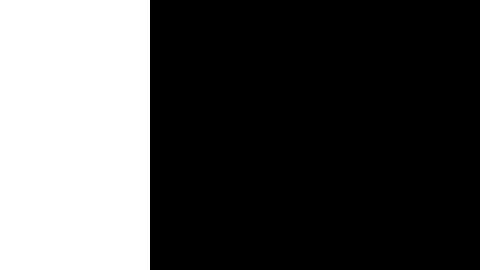} &
         \includegraphics[width=0.16\linewidth]{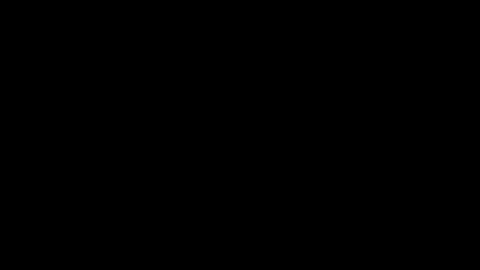} &
          \includegraphics[width=0.16\linewidth]{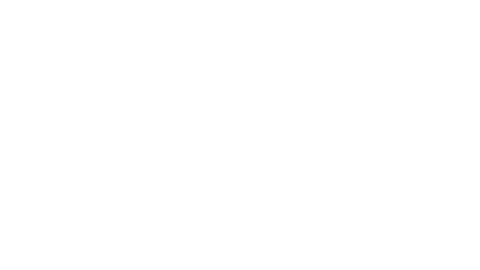} &
         \includegraphics[width=0.16\linewidth]{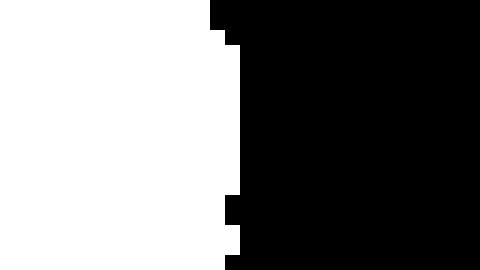} &
         \includegraphics[width=0.16\linewidth]{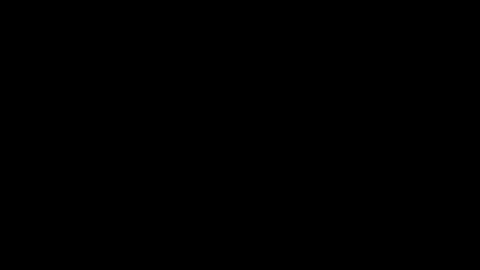}\\

         \multicolumn{6}{c}{Predictions from NeRF trained with right camera}\\
          \includegraphics[width=0.16\linewidth]{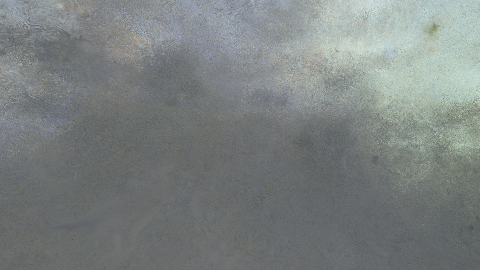} &
         \includegraphics[width=0.16\linewidth]{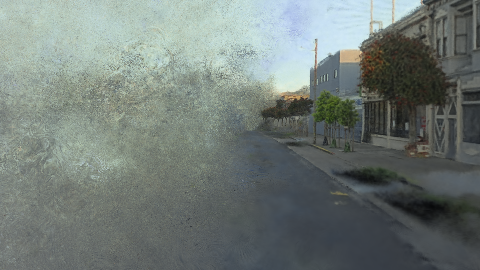} &
         \includegraphics[width=0.16\linewidth]{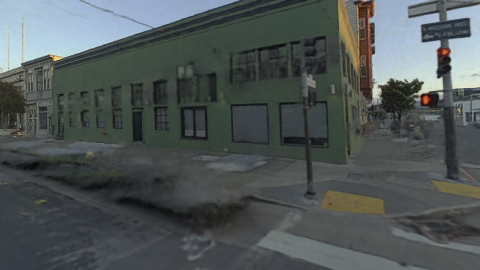}&
        \includegraphics[width=0.16\linewidth]{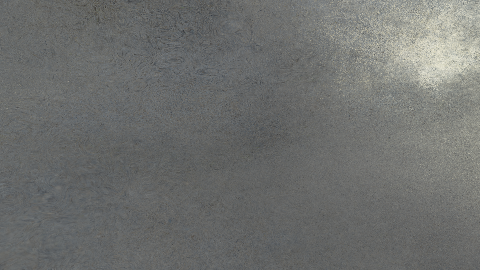} &
         \includegraphics[width=0.16\linewidth]{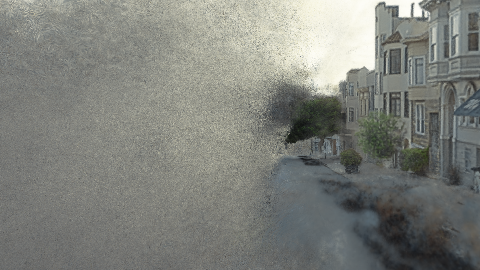} &
         \includegraphics[width=0.16\linewidth]{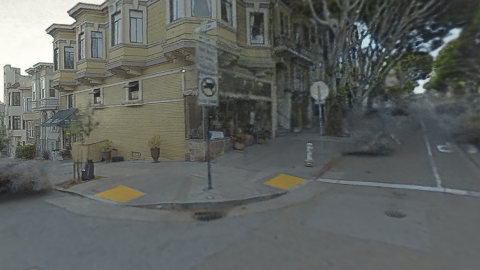}\\

          \includegraphics[width=0.16\linewidth]{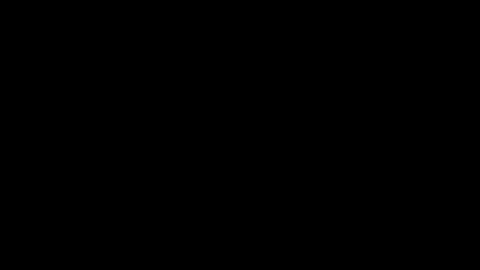} &
         \includegraphics[width=0.16\linewidth]{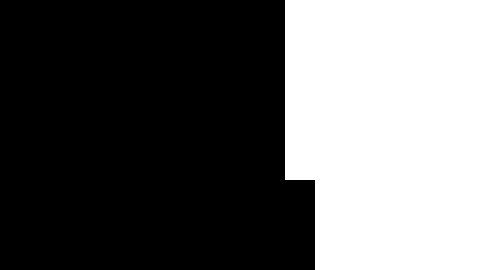} &
         \includegraphics[width=0.16\linewidth]{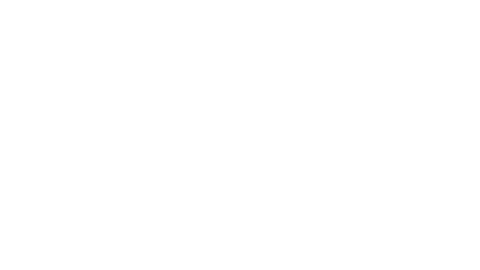} &
          \includegraphics[width=0.16\linewidth]{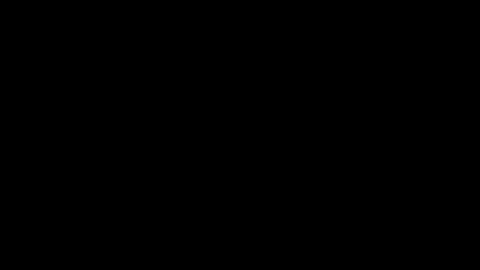} &
         \includegraphics[width=0.16\linewidth]{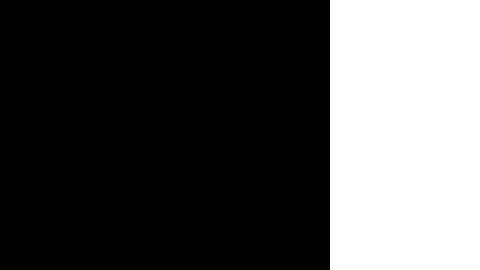} &
         \includegraphics[width=0.16\linewidth]{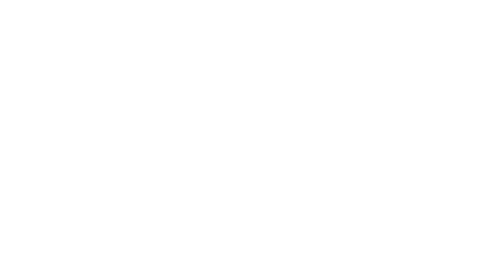}\\

    \end{tabular}
\caption{Results using visibility grids on a Pandaset~\cite{xiao2021pandaset} sequence -- Prediction from different NeRFs.}
\label{fig:mask_supp}
\end{figure*}

Fig.~\ref{fig:mask_supp} shows the predicted images and masks from each NeRF trained with different cameras. The visibility masks are coherent with the predicted RGB images, allowing correct filtering for SOAC.

\end{document}